\documentclass[10pt,twocolumn,letterpaper]{article}

\usepackage[pagenumbers]{cvpr} % To force page numbers, e.g. for an arXiv version

\usepackage{graphicx}
\usepackage{amsmath,amssymb}
\usepackage{mathtools}
\usepackage{xcolor}
\usepackage{booktabs} % For professional-looking tables
\usepackage{multirow} % For multi-row cells in tables
\usepackage{caption}
\usepackage{subcaption} % Required for side-by-side subfigures
\usepackage{natbib}   % Required for ieeenat_fullname bibstyle
\usepackage[inline]{enumitem}
\usepackage{csquotes}

\definecolor{cvprblue}{rgb}{0.21,0.49,0.74}
\usepackage[pagebackref,breaklinks,colorlinks,allcolors=cvprblue]{hyperref}

\usepackage[capitalize]{cleveref}
\crefname{section}{Sec.}{Secs.}
\Crefname{section}{Section}{Sections}
\Crefname{table}{Table}{Tables}
\crefname{table}{Tab.}{Tabs.}

\makeatletter
\newcommand\footnoteref[1]{\protected@xdef\@thefnmark{\ref{#1}}\@footnotemark}
\makeatother

\begin{document}

%%%%%%%%% PAPER ID  - PLEASE UPDATE
\def\paperID{16098} % *** Enter the Paper ID here
\def\confName{CVPR}
\def\confYear{2026}

\title{Weakly Supervised Concept Learning for Object-centric Visual Reasoning} % TODO: shorten title to save the line (exchange CV by just Vision?)

%Label efficient two-stage neuro-symbolic reasoning by weak and self- concept supervision

\author{
Sparsh Tiwari\\
University of Lübeck, Germany\\
{\tt\small sparsh.tiwari@uni-luebeck.de}
\and
Bettina Finzel\\
University of Bamberg, Germany\\
{\tt\small bettina.finzel@uni-bamberg.de}
\and
Gesina Schwalbe\\
University of Ulm, Germany\\
University of Lübeck, Germany\\
{\tt\small gesina.schwalbe@uni-ulm.de}
}

\maketitle

\begin{abstract}

Neuro-symbolic systems promise to combine deep neural network's (DNN) processing of raw sensor inputs with few-shot performance of symbolic artificial intelligence (AI).

Two-stage approaches explicitly decouple DNN-based perception from subsequent rule-based reasoning. This avoids optimization and interpretability issues of end-to-end differentiable approaches, but requires costly labels for the perception output.
This paper introduces an efficient weak supervision scheme for the perception stage to ground its output symbols for logical induction in object-centric reasoning tasks. % TODO: example?
It combines a slot-based architecture for object-centricity with a Variational Autoencoder (VAE) for self-supervision, competing with concept guidance on latent dimensions for human-interpretable grounding. The resulting predictions are translated into symbolic background knowledge for reasoning frameworks, such as Inductive Logic Programming (ILP), Decision Trees, and Bayesian Networks. 
Our extensive empirical evaluation on synthetic and real-world datasets shows that our approach can discover complex, abstract rules for object-centric reasoning whilst reducing supervision to as little as 1\% of labels, and being robust even under substantial domain shift.
Notably, at 1\% supervision it even outperforms state-of-the-art foundation model baselines in domain generalization.
\end{abstract}

\begin{figure}
    \centering
    \vspace*{-\baselineskip}%
    \hspace*{-0.05\linewidth}%
    \includegraphics[width=\linewidth]{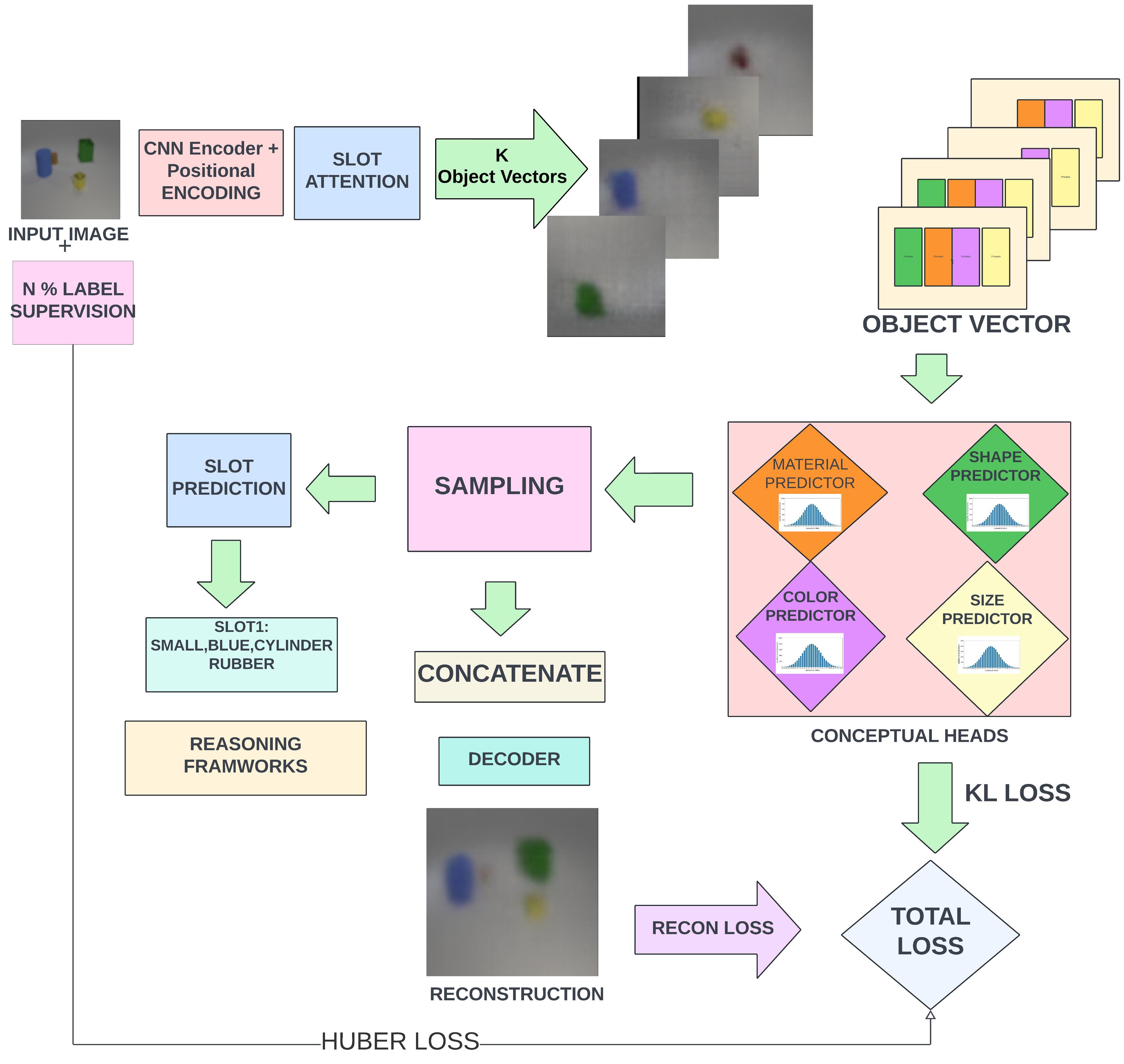}%
    \vspace*{-.5\baselineskip}%
    \caption{Architecture and training scheme of our two-stage neuro-symbolic visual reasoner, combining a VAE with slot attention and concept heads in the first stage.}
    \label{fig:Overall architecture}
    \vspace*{-.5\baselineskip}%
\end{figure}

\section{Introduction}
The dichotomy between connectionist and symbolic paradigms has defined much of the research landscape in artificial intelligence \cite{Minsky1991LogicalVA}. On one hand, deep learning models have achieved unprecedented success in perceptual tasks by extracting features from vast datasets \cite{goodfellow2016deep}. On the other, symbolic AI, particularly ILP~\cite{muggleton1991inductive}, provides a powerful framework for transparent, human-readable reasoning and robust generalization from structured knowledge. A central, unresolved challenge is the \textbf{symbol grounding problem} \cite{bengio2013representation}: how to autonomously map high-dimensional, unstructured sensory data, such as images, to a discrete symbolic and human-interpretable background knowledge that a reasoning engine can manipulate.
Fully unsupervised methods for learning disentangled representations, including end-to-end neuro-symbolic approaches \cite{kikaj2025deepgraphlog,garcez2023neurosymbolic}, lack inductive biases and often fail to produce representations that align with salient, human-interpretable concepts \cite{locatello2019challenging,kazhdan2021disentanglement}.
Full supervision of concepts instead relies on costly dense annotations, limiting scalability.

In this work, we navigate the spectrum between these extremes, proposing a neuro-symbolic framework guided by a minimal supervisory signal. Our hypothesis is that sparse labels are sufficient to guide a generative model toward learning a high-fidelity, object-centric symbolic background knowledge suitable for subsequent reasoning. Our approach is a two-stage system:
% TODO: remove enumration for saving space
\begin{enumerate*}[label=(\arabic*), itemjoin={;}, itemjoin*={; and~}]
    \item \textbf{A perception module}, a slot-based VAE \cite{locatello2020objectcentric} trained with as little as 1\% of concept labels learns to disentangle the core properties of objects in a scene (e.g., shape, color, size)
    \item \textbf{a symbolic reasoning engine:} The VAE's predictions are converted into logical predicates, which serve as background knowledge for a reasoning framework like the ILP system Popper \cite{cropper2020learningprogramslearningfailures} to induce complex, relational rules from few visual examples.
\end{enumerate*}
%
%This modular, two-stage design is a deliberate alternative to end-to-end differentiable neuro-symbolic systems. Such end-to-end models often face difficult optimization challenges when applied to high-dimensional perceptual data, especially under sparse supervision. % cite if you want to state this!
Our decoupling of perception and reasoning enhances flexibility and interpretability, allowing us to pair the perception module with the \textit{optimal} downstream reasoner for a given task. This is a critical advantage, as our findings show that ILP (Popper) is perfectly robust to perceptual noise for combinatorial rules, while Bayesian Networks are superior for probabilistic counting.

Our main \textbf{contributions} are:
\begin{itemize}
    \item A \textbf{weakly-supervised, object-centric VAE architecture} that demonstrates superior conceptual generalization from minimal supervision. We show in a real-world medical case study that with only 1\% of labels, our model's performance under domain shift significantly exceeds that of foundation models like DINOv2.
    
    \item A flexible, \textbf{two-stage neuro-symbolic reasoning pipeline} which allows to leverage that different reasoners %(ILP, Bayesian Networks, NS-CL) 
    specialize in different logical tasks, such as combinatorial versus numerical reasoning.

    \item An \textbf{extensive empirical analysis} of the perception-reasoning bottleneck: While predicate fidelity degrades with lower supervision, reasoners like ILP (Popper\cite{cropper2020learningprogramslearningfailures}) maintain robustness to perceptual noise for certain rule types.
    %We also identify the limits of this pipeline, showing that 
    For complex, multi-concept conjunctions we identify as remaining challenge and potential future work to accurately predict all required concepts simultaneously.
\end{itemize}
%%%%%%%%%%%%%%%%%%%%%%%%%%%%%%%%%%%%%%%%%%%%%%%%%%%%%%%%%%%%%%%%%%%%%%
% Related Work
%%%%%%%%%%%%%%%%%%%%%%%%%%%%%%%%%%%%%%%%%%%%%%%%%%%%%%%%%%%%%%%%%%%%%%
\section{Related Work}
% TODO: sec intro can be removed for sake of space
% Our work is situated at the intersection of object-centric representation learning, neuro-symbolic reasoning, and Inductive Logic Programming.

\subsection{Object-Centric Representation Learning} % TODO: use less space-consuming paragraph highlighting

The premise that scenes are compositions of objects is a powerful inductive bias. Unsupervised models like Slot Attention \cite{locatello2020objectcentric} have demonstrated a remarkable ability to segment scenes into object-centric \enquote{slots} through an iterative attention mechanism. The resulting representations are permutation-equivariant and provide a structured basis for downstream tasks. However, as shown by Locatello et al.~\cite{locatello2019challenging}, unsupervised learning of disentangled representations is fundamentally ill-posed without strong inductive biases. Methods like $\beta$-VAE \cite{higgins2016betavae} or FactorVAE~\cite{kim2018disentangling} encourage statistical independence in the latent space, but this offers no guarantee that the learned factors will align with semantically meaningful concepts. For instance, concepts that are meaningful to humans are often statistically \emph{dependent} (e.g., the concepts \textsf{apple} and \textsf{red} are correlated). %, whereas a purely statistical model may find it optimal to disentangle color from shape, mixing concepts like failing to capture the unified \textsf{red apple} concept. 

% The premise that scenes are compositions of objects is a powerful inductive bias. Unsupervised models like Slot Attention \cite{locatello2020objectcentric} have demonstrated a remarkable ability to segment scenes into object-centric "slots" through an iterative attention mechanism. The resulting representations are permutation-equivariant and provide a structured basis for downstream tasks. However, as shown by Locatello et al. \cite{locatello2019challenging}, unsupervised learning of disentangled representations is fundamentally ill-posed without inductive biases. While methods like $\beta$-VAE \cite{higgins2016betavae}, FactorVAE \cite{kim2018disentangling}, and $\beta$-TCVAE \cite{4809} encourage statistical independence in the latent space, they offer no guarantee that the learned factors will align with semantically meaningful concepts.
% TODO: illustrate this with an example / explain: concepts that are meaningful for humans are not necessarily statistically independent, like "apple" and "red".
Our choice of a VAE with Slot Attention is motivated by its proven efficacy in object discovery; by introducing a weak supervisory signal \cite{Li_2024_CVPR}, we explicitly guide the disentanglement process toward a desired conceptual background knowledge.

\subsection{Neuro-Symbolic Approaches}
Bridging the neural-symbolic gap is a central goal of modern AI. % TODO: cite
Approaches can be broadly categorized into three paradigms. % TODO: cite
First, \textbf{end-to-end differentiable systems} like DeepProblog~\cite{manhaeve2018deepproblogneuralprobabilisticlogic} % TODO: cite
integrate probabilistic logical reasoning directly into a neural network for joint optimization.
However, such tightly-coupled systems can be notoriously difficult to train, especially on complex perceptual data with sparse labels, due to the challenges of navigating a vast, combinatorial search space via gradient descent. % TODO: cite
Alternatively, \textbf{implicit neural reasoners} like the Neuro-Symbolic Concept Learner (NS-CL) \cite{mao2018neurosymbolic} or VQA transformers \cite{mamaghan2024exploring} also maintain a modular structure, but use neural networks for both perception and reasoning. These models perform reasoning \emph{implicitly}, learning a function that approximates a logical rule without producing an explicit, human-readable program. Our two-stage approach offers a pragmatic \textbf{modular, explicit-reasoning} alternative, decoupling perception (deep generative models) from reasoning (specialized symbolic solvers). This modularity provides two key advantages: (1) \textbf{Flexibility}, as the same generated predicates can feed diverse reasoners, allowing selection of the optimal tool; and (2) \textbf{Interpretability}, by providing a clear symbolic predicate checkpoint and an explicit, verifiable logical rule as output.

\section{Our Approach}

Our neuro-symbolic framework comprises two main stages: a perception module that learns concepts from weakly supervised visual data, and a predicate generation pipeline that translates these concepts into a symbolic vocabulary for a downstream reasoning engine as illustrated in (\cref{fig:Overall architecture}).

\subsection{Perception Module} % TODO: shorten heading

The perception module is a Variational Autoencoder integrating a \textbf{slot attention} mechanism \cite{locatello2020objectcentric}, a design chosen for its proven strengths in learning structured, object-centric representations. The VAE serves as a robust generative foundation, consistent with state-of-the-art approaches for learning disentangled latent spaces \cite{kim2018disentangling,liu2022learning}. While a standard VAE captures a global scene representation, the integrated Slot Attention module performs the critical task of object discovery. It uses an iterative, competitive attention mechanism to group perceptual features from a CNN encoder into a fixed number of object-centric feature vectors $\mathbf{z}_{\text{slot}}^{(i)}$, the \emph{slots}. Importantly, shared weights ensure that each slot is a general-purpose representational unit capable of binding to any object, rather than specializing to a particular object type. This eases reading out predicate values.

However, the purely unsupervised learning of semantically meaningful concepts is theoretically impossible without inductive biases \cite{seitzer2023bridginggaprealworldobjectcentric}. Therefore, we introduce a \textbf{weak supervisory signal}. This minimal supervision (from as little as 1\% of labels) acts as the crucial inductive bias needed to guide the VAE's latent space to align with the desired human-interpretable concepts (shape, color, etc.). This ensures that the generated predicates are not only consistent but also semantically grounded and interpretable. %, making them reliable for downstream symbolic reasoning engines like Popper.

\begin{figure*}[tb]
    \centering
    \newlength{\myfigwidth}\setlength{\myfigwidth}{.28\linewidth}
    % Subfigure for Coordinates
    \vspace*{-.5\baselineskip}%
    \hspace*{-.15\myfigwidth}%
    \begin{subfigure}[t]{\myfigwidth}
        \strut\\[-1.5\baselineskip]%
        \includegraphics[width=\linewidth]{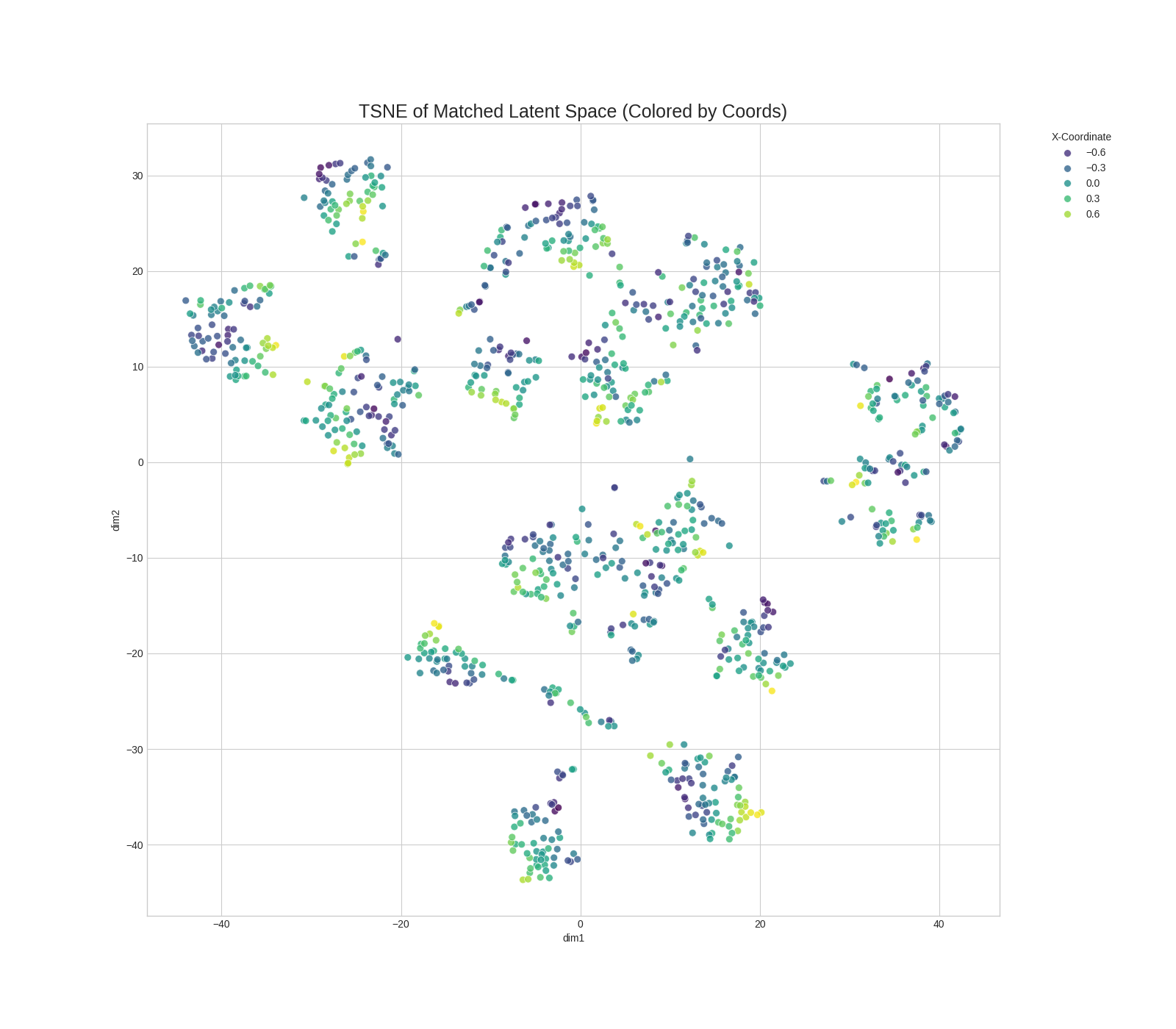}%
        \vspace*{-.8\baselineskip}%
        \caption{by coords}
        \label{fig:latent_coords}
        \vspace*{-\baselineskip}%
    \end{subfigure}%
    \hspace*{-0.05\myfigwidth}%
    %\hfill% Adds space between figures
    % Subfigure for Color
    \begin{subfigure}[t]{\myfigwidth}
        \strut\\[-1.5\baselineskip]%
        \includegraphics[width=\linewidth]{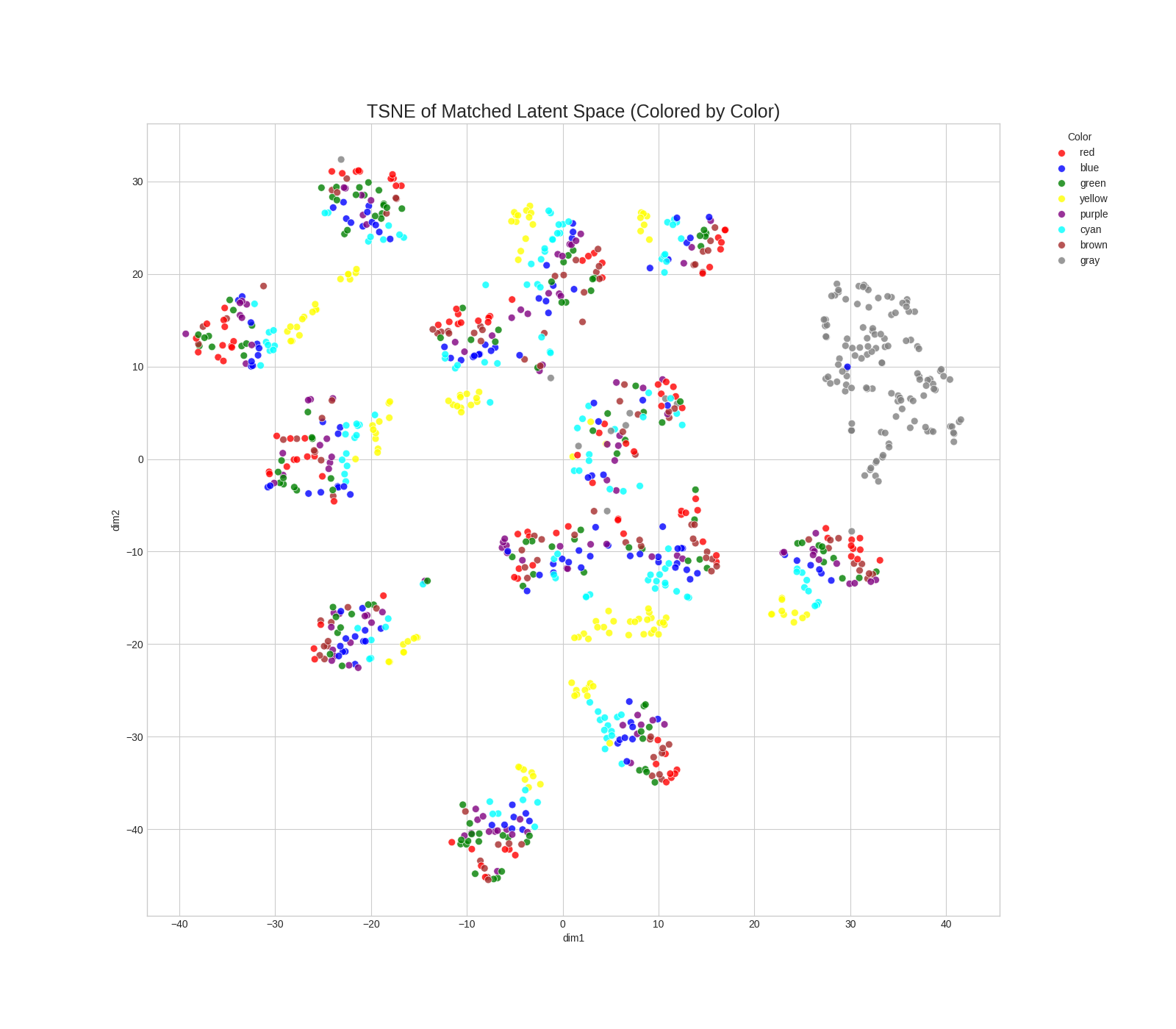}%
        \vspace*{-.8\baselineskip}%
        \caption{by color}
        \label{fig:latent_color}
        \vspace*{-\baselineskip}%
    \end{subfigure}%
    %\\%
    % % Subfigure for Material
    % \begin{subfigure}{0.15\textwidth}
    %     \includegraphics[width=\linewidth]{Latent_plots/latent_by_material_epoch_150 (1).png}
    %     \caption{by Material}
    %     \label{fig:latent_material}
    % \end{subfigure}
    % \hfill
    % Subfigure for Shape
    \hspace*{-0.05\myfigwidth}%
    \begin{subfigure}[t]{\myfigwidth}
        \strut\\[-1.5\baselineskip]%
        \includegraphics[width=\linewidth]{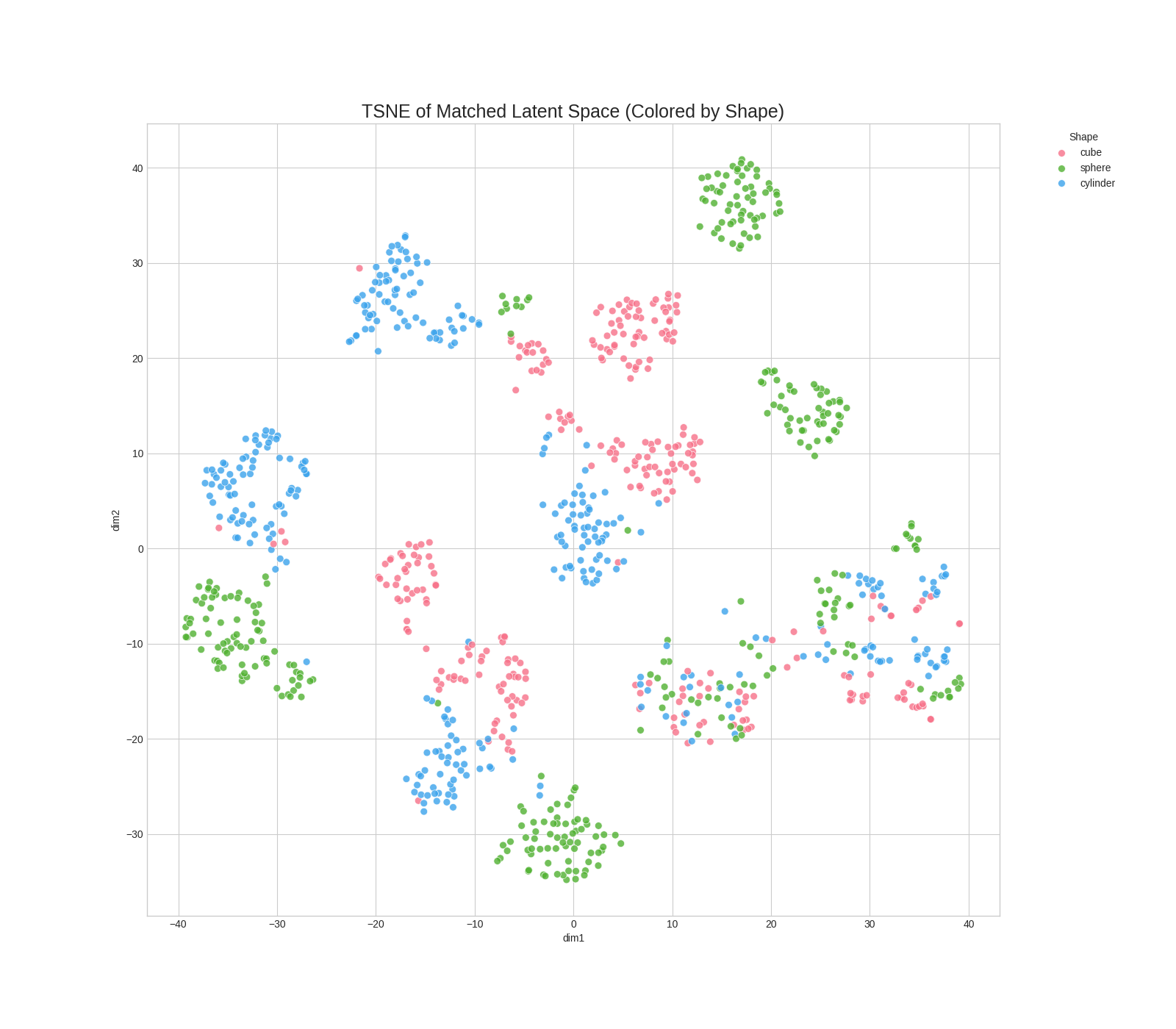}%
        \vspace*{-\baselineskip}%
        \caption{by shape}%
        \label{fig:latent_shape}%
        \vspace*{-\baselineskip}%
    \end{subfigure}%
    %\hfill%
    % % Subfigure for Size
    % \begin{subfigure}{0.15\textwidth}
    %     \includegraphics[width=\linewidth]{Latent_plots/latent_by_size_epoch_150__1_.png}
    %     \caption{by Size}
    %     \label{fig:latent_size}
    % \end{subfigure}
    \parbox[t]{\linewidth-2.85\myfigwidth}{%
    \caption{UMAP visualizations of the latent space, colored by different ground-truth concepts.}
    \label{fig:all_latent_plots}
    }
\end{figure*}

\paragraph{Structured Latent Space.} A key architectural choice is the explicit partitioning of the latent space for each slot. The latent representation $\mathbf{z}_{\text{slot}}$ is a concatenation of sub-vectors, each dedicated to a specific conceptual property:
\begin{equation}
    \mathbf{z}_{\text{slot}} = [\mathbf{z}_{\text{coord}} \mathbin{\|} \mathbf{z}_{\text{presence}} \mathbin{\|} \mathbf{z}_{\text{shape}} \mathbin{\|} \mathbf{z}_{\text{color}} \mathbin{\|} \mathbf{z}_{\text{size}} \mathbin{\|} \mathbf{z}_{\text{material}}]
\end{equation}
This structure is enforced by using separate \emph{conceptual heads}—implemented as 2-layer MLPs \cite{mamaghan2024exploring}—that each project the slot's feature representation onto the mean ($\mu$) and log-variance ($\log\sigma^2$) of a Gaussian distribution (cf.\ VAEs \cite{higgins2016betavae}). %This design creates a continuous, probabilistic latent space for all concepts.
Categorical values (e.g., of \textsf{shape}) are subsequently derived by separate predictor networks fed by these sampled latent sub-vectors.
Notably we add the special heads for coordinate (coord) and presence prediction.
This partitioning of $\mathbf{z}_{\text{slot}}$ encourages to encode distinct semantic properties into non-overlapping parts of the latent code as seen in (\cref{fig:all_latent_plots}).

% TODO: GEsina: mention that VAE / disentangled encodings are helpful/used for abstract visual reasoning \cite{vansteenkiste2019are,steenbrugge2018improving}

\paragraph{Weakly Supervised Training Objective.} For training we optimize a composite loss function $\mathcal{L}_{\text{total}}$. Crucially, the supervisory component of the loss ($\mathcal{L}_{\text{sup}}, \mathcal{L}_{\text{coord}}, \mathcal{L}_{\text{presence}}$) is applied only to a small fraction of the training data:
\begin{equation}
    \mathcal{L}_{\text{total}} = \mathcal{L}_{\text{recon}} + \beta\mathcal{L}_{\text{KL}} + \delta\mathcal{L}_{\text{sup}} + \lambda\mathcal{L}_{\text{coord}} + \gamma\mathcal{L}_{\text{presence}}
\end{equation}
where $\mathcal{L}_{\text{recon}}$ is a pixel-wise mean squared error (MSE) reconstruction loss; $\mathcal{L}_{\text{KL}}$ is the VAE Kullback-Leibler divergence term ($\beta$ increased via a warmup schedule \cite{higgins2016betavae});
$\mathcal{L}_{\text{coord}}$ is a MSE on predicted coordinates; $\mathcal{L}_{\text{presence}}$ is a binary cross-entropy (BCE) loss on the dedicated presence predictor head;
and $\mathcal{L}_{\text{sup}}$ is the supervised concept (Huber) loss on the remaining heads, calculated on pairs of slot$\leftrightarrow$ground-truth object matched via the \textbf{Hungarian algorithm} \cite{Kuhn1955TheHM}.

\subsection{Automated Predicate Generation Pipeline}
During inference, the VAE is used to translate a given input image into symbols by executing the following pipeline:
\begin{enumerate}
    \item \textbf{Inference and Filtering:} The model's forward pass yields $k$ slot representations with property and presence predictions. Slots with a presence probability below a threshold (here used: 0.5) are discarded.
    \item \textbf{Symbol Grounding:} For each active slot (e.g., \texttt{obj\_i}), predicted logits are converted to symbolic atoms. The argmax of categorical logits determines the predicate (e.g., \texttt{color(obj\_i, red)}).
    \item \textbf{Background Knowledge Construction:} Repeating this for all active slots and concepts generates a complete symbolic scene description (first-order logic predicates), forming the background knowledge for the reasoning.
\end{enumerate}

%%%%%%%%%%%%%%%%%%%%%%%%%%%%%%%%%%%%%%%%%%%%%%%%%%%%%%%%%%%%%%%%%%%%%%
% Experiments
%%%%%%%%%%%%%%%%%%%%%%%%%%%%%%%%%%%%%%%%%%%%%%%%%%%%%%%%%%%%%%%%%%%%%%
\section{Experiments and Results}

Our empirical evaluation is designed to systematically investigate the weak supervision (RQ1), robustness (RQ2, RQ4), and reasoning capabilities (RQ3) of our neuro-symbolic architecture across four research questions.

\noindent\textbf{Datasets:} We use the CLEVR\cite{johnson2016clevrdiagnosticdatasetcompositional}, CLEVR-Tex~\cite{karazija2021clevrtextexturerichbenchmarkunsupervised}, a 2D version\footnote{\label{see-supplementary}further details available in the supplementary material} of CLEVR, dSprites~\cite{dsprites17}, 3D-Shapes~\cite{3dshapes18} datasets, with a case study on a real-world medical imaging task using HAM~\cite{Tschandl2018_HAM10000} and Melanoma~\cite{muhammad_hasnain_javid_2022} datasets.

\begin{figure}
    \centering
    \hspace*{-.05\linewidth}%
    \includegraphics[width=1.1\linewidth]{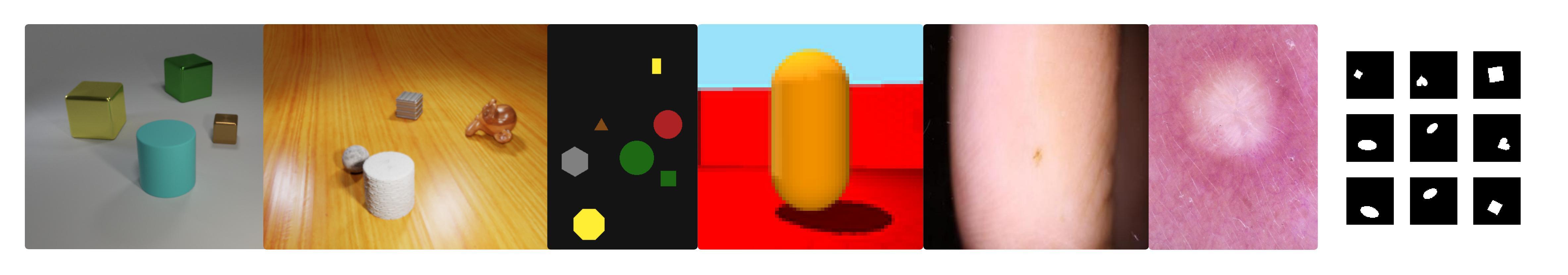}
    \caption{\textit{Left to right:} Samples from the used datasets Clevr, Clevr-Tex, 2D version of Clevr, 3D shapes, melanoma , HAM10000, Dsprites}
    \label{fig:Datasets used in the experiment section}
\end{figure}

\paragraph{Reproducibility Details:}The \textbf{SlotVAE} is trained for 200 epochs (batch size 64) using the Adam optimizer with a learning rate of $4 \times 10^{-4}$. The architecture features a 5-layer CNN Encoder and a 4-layer Decoder. The Slot attention is configured with 10 slots, 3 attention iterations, and a hidden layer of 64.Key loss function weights are $\beta_{\text{final}}=0.0001$, $\delta=10.0$, $\gamma=5.0$, and $\lambda_{\text{coord}}=5.0$

\noindent\textbf{SlotVAE Model:}
For the CLEVR-like datasets we set the number of slots to $k=10$ (maximum number of objects), and to $k=2$ else (single object + background).
For the CLEVR-like, dSprites and 3D-Shapes datasets we trained heads for coords, presence, and the symbolic properties \textsf{shape}, \textsf{color}, \textsf{size}, \textsf{material}.
In case of the skin lesion benchmarks HAM and Melanoma, we focus on three concepts for predicate quality and rule formulation: \textsf{dx}, the seven diagnostic classes (3 malignant: \texttt{mel}, \texttt{bcc}, \texttt{akiec}; 4 benign: \texttt{nv}, \texttt{bkl}, \texttt{vasc}, \texttt{df}); the \textsf{age}; and \textsf{localization}, which refers to the body part where the lesion exist.
%
%During inference, the image is translated into slots, which are filtered and turned into a structured, symbolic scene description for downstream reasoning (e.g., \texttt{cube(obj1)}, \texttt{red(obj1)}). % known from the approach section

\noindent\textbf{Downstream Reasoning Frameworks:}
To ensure a fair and highly controlled comparison, the VAE-generated predicates and rule labels were formatted for four distinct reasoning frameworks to learn rules: \textbf{ILP (Popper \cite{cropper2020learningprogramslearningfailures})}, \textbf{Decision Tree (DT)}, \textbf{Bayesian Network (BN)}, and \textbf{Neural-Symbolic (NS-CL)} \cite{mao2018neurosymbolic}.
For all training sets (e.g., 500 or 5000 images), we maintained a consistent and challenging 4:1 ratio of negative to positive examples. The rule learning involves training each framework on the noisy, VAE-generated predicates and comparing its performance to a baseline trained on perfect, ground-truth predicates.

% TODO: overview over research questions
% TODO: what is the cutoff threshold for the presence prediction? Also 0.5? Then put here or state more clearly above.
% TODO: What is the (default) number of slots? How large is the architecture?

\subsection{RQ1: Effects of Weakening Supervision}\label{rq1}
We here investigate the effect of reduced supervision on predicate quality and reasoning.
%
%\paragraph{Setup.}
For that we considered five levels of supervision on CLEVR, varying the percentage of labeled data: 1\%, 5\%, 15\%, 25\%, 50\% and 75\%. On a dataset of 70,000 images, 1\% supervision corresponds to just 700 labeled examples.
For each level we trained our SlotVAE model 5 times.

\paragraph{\textbf{Results on Predicate Quality}.} As expected and shown in \cref{tab:concept_accuracy}, predicate quality consistently scales with the level of supervision. With 100\% supervision, the model achieves near-perfect accuracy on most concepts. Notably, even with only 1\% of labels, the model learns concepts like size ($88.08\pm2.92\%$) and shape ($71.40\pm2.18\%$) with high fidelity. However, concepts with high perceptual variance, such as color, prove more challenging; color accuracy at 1\% supervision is only $21.8\pm2.41\%$. This highlights a critical finding: \emph{The data-efficiency of learning is concept-dependent. Structural properties are learned readily, while surface properties are more data-hungry.}

\begin{table}[tb]
\centering
\caption{Concept accuracy (\%) and coordinate MAE vs.\ concept supervision level (\%). Values are (mean$\pm$std.) over 5 runs.}
\label{tab:concept_accuracy}
\scriptsize % Use a smaller font
\resizebox{\columnwidth}{!}{%
\begin{tabular}{@{}lcccccc@{}}
\toprule
\textbf{Superv.} & \textbf{Shape} & \textbf{Color} & \textbf{Size} & \textbf{Material} & \textbf{Presence} & \textbf{Coord.} \\
%\textbf{Level (\%)} & \textbf{(\%)} & \textbf{(\%)} & \textbf{(\%)} & \textbf{(\%)} & \textbf{(\%)} & \\
\midrule
1   & 71.40 $\pm$ 2.18 & 21.82 $\pm$ 2.41 & 88.08 $\pm$ 2.92 & 83.58 $\pm$ 4.00 & 65.86 $\pm$ 2.38 & 0.2646 \\
5   & 80.79 $\pm$ 3.41 & 44.46 $\pm$ 2.84 & 96.14 $\pm$ 2.31 & 90.42 $\pm$ 3.18 & 78.49 $\pm$ 2.89 & 0.1603 \\
15  & 84.14 $\pm$ 1.68 & 78.72 $\pm$ 3.03 & 97.41 $\pm$ 0.54 & 94.14 $\pm$ 0.70 & 83.84 $\pm$ 2.42 & 0.0989 \\
25  & 84.07 $\pm$ 2.89 & 82.14 $\pm$ 4.61 & 97.51 $\pm$ 0.62 & 93.88 $\pm$ 1.30 & 84.79 $\pm$ 2.17 & 0.0873 \\
50  & 86.44 $\pm$ 2.74 & 88.21 $\pm$ 3.31 & 97.80 $\pm$ 0.62 & 95.40 $\pm$ 1.42 & 88.65 $\pm$ 1.24 & 0.0772 \\
75  & 89.44 $\pm$ 2.35 & 89.21 $\pm$ 2.41 & 97.95 $\pm$ 0.54 & 95.90 $\pm$ 1.22 & 90.65 $\pm$ 1.14 & 0.0572 \\
100 & 93.10 $\pm$ 1.79 & 91.38 $\pm$ 2.67 & 98.80 $\pm$ 0.45 & 96.76 $\pm$ 1.18 & 93.80 $\pm$ 1.05 & 0.0425 \\
\bottomrule
\end{tabular}
}
\end{table}

\paragraph{Results on Reasoning Tractability.}
Our experiment illustrates\footnote{Find the full results in the supplementary material.} how predicate fidelity directly impacts the tractability of the downstream ILP solver (Popper). We tested this on three rules of increasing difficulty:
(1) An \textbf{Easy} rule (``Contains a yellow sphere''),
(2) a \textbf{Medium} conjunctive rule (``Large blue sphere \& small yellow sphere''), and
(3) a \textbf{Hard} relational rule (``Blue sphere in front of yellow cube'').
For the \textbf{Easy} rule, Popper succeeded even with the 1\% supervision model, as the high-fidelity shape predicate (71.4\%) provided a sufficient signal despite poor color accuracy (21.8\%).
For the \textbf{Medium} rule, the 1\% model failed (timeout); the task only became tractable with $\geq 25\%$ supervision, which provided reliable color predicates (82.1\%).
The \textbf{Hard} rule, which requires high-fidelity color and spatial predicates, failed at both 1\% and 25\% supervision. It only became solvable (at 75\% supervision) once \textit{all} required predicates (color acc: 89.4\%, coord. MAE: 0.057) were sufficiently accurate. This demonstrates a clear trade-off: \emph{While minimal supervision suffices for simple rules, complex relational rules require a higher-fidelity vocabulary that is only achievable with increased supervision.}

% This demonstrates a clear trade-off: while minimal supervision is sufficient for grounding simple concepts, inducing complex relational rules requires a higher-fidelity symbolic BK that can only be achieved with increased supervision.

\paragraph{Ablation: Impact of Number of Slots.}
We performed an ablation study to determine the impact of the number of slots ($k$) on predicate quality and its interaction with weak supervision.
For the HAM skin-lesion dataset we hypothesized that $k=2$ (lesion, background) would provide the strongest inductive bias for the HAM dataset, given its single-lesion structure. %We trained models with $k \in \{1, 2, 5, 10\}$ at supervision levels from 1\% to 75\%, evaluating \texttt{dx} (diagnostic category) and localization (body part) accuracy, and age mean absolute error (MAE).
Results (\cref{tab:slot_comparison}) show the inductive bias is most critical when supervision is minimal. At \textbf{1\% supervision}, the $k=1$ (global) model struggled (49.5\% dx Acc.). Setting $k=2$ provided the minimal lesion/background partition, yielding a \textbf{+18.2\% absolute improvement} (67.7\% dx Acc.). Higher $k$ values ($k=5, 10$) degraded performance by segmenting the single lesion into semantically meaningless regions. While $k=1$ became the top performer at $\geq$15\% supervision (90.0\% at 75\%), in the single-object case $k=2$ remained the most robust, data-efficient configuration.
For the multi-object case we found similar results, where minimal supervision heavily benefits from choosing the maximum number of objects as slot number (10; find details in the supplementary).
This demonstrates that \emph{providing the correct structural bias is most critical when the supervisory signal is weakest}.

\begin{table}[tb]
\caption{Effect of slot number ($k$) on concept learning in SlotVAE, as mean$\pm$std for concept accuracy (\%) and age MAE.}
\label{tab:slot_comparison}
\centering % Kept for good practice
\resizebox{\columnwidth}{!}{%
\scriptsize % Use scriptsize to make the font smaller
\begin{tabular}{@{}c c ccc@{}} % Simplified column definitions
\toprule
\textbf{Superv. (\%)} & \textbf{Slots ($k$)} & \textbf{dx Acc. (\%)} & \textbf{loc Acc. (\%)} & \textbf{Age MAE (Years)} \\
\midrule
\multirow{4}{*}{\textbf{1}} 
& 1 & 49.50 $\pm$ 2.21 & 20.00 $\pm$ 4.69 & 0.599 $\pm$ 0.162 \\
& 2 & \textbf{67.67 $\pm$ 2.08} & \textbf{21.00 $\pm$ 6.08} & 0.784 $\pm$ 0.207 \\
& 5 & 66.00 $\pm$ 2.00 & 19.80 $\pm$ 6.10 & 0.846 $\pm$ 0.144 \\
& 10 & 64.00 $\pm$ 4.58 & 18.00 $\pm$ 8.66 & 1.037 $\pm$ 0.101 \\
\midrule
\multirow{4}{*}{\textbf{15}} 
& 1 & \textbf{75.00 $\pm$ 2.59} & \textbf{27.00 $\pm$ 9.90} & 0.628 $\pm$ 0.039 \\
& 2 & 70.67 $\pm$ 3.06 & 25.00 $\pm$ 5.57 & 0.691 $\pm$ 0.148 \\
& 5 & 73.00 $\pm$ 3.00 & 20.60 $\pm$ 3.78 & 0.699 $\pm$ 0.108 \\
& 10 & 68.25 $\pm$ 6.02 & 19.00 $\pm$ 2.94 & 0.710 $\pm$ 0.091 \\
\midrule
\multirow{4}{*}{\textbf{25}} 
& 1 & \textbf{77.33 $\pm$ 2.31} & \textbf{34.33 $\pm$ 3.21} & 0.756 $\pm$ 0.060 \\
& 2 & 77.00 $\pm$ 5.66 & 30.50 $\pm$ 3.54 & 0.716 $\pm$ 0.181 \\
& 5 & 71.67 $\pm$ 3.14 & 25.67 $\pm$ 6.44 & 0.699 $\pm$ 0.145 \\
& 10 & 69.50 $\pm$ 3.70 & 20.50 $\pm$ 2.38 & 0.708 $\pm$ 0.049 \\
\midrule
\multirow{4}{*}{\textbf{50}} 
& 1 & \textbf{86.00 $\pm$ 3.84} & \textbf{32.00 $\pm$ 2.45} & 0.723 $\pm$ 0.120 \\
& 2 & 74.00 $\pm$ 3.48 & 29.00 $\pm$ 2.45 & 0.603 $\pm$ 0.115 \\
& 5 & 78.33 $\pm$ 5.68 & 28.50 $\pm$ 5.68 & 0.728 $\pm$ 0.104 \\
& 10 & 70.75 $\pm$ 2.75 & 19.25 $\pm$ 2.36 & 0.718 $\pm$ 0.146 \\
\midrule
\multirow{4}{*}{\textbf{75}} 
& 1 & \textbf{90.00 $\pm$ 2.10} & \textbf{45.50 $\pm$ 9.19} & 0.717 $\pm$ 0.015 \\
& 2 & 76.50 $\pm$ 12.02 & 29.00 $\pm$ 8.49 & 1.132 $\pm$ 0.488 \\
& 5 & 83.80 $\pm$ 7.56 & 35.00 $\pm$ 11.11 & 0.743 $\pm$ 0.070 \\
& 10 & 73.25 $\pm$ 1.71 & 23.75 $\pm$ 2.50 & 0.726 $\pm$ 0.056 \\
\bottomrule
\end{tabular}%
}
\end{table}

\subsection{RQ2: Assessing Robustness to Domain Shift}

This experiment assesses the robustness of the learned conceptual representations under different visual domain shifts. We expect that---while zero-shot performance is limited---structural features (e.g., shape, size) are transferable via fine-tuning, and that adaptation success is asymmetric, favoring transfer from complex to simple visual domains.
%
%\subsection*{Methodology}
%
To evaluate this we use a three-stage protocol: (1) zero-shot transfer, (2) linear probing of frozen features, and (3) full fine-tuning. This protocol was applied across three distinct domain shifts: dSprites$\leftrightarrow$3DShapes, CLEVR$\leftrightarrow$CLEVR-Tex, and CLEVR$\leftrightarrow$2D CLEVR.\footnoteref{see-supplementary}

%\subsubsection*{Results and Analysis}
\paragraph{\textit{CLEVR$\leftrightarrow$2D CLEVR}.}
This shift, which removes 3D geometric cues (e.g., shading), demonstrated a critical adaptation asymmetry. The 3D$\to$2D transfer resulted in complete model collapse, as the encoder was over-specialized to 3D features that were absent post-transfer. In contrast, the reverse 2D$\to$3D transfer showed clear adaptation success. While linear probing on the frozen 2D features was ineffective, full fine-tuning allowed the model to learn robust representations. Latent space analysis confirms that the model successfully learned to disentangle foundational structural concepts, with UMAP plots\footnoteref{see-supplementary} showing clear, well-separated clusters for both \textsf{size} and \textsf{shape}. The \textsf{color} concept, while achieving reasonable classification accuracy, remained more challenging to separate, showing partially entangled clusters in its latent space. This indicates that \emph{features for structural concepts learned in an information-poor (2D) domain are transferable and can be successfully adapted to a more complex (3D) domain}.

\paragraph{\textit{dSprites$\leftrightarrow$3DShapes:} 2D-to-3D Generalization.}
We assessed the generalization of geometric concepts across dimensions by transferring between dSprites (2D) and 3DShapes (3D). While zero-shot transfer failed, full fine-tuning yielded effective bidirectional adaptation. The dSprites$\to$3DShapes transfer achieved $\sim$99\% shape accuracy, indicating the learned 2D geometric features were sufficiently abstract to represent simple 3D objects. The reverse transfer also produced a well-disentangled latent space for shape\footnote{Find respective plots in the supplementary material.}, confirming that \emph{representations for foundational concepts like geometry are robust to dimensionality and rendering style changes}.

\paragraph{\textit{CLEVR$\leftrightarrow$CLEVR-Tex:} Robustness to Surface Appearance.}
This shift revealed a critical adaptation asymmetry. The simple-to-complex transfer (CLEVR$\to$CLEVR-Tex) exhibited a dichotomy in concept robustness: structural concepts (shape: 91\%; size: 98.5\%) adapted post-finetuning, whereas surface-level concepts like color failed (22\% accuracy). Conversely, the complex-to-simple transfer (CLEVR-Tex$\to$CLEVR) was highly effective, demonstrating robust generalization (90\%) across all concepts, including color. This finding implies that exposure to visually diverse, textured data is a necessary inductive bias for learning fundamental and robust representations of surface-level properties.

\paragraph{Summary:}
\emph{Perceptual robustness is concept- and direction-dependent}, generally hindering zero-shot generalization. Post-finetuning, structural features proved highly transferable:
%we observed effective bidirectional adaptation for dSprites$\leftrightarrow$3DShapes (simple geometry) and successful 2D$\to$3D CLEVR adaptation (information-poor to rich), with latent analysis confirming disentanglement of size and shape.
We observed successful 2D$\to$3D adaptation in two different settings, with latent analysis confirming disentanglement of size and shape.
%A critical asymmetry emerged in the CLEVR$\leftrightarrow$CLEVR-Tex 
Regarding shift of surface appearance, complex-to-simple (textured$\to$uniform) transfer was robust, but simple-to-complex (uniform$\to$textured) transfer exhibited a conceptual dichotomy, where structural concepts adapted but color accuracy collapsed to 22\%.
Finally, the destructive 3D$\to$2D (information-loss) shift proved intractable. Thus, while structural features are highly transferable, robust surface-feature adaptation is asymmetrically dependent on the source domain's visual complexity providing a sufficient inductive bias.

\subsection{RQ3: Downstream Logical Reasoning}

To evaluate the robustness of downstream logical reasoning, we designed a multi-stage pipeline. The perception front-end consists of a single, pre-trained SlotVAE. Based on our analysis in \hyperlink{rq1}{RQ1} (see \cref{tab:concept_accuracy}), we deliberately selected the model checkpoint trained with only \textbf{15\% label supervision}. This choice represents a practical trade-off, providing high accuracy on structural concepts (e.g., 97.4\% size) and sufficient fidelity on challenging ones (e.g., 78.7\% color). This single, imperfect model was \textbf{frozen} to serve as a realistic predicate generator for all subsequent reasoning experiments.

We defined five logical rules of increasing complexity on the synthetic multi-object datasets, from \textbf{Simple Existential} (``Is there a large, red sphere?'') to \textbf{Conjunctive} (``...a blue sphere AND a yellow cube?''), \textbf{Disjunctive} (``...a green metal cylinder OR a yellow rubber cube?''), \textbf{Cardinality} (``...exactly two metal objects?''), and \textbf{Universal Quantifier} (``Are ALL spheres blue?''). Ground-truth labels for these rules were generated from the dataset's scene graphs.
We then analyzed the performance of the four reasoning frameworks regarding their ability to induce accurate logical rules from the 15\%-supervised VAE's predicates. The analysis reveals distinct specializations and a striking robustness in the ILP framework.

\paragraph{ILP's Superior Robustness on Combinatorial Rules.}
A standout result is the remarkable resilience of the ILP (Popper) framework to perceptual noise. As evidenced in \cref{tab:gap_5000}, for the Simple Existential, Conjunctive, and Disjunctive rules, Popper achieves a perfect F1-score of 1.0. This \textbf{zero performance gap} indicates that for these fundamental first-order logic structures, Popper's search mechanism is completely resilient to the perceptual noise, successfully identifying the correct symbolic hypothesis. In contrast, all other frameworks (BN, DT, NS-CL) struggled to find a meaningful signal for these rules, performing poorly on both the VAE and Ground Truth predicate sets.

\paragraph{Framework Specialization on Rule Types.}
Conversely, ILP's strict logical formalism caused it to fail on non-combinatorial tasks (\cref{tab:gap_5000}), where other frameworks demonstrated clear specializations. The \textbf{Decision Tree} was most effective for the \textbf{Cardinality} (counting) rule (0.571 F1), while the \textbf{Bayesian Network} (0.760 F1) and \textbf{NS-CL} (0.754 F1) excelled at the \textbf{Universal Quantifier} rule, indicating probabilistic methods are better suited for generalized reasoning on noisy data. The choice of reasoner is therefore a critical determinant of the pipeline's success. Despite using an imperfect perception module (15\% supervision, 78.7\% color acc.) and a 4:1 class imbalance, the generated predicates were of sufficient fidelity for successful—and in ILP's case, perfect—rule induction. This success, along with the ability of models like NS-CL to overcome predicate noise by scaling with data (e.g., improving from 0.620 to 0.754 F1 on the Universal rule), \emph{validates our modular, two-stage design, which allows for selecting the optimal reasoning engine for the problem's logical structure}.

\begin{table}[tb]
\centering
\caption{Performance on \textbf{5000 Images} (15\% VAE Supervision). Values are F1-Score (Mean $\pm$ Std. Dev.) over 5 runs.}
\label{tab:gap_5000}
\scriptsize
\resizebox{\columnwidth}{!}{%
\begin{tabular}{@{}llcc@{}}
\toprule
\textbf{Rule} & \textbf{Framework} & \textbf{GT F1 (mean)} & \textbf{VAE F1 (mean$\pm$std)} \\
\midrule
\multirow{4}{*}{\begin{tabular}[c]{@{}l@{}}"Exactly two \\ metal objects?"\end{tabular}} & Bayesian Network (BN) & 0.800 & 0.366 $\pm$ 0.250 \\
& \textbf{Decision Tree (DT)} &\textbf{ 1.000} & \textbf{0.571} $\pm$ \textbf{0.010} \\
& NS-CL & 0.516 & 0.221 $\pm$ 0.151 \\
& ILP (Popper) & 0.000 & 0.000 $\pm$ 0.000 \\
\midrule
\multirow{4}{*}{\begin{tabular}[c]{@{}l@{}}"Blue sphere AND \\ yellow cube?"\end{tabular}} & Bayesian Network (BN) & 0.036 & 0.010 $\pm$ 0.011 \\
& Decision Tree (DT) & 0.247 & 0.186 $\pm$ 0.023 \\
& NS-CL & 0.034 & 0.009 $\pm$ 0.017 \\
& \textbf{ILP (Popper)} & \textbf{1.000} & \textbf{1.000 $\pm$ 0.000} \\
\midrule
\multirow{4}{*}{\begin{tabular}[c]{@{}l@{}}"Green metal cylinder \\ OR yellow rubber cube?"\end{tabular}} & Bayesian Network (BN) & 0.960 & 0.940 $\pm$ 0.004 \\
& Decision Tree (DT) & 0.927 & 0.914 $\pm$ 0.005 \\
& NS-CL & 0.950 & 0.945 $\pm$ 0.005 \\
& \textbf{ILP (Popper)} & \textbf{1.000} & \textbf{1.000 $\pm$ 0.000} \\
\midrule
\multirow{4}{*}{\begin{tabular}[c]{@{}l@{}}"Large, red \\ sphere?"\end{tabular}} & Bayesian Network (BN) & 0.147 & 0.119 $\pm$ 0.102 \\
& Decision Tree (DT) & 0.377 & 0.338 $\pm$ 0.023 \\
& NS-CL & 0.448 & 0.390 $\pm$ 0.031 \\
& \textbf{ILP (Popper)} & \textbf{1.000} & \textbf{1.000 $\pm$ 0.000} \\
\midrule
\multirow{4}{*}{\begin{tabular}[c]{@{}l@{}}"Are ALL \\ spheres blue?"\end{tabular}} & Bayesian Network (BN) & 0.861 & 0.760 $\pm$ 0.020 \\
& Decision Tree (DT) & 0.835 & 0.645 $\pm$ 0.019 \\
& \textbf{NS-CL} & \textbf{0.871} & \textbf{0.754 $\pm$ 0.019} \\
& ILP (Popper) & 0.000 & 0.000 $\pm$ 0.000 \\
\bottomrule
\end{tabular}
}
\end{table}

% \begin{table}[tb]
% \centering
% \caption{Framework Strengths by Rule Type (F1-Score on VAE @ \textbf{5000 Images})}
% \label{tab:strengths_5000_final}
% \scriptsize
% \begin{tabular}{@{}lccccc@{}}
% \toprule
% \textbf{Rule} & \textbf{BN} & \textbf{DT} & \textbf{NS-CL} & \textbf{ILP} & \textbf{Excelling Framework} \\
% \midrule
% Cardinality & \textbf{0.723} & 0.661 & 0.640 & 0.000 & \textbf{Bayesian Network} \\
% Conjunctive & 0.000 & 0.182 & 0.721 & \textbf{1.000} & \textbf{ILP (Popper)} \\
% Disjunctive & 0.938 & 0.936 & 0.966 & \textbf{1.000} & \textbf{ILP (Popper)} \\
% Simple\_Exist & 0.149 & 0.369 & 0.888 & \textbf{1.000} & \textbf{ILP (Popper)} \\
% Univ. Quant & 0.749 & 0.764 & \textbf{0.897} & 0.000 & \textbf{NS-CL} \\
% \bottomrule
% \end{tabular}
% \end{table}

\subsection{RQ4: Baseline Comparison}% on Real World Domain Shift}
Finally we test the framework's data efficiency and real-world applicability in a high-stakes domain shift case study and baseline comparison. The perception modules, trained on the in-domain HAM skin lesion benchmark with sparse labels, are evaluated regarding their ability to generalize zero-shot to the Melanoma dataset for \textbf{malignant/benign classification.}
%
% Our experiments focus on three concepts for predicate quality and rule formulation: \textbf{dx}, the seven diagnostic classes (3 malignant: \texttt{mel}, \texttt{bcc}, \texttt{akiec}; 4 benign: \texttt{nv}, \texttt{bkl}, \texttt{vasc}, \texttt{df}), and \textbf{localization}, which refers to the body part where the lesion exist.
This is measured via \textbf{in-domain accuracy} (on HAM) versus \textbf{zero-shot accuracy} (on Melanoma), additionally computing the harmonic mean (H-mean) for a balanced assessment of specialization versus generalization.

\paragraph{\textbf{Baselines}.}
We compare our weakly-supervised \textbf{SlotVAE} (an object-centric model) against three state-of-the-art baselines representing different architectural paradigms:
\begin{itemize}
    \item \textbf{DINOv2} \cite{oquab2024dinov2learningrobustvisual}: A state-of-the-art, \textbf{fixed-region} foundation model pretrained on a huge amount of data and based on a Vision Transformer.
    \item \textbf{ResNet+SA} \cite{biza2023invariantslotattentionobject}: A modern \textbf{object-centric} baseline combining a ResNet backbone with Slot Attention.
    \item \textbf{GlobalVAE} \cite{watters2019spatialbroadcastdecodersimple}: A standard VAE with a spatial broadcast decoder, i.e., a \textbf{global} representation baseline.
\end{itemize}
We each train models for the HAM supervision levels 1\%, 15\%, 25\%, 50\%, 75\%, resulting in 35 compared models.

\noindent\textbf{Downstream Reasoning.}
In this setting, for 1000 test images, each model generated predicates \texttt{vae\_dx}, \texttt{vae\_loc}, \texttt{vae\_age}), which were compared to a ground truth set. These sets were used to train three downstream reasoners (Decision Tree (DT), Bayesian Network (BN), NS-CL) over 5 runs to learn three diagnostic rules: \textbf{Simple} (\texttt{is\_malignant}); \textbf{Disjunctive} (\texttt{is\_mel\_or\_bcc}); and \textbf{Conjunctive} (\texttt{is\_mel\_on\_back}, a \texttt{dx} AND \texttt{loc} conjunction). Performance is reported as F1-score mean$\pm$ std.

\begin{figure}[tbh]
    \centering
    \vspace*{-\baselineskip}%
    \hspace*{-.1\linewidth}\includegraphics[width=1.20\linewidth]{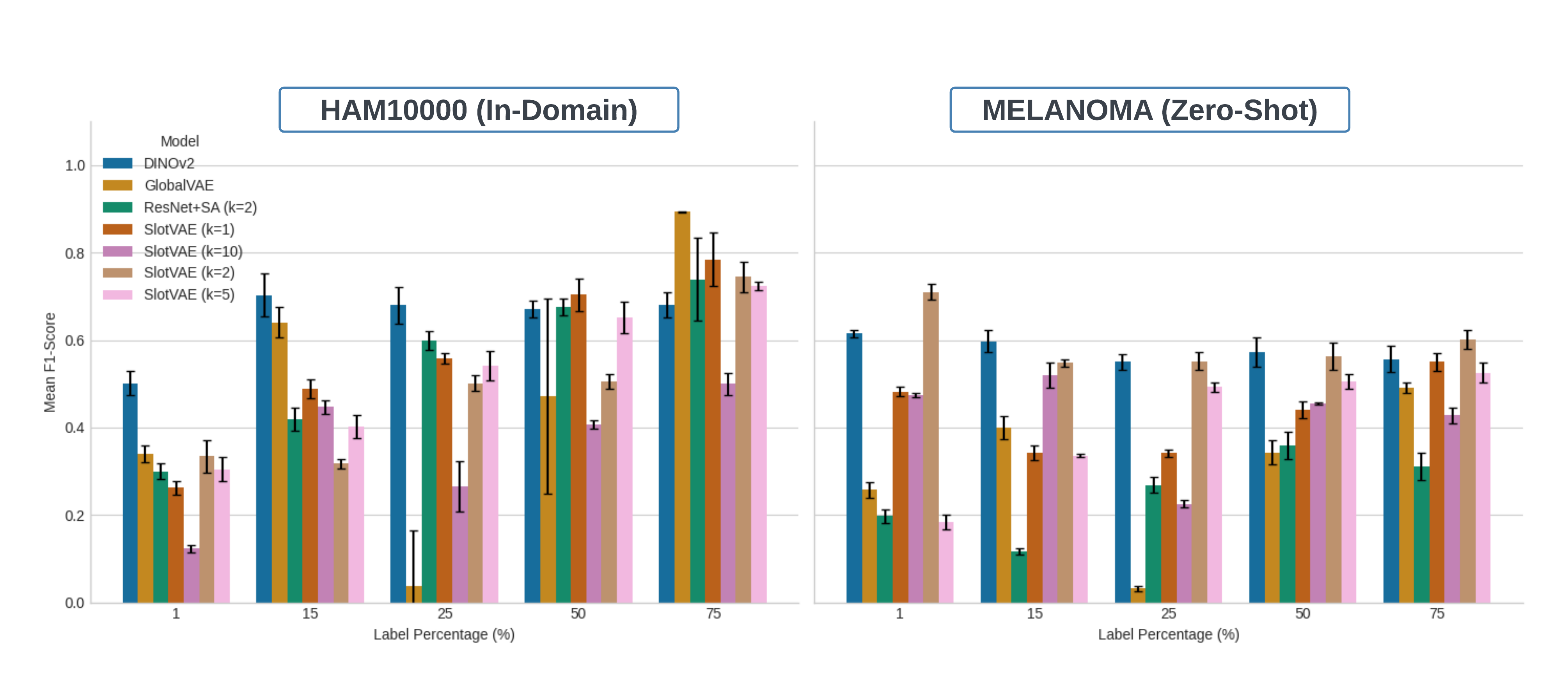}%
    \vspace*{-\baselineskip}%
    \caption{Model accuracy on in-domain (HAM, left) and out of domain (Melanoma, right) tasks across supervision levels. At 1\% supervision, our SlotVAE[k=2] achieves significantly higher zero-shot accuracy (0.746) than all baselines for 1000 Images.}
    \label{fig:melanoma_barchart}
\end{figure}

Results (\cref{tab:full_melanoma_results_acc,fig:melanoma_barchart}) reveal a critical trade-off between in-domain performance and zero-shot generalization, particularly under sparse supervision.

\paragraph{Results: In-Domain vs. Zero-Shot Tradeoff.}
We observed a \emph{clear trade-off between in-domain specialization and zero-shot generalization}. On the in-domain HAM task (\cref{fig:melanoma_barchart}, left), foundation models like DINOv2 were superior at 15\% supervision (0.88 vs.\ 0.80), while our SlotVAE became competitive at 75\% (0.897 vs.\ 0.894), consistent with its objective of finding robust representations rather than maximizing source-task performance. This focus on robust concepts yielded a significant advantage in the zero-shot transfer task (\cref{fig:melanoma_barchart}, right), which revealed a striking reversal under data scarcity. At \textbf{1\% supervision}, our SlotVAE achieved a mean zero-shot accuracy of \textbf{0.746} over 5 runs, substantially outperforming both DINOv2 (0.679) and ResNet+SA (0.579). \emph{This indicates a superior capability for generalizing robust diagnostic features from minimal supervision}.

% \begin{figure}[h!]
%     \centering
%     \includegraphics[width=0.75\linewidth]{Latent_plots/RQ4/model_results_heatmap (2).png}
%     \caption{Best Harmonic Mean Accuracy by model and supervision level. While DINOv2 (top) peaks, our SlotVAE (bottom) is the most competitive model under extreme data scarcity (1\% labels) and shows strong, consistent performance overall.}
%     \label{fig:melanoma_heatmap}
% \end{figure}

% \paragraph{Overall Performance and Training Efficiency.}
% To balance both specialization and generalization, we analyze the harmonic mean. While DINOv2 achieves the highest single H-Mean score (0.785 at 25\% labels), our SlotVAE is the most consistent and data-efficient performer. It achieves the highest H-Mean at both the 1\% (0.720) and 50\% (0.748) supervision levels, establishing it as the most balanced model. % S

% Furthermore, Table \ref{tab:full_melanoma_results} highlights the training efficiency of our approach. Our SlotVAE often achieves its best-performing checkpoints in significantly fewer epochs (e.g., 65 epochs at 1\% supervision, 45 epochs at 50\%) compared to the fixed 100-epoch training of the baselines. This suggests our weakly-supervised VAE architecture not only generalizes more effectively from sparse data but also converges to a robust solution more rapidly.

\begin{table}[tb]
\centering
\caption{Evaluation results for diagnostic \textbf{accuracy} on HAM (in-domain) and Melanoma (zero-shot) datasets. All values are mean$\pm$std. over 5 runs for 1000 images}
\label{tab:full_melanoma_results_acc}
\scriptsize % Use smaller font
\resizebox{\columnwidth}{!}{%
\begin{tabular}{@{}l cccc@{}}
\toprule
\textbf{Model} & \textbf{Superv. (\%)} & \textbf{In-Domain} & \textbf{Zero-Shot} & \textbf{H-Mean} \\
\midrule
DINOv2 & 1 & \textbf{0.8153 $\pm$ 0.0192} & 0.6947 $\pm$ 0.0084 & \textbf{0.7500 $\pm$ 0.0042} \\
GlobalVAE & 1 & 0.7787 $\pm$ 0.0170 & 0.4873 $\pm$ 0.0121 & 0.5993 $\pm$ 0.0077 \\
RN+SA (k=2) & 1 & 0.8077 $\pm$ 0.0172 & 0.5473 $\pm$ 0.0142 & 0.6525 $\pm$ 0.0148 \\
SlotVAE (k=2) & 1 & 0.6567 $\pm$ 0.0165 & \textbf{0.7467 $\pm$ 0.0100} & 0.6988 $\pm$ 0.0136 \\
\midrule
DINOv2 & 15 & \textbf{0.8827 $\pm$ 0.0552} & \textbf{0.6730 $\pm$ 0.0026} & \textbf{0.7632 $\pm$ 0.0185} \\
GlobalVAE & 15 & 0.8620 $\pm$ 0.0877 & 0.5620 $\pm$ 0.0213 & 0.6784 $\pm$ 0.0133 \\
RN+SA (k=2) & 15 & 0.8327 $\pm$ 0.0080 & 0.5217 $\pm$ 0.0106 & 0.6414 $\pm$ 0.0090 \\
SlotVAE (k=2) & 15 & 0.8000 $\pm$ 0.0082 & 0.6403 $\pm$ 0.0139 & 0.7112 $\pm$ 0.0054 \\
\midrule
DINOv2 & 25 & \textbf{0.8843 $\pm$ 0.0393} & \textbf{0.6660 $\pm$ 0.0090} & \textbf{0.7596 $\pm$ 0.0193} \\
GlobalVAE & 25 & 0.8050 $\pm$ 0.0139 & 0.5057 $\pm$ 0.0155 & 0.6211 $\pm$ 0.0147 \\
RN+SA (k=2) & 25 & 0.8463 $\pm$ 0.0015 & 0.5593 $\pm$ 0.0137 & 0.6735 $\pm$ 0.0095 \\
SlotVAE (k=2) & 25 & 0.8233 $\pm$ 0.0167 & 0.6400 $\pm$ 0.0104 & 0.7201 $\pm$ 0.0072 \\
\midrule
DINOv2 & 50 & \textbf{0.8880 $\pm$ 0.0135} & \textbf{0.6870 $\pm$ 0.0079} & \textbf{0.7747 $\pm$ 0.0099} \\
GlobalVAE & 50 & 0.8413 $\pm$ 0.0214 & 0.5690 $\pm$ 0.0075 & 0.6788 $\pm$ 0.0109 \\
RN+SA (k=2) & 50 & 0.8737 $\pm$ 0.0032 & 0.5920 $\pm$ 0.0121 & 0.7057 $\pm$ 0.0093 \\
SlotVAE (k=2) & 50 & 0.8310 $\pm$ 0.0082 & 0.6660 $\pm$ 0.0125 & 0.7394 $\pm$ 0.0106 \\
\midrule
DINOv2 & 75 & 0.8943 $\pm$ 0.0083 & \textbf{0.6773 $\pm$ 0.0086} & \textbf{0.7708 $\pm$ 0.0052} \\
GlobalVAE & 75 & \textbf{0.9583 $\pm$ 0.0147} & 0.5723 $\pm$ 0.0125 & 0.7165 $\pm$ 0.0062 \\
RN+SA (k=2) & 75 & 0.8913 $\pm$ 0.0096 & 0.5683 $\pm$ 0.0040 & 0.6941 $\pm$ 0.0030 \\
SlotVAE (k=2) & 75 & 0.8970 $\pm$ 0.0144 & 0.6623 $\pm$ 0.0172 & 0.7619 $\pm$ 0.0149 \\
\bottomrule
\end{tabular}
}
\end{table}

\paragraph{Results: Baseline Comparison on Reasoning.}
To complete our analysis, we evaluate pipeline accuracy by assessing how downstream reasoners learn logical rules of varying complexity from predicates generated by our SlotVAE and SOTA baselines.
%
%For that we use a pool of 30 frozen predicate generators, comprising 6 distinct model configurations (our SlotVAE ablations and SOTA baselines) each trained at 5 different HAM supervision levels (1\%, 15\%, 25\%, 50\%, and 75\%).
% For 1000 test images, each model generated predicates \texttt{vae\_dx}, \texttt{vae\_loc}, \texttt{vae\_age}), which were compared to a ground truth set. These sets were used to train three \textbf{downstream reasoners} (Decision Tree (DT), Bayesian Network (BN), NS-CL) over 5 runs to learn three diagnostic rules: \textbf{Simple} (\texttt{is\_malignant}); \textbf{Disjunctive} (\texttt{is\_mel\_or\_bcc}); and \textbf{Conjunctive} (\texttt{is\_mel\_on\_back}, a \texttt{dx} AND \texttt{loc} conjunction). Performance is reported as F1-score mean$\pm$ std.
%
Results summarized in \cref{tab:reasoning_comparison}, reveal that the reasoning frameworks (DT, NS-CL) are highly effective when given clean data, but the overall pipeline performance is critically bottlenecked by the perceptual fidelity of the predicate generators, especially for complex rules.

%\paragraph{Establishing the Reasoning Baseline.}
On ground truth predicates, DT and NS-CL achieved near-perfect F1-scores ($\geq 0.99$) for all rules, confirming the task is solvable. The BN performed poorly (F1 $\approx$ 0.0-0.2) is excluded from this analysis.

\begin{table*}[tb]
\centering
\caption{Comparative F1-Scores (Mean $\pm$ Std. Dev.) of Downstream Reasoners on 5000 [neg:4000,pos:1000] HAM, including SlotVAE ablations. Each cell shows the F1-score for NS-CL / Decision Tree. The Bayesian Network (F1 $\in [0.0,0.2]$) is omitted for clarity.}
\label{tab:reasoning_comparison}
\scriptsize
\begin{tabular}{@{}c|l|c|c|c@{}}
\toprule
\textbf{Superv. (\%)} & \textbf{Model (Predicate Generator)} & \textbf{Rule 1: is\_malignant} & \textbf{Rule 2: is\_mel\_or\_bcc} & \textbf{Rule 3: is\_mel\_on\_back} \\
\midrule
\textbf{GT} & \textbf{GroundTruth (Oracle)} & \textbf{1.000 $\pm$ 0.000 / 1.000 $\pm$ 0.000} & \textbf{1.000 $\pm$ 0.000 / 1.000 $\pm$ 0.000} & \textbf{0.998 $\pm$ 0.005 / 1.000 $\pm$ 0.000} \\
\midrule
\multirow{7}{*}{\textbf{1\%}} 
& DINOv2 & \textbf{0.370 $\pm$ 0.110 / 0.298 $\pm$ 0.106} & \textbf{0.160 $\pm$ 0.089 / 0.076 $\pm$ 0.028} & 0.000 $\pm$ 0.000 / 0.036 $\pm$ 0.027 \\
& GlobalVAE & 0.166 $\pm$ 0.054 / 0.137 $\pm$ 0.043 & 0.046 $\pm$ 0.028 / 0.041 $\pm$ 0.021 & 0.000 $\pm$ 0.000 / 0.000 $\pm$ 0.000 \\
& ResNet+SA (k=2) & 0.290 $\pm$ 0.038 / 0.188 $\pm$ 0.117 & 0.000 $\pm$ 0.000 / 0.034 $\pm$ 0.037 & 0.000 $\pm$ 0.000 / 0.000 $\pm$ 0.000 \\
& SlotVAE (k=1) & 0.136 $\pm$ 0.032 / 0.125 $\pm$ 0.011 & 0.012 $\pm$ 0.018 / 0.055 $\pm$ 0.013 & 0.000 $\pm$ 0.000 / 0.007 $\pm$ 0.017 \\
& SlotVAE (k=2) & 0.083 $\pm$ 0.046 / 0.157 $\pm$ 0.057 & 0.019 $\pm$ 0.022 / 0.046 $\pm$ 0.019 & 0.000 $\pm$ 0.000 / 0.007 $\pm$ 0.015 \\
& SlotVAE (k=5) & 0.000 $\pm$ 0.000 / 0.117 $\pm$ 0.055 & 0.000 $\pm$ 0.000 / 0.048 $\pm$ 0.032 & 0.000 $\pm$ 0.000 / 0.045 $\pm$ 0.016 \\
& SlotVAE (k=10) & 0.001 $\pm$ 0.003 / 0.017 $\pm$ 0.015 & 0.000 $\pm$ 0.000 / 0.019 $\pm$ 0.009 & 0.000 $\pm$ 0.000 / 0.000 $\pm$ 0.000 \\
\midrule
\multirow{7}{*}{\textbf{15\%}} 
& DINOv2 & \textbf{0.617 $\pm$ 0.029 / 0.633 $\pm$ 0.017} & \textbf{0.548 $\pm$ 0.026 / 0.507 $\pm$ 0.044} & 0.000 $\pm$ 0.000 / 0.112 $\pm$ 0.040 \\
& GlobalVAE & 0.522 $\pm$ 0.044 / 0.550 $\pm$ 0.028 & 0.369 $\pm$ 0.038 / 0.187 $\pm$ 0.021 & 0.000 $\pm$ 0.000 / 0.057 $\pm$ 0.041 \\
& ResNet+SA (k=2) & 0.399 $\pm$ 0.024 / 0.424 $\pm$ 0.020 & 0.291 $\pm$ 0.034 / 0.304 $\pm$ 0.037 & 0.000 $\pm$ 0.000 / 0.015 $\pm$ 0.021 \\
& SlotVAE (k=1) & 0.393 $\pm$ 0.036 / 0.393 $\pm$ 0.025 & 0.247 $\pm$ 0.056 / 0.204 $\pm$ 0.070 & 0.000 $\pm$ 0.000 / 0.014 $\pm$ 0.019 \\
& SlotVAE (k=2) & 0.225 $\pm$ 0.085 / 0.327 $\pm$ 0.081 & 0.107 $\pm$ 0.078 / 0.227 $\pm$ 0.019 & 0.000 $\pm$ 0.000 / 0.000 $\pm$ 0.000 \\
& SlotVAE (k=5) & 0.083 $\pm$ 0.019 / 0.082 $\pm$ 0.021 & 0.020 $\pm$ 0.025 / 0.063 $\pm$ 0.016 & 0.000 $\pm$ 0.000 / 0.000 $\pm$ 0.000 \\
& SlotVAE (k=10) & 0.117 $\pm$ 0.007 / 0.124 $\pm$ 0.027 & 0.019 $\pm$ 0.035 / 0.082 $\pm$ 0.020 & 0.000 $\pm$ 0.000 / 0.000 $\pm$ 0.000 \\
\midrule
\multirow{7}{*}{\textbf{25\%}} 
& DINOv2 & \textbf{0.691 $\pm$ 0.015 / 0.703 $\pm$ 0.022} & \textbf{0.638 $\pm$ 0.021 / 0.621 $\pm$ 0.036} & 0.000 $\pm$ 0.000 / 0.014 $\pm$ 0.019 \\
& GlobalVAE & 0.056 $\pm$ 0.018 / 0.047 $\pm$ 0.019 & 0.045 $\pm$ 0.015 / 0.040 $\pm$ 0.023 & 0.000 $\pm$ 0.000 / 0.000 $\pm$ 0.000 \\
& ResNet+SA (k=2) & 0.509 $\pm$ 0.015 / 0.512 $\pm$ 0.016 & 0.407 $\pm$ 0.050 / 0.355 $\pm$ 0.083 & 0.000 $\pm$ 0.000 / 0.022 $\pm$ 0.020 \\
& SlotVAE (k=1) & 0.547 $\pm$ 0.027 / 0.506 $\pm$ 0.054 & 0.401 $\pm$ 0.048 / 0.285 $\pm$ 0.052 & 0.000 $\pm$ 0.000 / \textbf{0.122 $\pm$ 0.066} \\
& SlotVAE (k=2) & 0.439 $\pm$ 0.041 / 0.459 $\pm$ 0.046 & 0.279 $\pm$ 0.132 / 0.287 $\pm$ 0.135 & 0.000 $\pm$ 0.000 / 0.022 $\pm$ 0.032 \\
& SlotVAE (k=5) & 0.265 $\pm$ 0.025 / 0.259 $\pm$ 0.059 & 0.082 $\pm$ 0.071 / 0.089 $\pm$ 0.042 & 0.000 $\pm$ 0.000 / 0.000 $\pm$ 0.000 \\
& SlotVAE (k=10) & 0.211 $\pm$ 0.018 / 0.170 $\pm$ 0.063 & 0.002 $\pm$ 0.004 / 0.073 $\pm$ 0.025 & 0.000 $\pm$ 0.000 / 0.015 $\pm$ 0.021 \\
\midrule
\multirow{7}{*}{\textbf{50\%}} 
& DINOv2 & \textbf{0.723 $\pm$ 0.015 / 0.719 $\pm$ 0.013} & \textbf{0.669 $\pm$ 0.022 / 0.664 $\pm$ 0.019} & 0.008 $\pm$ 0.018 / \textbf{0.139 $\pm$ 0.068} \\
& GlobalVAE & 0.496 $\pm$ 0.022 / 0.506 $\pm$ 0.029 & 0.399 $\pm$ 0.064 / 0.414 $\pm$ 0.039 & 0.000 $\pm$ 0.000 / 0.102 $\pm$ 0.081 \\
& ResNet+SA (k=2) & 0.671 $\pm$ 0.016 / 0.667 $\pm$ 0.020 & 0.577 $\pm$ 0.018 / 0.582 $\pm$ 0.038 & 0.000 $\pm$ 0.000 / 0.071 $\pm$ 0.098 \\
& SlotVAE (k=1) & 0.686 $\pm$ 0.015 / 0.683 $\pm$ 0.008 & 0.607 $\pm$ 0.015 / 0.615 $\pm$ 0.018 & 0.000 $\pm$ 0.000 / 0.055 $\pm$ 0.031 \\
& SlotVAE (k=2) & 0.396 $\pm$ 0.063 / 0.343 $\pm$ 0.078 & 0.340 $\pm$ 0.035 / 0.239 $\pm$ 0.092 & 0.000 $\pm$ 0.000 / 0.007 $\pm$ 0.016 \\
& SlotVAE (k=5) & 0.389 $\pm$ 0.046 / 0.387 $\pm$ 0.054 & 0.203 $\pm$ 0.082 / 0.190 $\pm$ 0.100 & 0.000 $\pm$ 0.000 / 0.008 $\pm$ 0.018 \\
& SlotVAE (k=10) & 0.147 $\pm$ 0.044 / 0.159 $\pm$ 0.028 & 0.010 $\pm$ 0.010 / 0.067 $\pm$ 0.021 & 0.000 $\pm$ 0.000 / 0.000 $\pm$ 0.000 \\
\midrule
\multirow{7}{*}{\textbf{75\%}} 
& DINOv2 & 0.678 $\pm$ 0.027 / 0.679 $\pm$ 0.021 & 0.622 $\pm$ 0.022 / 0.617 $\pm$ 0.024 & 0.008 $\pm$ 0.017 / 0.184 $\pm$ 0.067 \\
& GlobalVAE & \textbf{0.929 $\pm$ 0.009 / 0.928 $\pm$ 0.011} & \textbf{0.912 $\pm$ 0.012 / 0.903 $\pm$ 0.006} & \textbf{0.908 $\pm$ 0.025 / 0.884 $\pm$ 0.022} \\
& ResNet+SA (k=2) & 0.739 $\pm$ 0.015 / 0.745 $\pm$ 0.010 & 0.683 $\pm$ 0.011 / 0.676 $\pm$ 0.018 & 0.205 $\pm$ 0.227 / 0.312 $\pm$ 0.140 \\
& SlotVAE (k=1) & 0.759 $\pm$ 0.018 / 0.751 $\pm$ 0.021 & 0.717 $\pm$ 0.022 / 0.707 $\pm$ 0.036 & 0.606 $\pm$ 0.018 / 0.400 $\pm$ 0.194 \\
& SlotVAE (k=2) & 0.718 $\pm$ 0.016 / 0.737 $\pm$ 0.016 & 0.670 $\pm$ 0.018 / 0.654 $\pm$ 0.025 & 0.000 $\pm$ 0.000 / 0.103 $\pm$ 0.102 \\
& SlotVAE (k=5) & 0.270 $\pm$ 0.058 / 0.207 $\pm$ 0.069 & 0.149 $\pm$ 0.037 / 0.065 $\pm$ 0.016 & 0.000 $\pm$ 0.000 / 0.015 $\pm$ 0.021 \\
& SlotVAE (k=10) & 0.356 $\pm$ 0.026 / 0.288 $\pm$ 0.070 & 0.014 $\pm$ 0.026 / 0.064 $\pm$ 0.018 & 0.008 $\pm$ 0.017 / 0.008 $\pm$ 0.018 \\
\bottomrule
\end{tabular}
\end{table*}

%\paragraph{Performance on Simple \& Disjunctive Rules (Rule 1, 2).}
\emph{For simple \texttt{dx}-only rules (Rule 1, 2), performance scaled with predicate quality.} At 1\% supervision, all models produced low F1-scores. With more supervision (15\%-75\%), predicates from DINOv2 and GlobalVAE improved, enabling high F1-scores (e.g., 0.90 for GlobalVAE + NS-CL), which aligns with their strong in-domain accuracy on the dominant \texttt{dx} concept.

A key finding is a \textbf{perception bottleneck}, revealed by a stark mismatch between high-in-domain diagnostic accuracy (\cref{tab:full_melanoma_results_acc}) and low downstream reasoning F1-scores (\cref{tab:reasoning_comparison}). \emph{At 1-50\% supervision, models like DINOv2 had high accuracy ($\approx$81-88\%) yet their predicates led to poor reasoning.} This failure was most pronounced on the conjunctive Rule 3, where most models scored an F1 $\approx$ 0.0.
This is a failure of the perception modules, not the reasoners (which achieved F1 $\approx$ 1.0 on Ground Truth). Rule 3 requires simultaneous correctness on both the \texttt{dx} and \texttt{localization} concepts. However, perceptual fidelity for the secondary localization concept was extremely low (e.g., SlotVAE max 44.64\% per \cref{tab:slot_comparison}). This creates a fatal bottleneck, rendering the conjunctive rule unsolvable. The sole exception was GlobalVAE at 75\% supervision, which aligned high in-domain accuracy ($\approx$96\%) with strong reasoning F1-scores ($\approx$0.90-0.93). This underscores \emph{the need for perception modules that can provide high-fidelity predicates for multiple concepts simultaneously}.

\paragraph{Summary and Discussion:}
% Our investigation in this real-world case study began with an architectural ablation on HAM, which validated our object-centric inductive bias (\cref{tab:slot_comparison}). We found a minimal $k=2$ (lesion + background) slot configuration is optimal, yielding a \textbf{+18.2\%} absolute \texttt{dx} accuracy improvement over $k=1$ at 1\% supervision (67.7\% vs 49.5\%). This demonstrates that providing the correct structural bias is most critical when the supervisory signal is weakest.
%
We benchmarked our SlotVAE predicate quality against baselines including DINOv2~\cite{oquab2024dinov2learningrobustvisual}, revealing a key trade-off: \emph{While the larger DINOv2 achieved higher in-domain accuracy, our SlotVAE demonstrated superior zero-shot generalization of predicates.} At 1\% supervision, our model's zero-shot accuracy was 0.746, significantly outperforming DINOv2 (0.6947) and all other baselines (\cref{fig:melanoma_barchart}), highlighting its data-efficiency for robust concept generalization.
Finally, we evaluated the \textit{utility} of all generated predicates using downstream reasoners (\cref{tab:reasoning_comparison}). While simple, single-concept rules (e.g., \texttt{is\_malignant}) were learnable, we observed a near-universal failure on the complex conjunctive rule 3. This failure was not in the reasoners (which achieved $F1 \approx 1.0$ on Ground Truth) but in the perception modules. The root cause appears to be an \textit{inductive bias mismatch}. The HAM task is scene-centric, where localization is a diffuse property. Architectures with spatial partitioning biases (our SlotVAE and DINOv2) failed to learn this global concept. Tellingly, the GlobalVAE, which learns a single holistic vector, was the only model to succeed on this rule (e.g., $F1 \approx 0.90$ at 75\% supervision).
Altogether, these findings support our weakly-supervised, object-centric approach, which excels at zero-shot generalization in data-scarce regimes. However, they also highlight that the primary bottleneck for complex, multi-concept reasoning is the non-trivial challenge of matching the perception module's inductive bias to the \textit{nature} of the concepts being learned—whether they are discrete objects or holistic scene properties.

\section{Conclusion}

This work demonstrates that our proposed weakly supervised, two-stage neuro-symbolic SlotVAE architecture is a data-efficient approach to bridge perception and reasoning: A minimal supervisory signal (as low as 1\%) is sufficient to ground simple concepts like shape and size. Our experiments reveal a clear trade-off between supervision and \textit{concept complexity}. While data-efficient for simple concepts (e.g., CLEVR shapes), concepts with high perceptual variance (e.g., CLEVR-TEX, HAM) require more supervision. This creates a bottleneck: complex, multi-concept rules fail if predicate fidelity for even one concept is low.

Despite this, the modular design offers significant advantages, including robust asymmetric domain adaptation and the flexibility to pair with specialized reasoners. We found ILP is remarkably robust to perceptual noise for combinatorial rules, while Bayesian Network and NS-CL models scale better for other logical tasks. Ultimately, our framework's success in the zero-shot medical task (RQ4) with 1\% data proves this architecture can learn robust concepts, offering a data-efficient path toward scalable, trustworthy AI.

For future work we propose investigating how our approach can be further boosted if not sequentially but in parallel or alternatingly training the perception and reasoning stage, building on most recent work on two-stage neuro-symbolic models \cite{roth2025enhancing}.
%We propose investigating a dual optimization process to "close the loop" beyond our current one-way (VAE $\rightarrow$ ILP) pipeline. A future architecture could implement a \textbf{logical consistency loss}: if a high-confidence rule induced by Popper is contradicted by the VAE's predicates on a training example, this logical mismatch would be backpropagated as a loss signal to refine the VAE's weights. This dual optimization would create a symbiotic system where the symbolic reasoner actively refines its neural counterpart.

% \begin{figure}
%     \centering
%     \includegraphics[width=0.23\linewidth]{Latent_plots//RQ4/model_evaluation_report.png}
%     \includegraphics[width=0.23\linewidth]{Latent_plots/RQ4/model_evaluation_report (3).png}
%     \includegraphics[width=0.23\linewidth]{Latent_plots/RQ4/model_evaluation_report (4).png}
%     \includegraphics[width=0.23\linewidth]{Latent_plots/RQ4/model_evaluation_report (1).png}
%     \includegraphics[width=0.23\linewidth]{Latent_plots/RQ4/model_evaluation_report (2).png}
%     \caption{One Shot performance of HAM on Melanoma Dataset wrt different labels {15,25,50,75,100} left to right }
%     \label{fig:placeholder}
% \end{figure}

%%%%%%%%%%%%%%%%%%%%%%%%%%%%%%%%%%%%%%%%%%%%%%%%%%%%%%%%%%%%%%%%%%%%%%
% Conclusion
%%%%%%%%%%%%%%%%%%%%%%%%%%%%%%%%%%%%%%%%%%%%%%%%%%%%%%%%%%%%%%%%%%%%%%

%%%%%%%%%%%%%%%%%%%%%%%%%%%%%%%%%%%%%%%%%%%%%%%%%%%%%%%%%%%%%%%%%%%%%%
% Bibliography
%%%%%%%%%%%%%%%%%%%%%%%%%%%%%%%%%%%%%%%%%%%%%%%%%%%%%%%%%%%%%%%%%%%%%%
{
    \small
    \bibliographystyle{ieeenat_fullname}
    \bibliography{literature}
}

\end{document}

%%%%%%%%% PAPER ID  - PLEASE UPDATE
%\def\paperID{16098} %  * Enter the Paper ID here
%\def\confName{CVPR}
%\def\confYear{2026}

\title{Supplementary Material:\\Weakly Supervised concept learning for object centric visual reasoning} % TODO: shorten title to save the line (exchange CV by just Vision?)

%Label efficient two-stage neuro-symbolic reasoning by weak and self- concept supervision

\author{
Sparsh Tiwari\\
University of Lübeck, Germany\\
{\tt\small sparsh.tiwari@uni-luebeck.de}
\and
Bettina Finzel\\
University of Bamberg, Germany\\
{\tt\small bettina.finzel@uni-bamberg.de}
\and
Gesina Schwalbe\\
University of Ulm, Germany\\
University of Lübeck, Germany\\
{\tt\small gesina.schwalbe@uni-ulm.de}
}

\maketitle

\tableofcontents

\clearpage
\section{Latent Space Analysis of Concept Disentanglement on CLEVR Dataset }

This section presents the latent space visualizations of the Slot-Variational Autoencoder (Slot-VAE) model, which was trained on the CLEVR dataset under conditions of sparse supervision ($15\%$ label supervision).The analysis was conducted on a model configured with $k=10$ object slots after $150$ training epochs

The  UMAP  projections of the latent vectors, colored by corresponding ground-truth conceptual factors (Coordinates, Shape, Size, Color, and Material), reveal a strong degree of  disentanglement  across all tested concepts. Notably, the latent space demonstrates effective clustering for the categorical attributes:
\begin{itemize}
    \item  Structural Concepts (Shape, Size, Material):  These attributes exhibit  clear, well-separated clusters , indicating robust object-centric representation learning.
    \item Surface Properties (Color): Even for the perceptually complex attribute of Color, the latent space forms distinct clusters corresponding to similar color patterns.This successful separation confirms the Slot-VAE's capacity to induce high-confidence, disentangled  predicates  for downstream symbolic reasoning tasks by accurately segmenting and distinguishing multiple colors within the input scenes[cite: 52].
\end{itemize}
The high degree of latent space organization validates the efficacy of the Slot-VAE architecture in grounding and disentangling multiple conceptual factors from complex visual input with minimal label supervision.

\begin{figure}[h!]
    \centering
    % Subfigure for Coordinates
    \begin{subfigure}{0.3\textwidth}
        \includegraphics[width=\linewidth]{Latent_plots/latent_by_coords_epoch_150__1_.png}
        \caption{by Coords}
        \label{fig:latent_coords}
    \end{subfigure}
     % Subfigure for Color
    \begin{subfigure}{0.3\textwidth}
        \includegraphics[width=\linewidth]{Latent_plots/latent_by_color_epoch_150__1_.png}
        \caption{by Color}
        \label{fig:latent_color}
    \end{subfigure}
    % Subfigure for Material
    \begin{subfigure}{0.3\textwidth}
        \includegraphics[width=\linewidth]{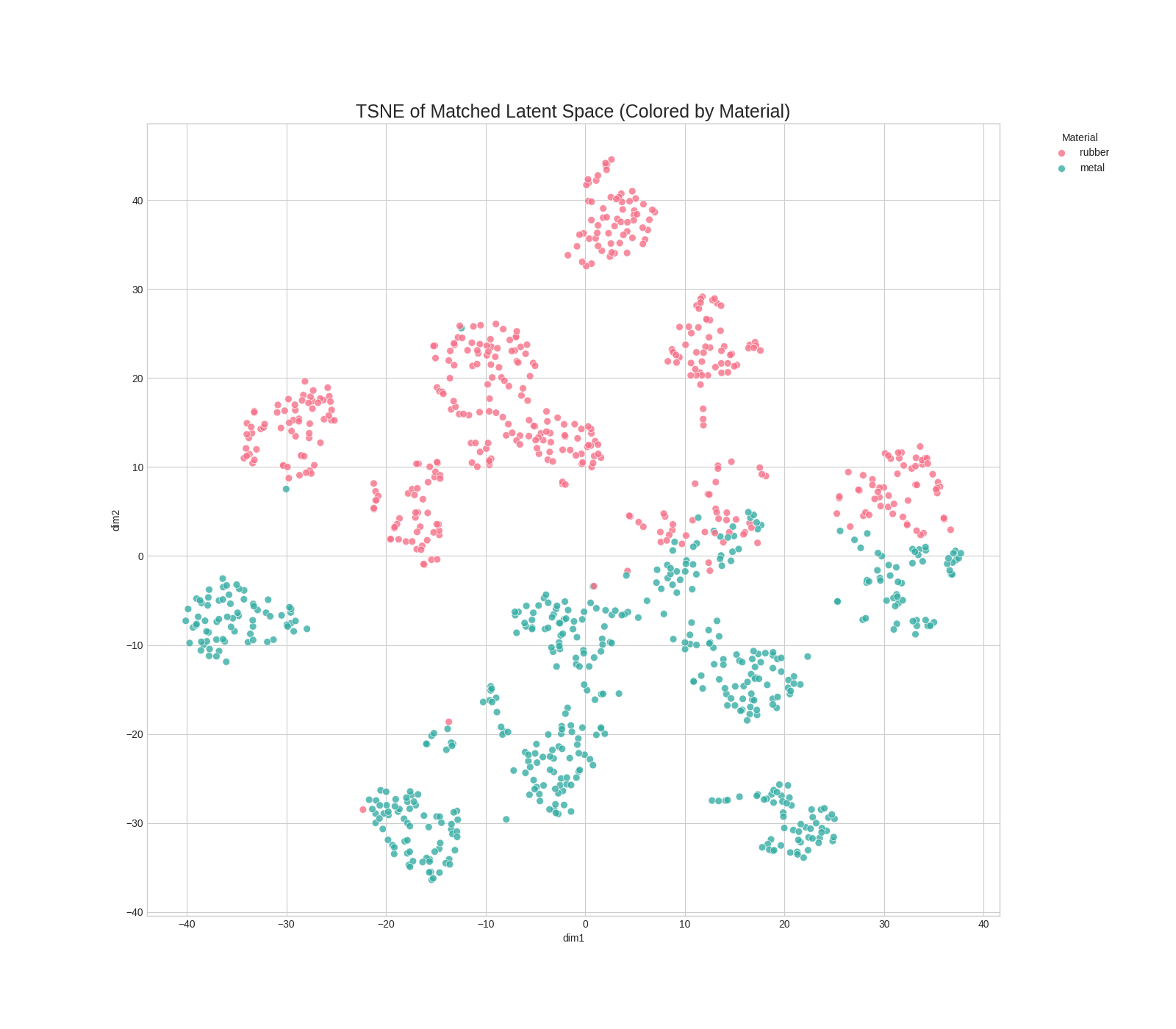}
        \caption{by Material}
        \label{fig:latent_material}
    \end{subfigure}
    % Subfigure for Shape
    \begin{subfigure}{0.3\textwidth}
        \includegraphics[width=\linewidth]{Latent_plots/latent_by_shape_epoch_150__1_.png}
        \caption{by Shape}
        \label{fig:latent_shape}
    \end{subfigure}
    % Subfigure for Size
    \begin{subfigure}{0.3\textwidth}
        \includegraphics[width=\linewidth]{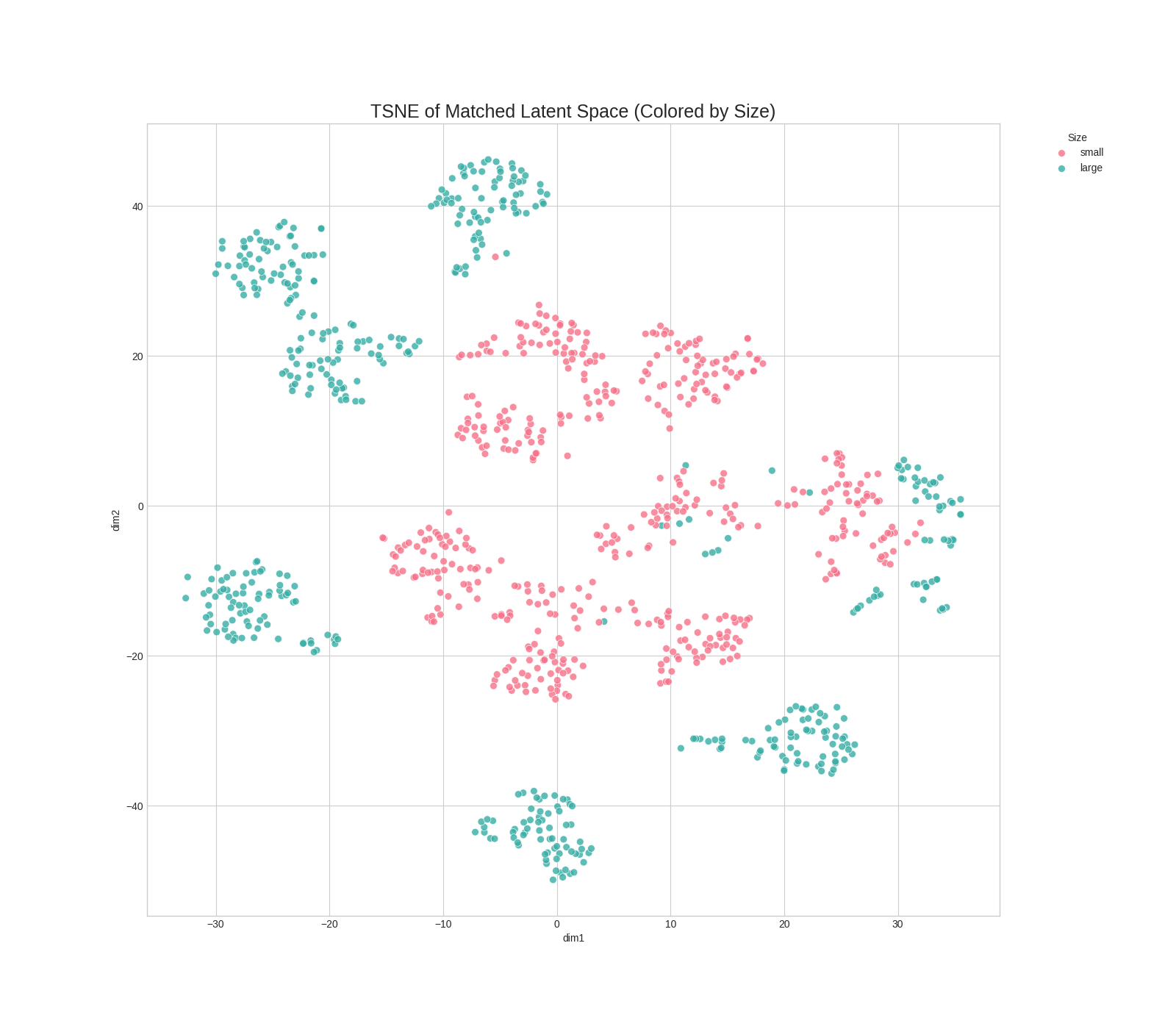}
        \caption{by Size}
        \label{fig:latent_size}
    \end{subfigure}
    
    \caption{UMAP visualizations of the latent space, colored by different ground-truth concepts for 15\% supervision.}
    \label{fig:all_latent_plots}
\end{figure}

\section{Downstream Reasoning Frameworks: Interpretation and Rule Induction}

The flexibility of the modular neuro-symbolic design allows the system to pair the VAE's symbolic outputs with specialized reasoning engines. We detail the mechanism by which the evaluated frameworks—Inductive Logic Programming (ILP), Statistical Tabular Methods (DT and BN), and Neural Symbolic (NS-CL)—interpret the noisy VAE predicates and induce the required logical rules in the paper's empirical analyses (RQ3 and RQ4).

\subsection{Explicit Logical Reasoning: Inductive Logic Programming (ILP)}
The ILP framework, utilizing the \textbf{Popper\cite{cropper2020learningprogramslearningfailures}} solver, embodies the most rigorous, explicit, and interpretable reasoning paradigm.

VAE Output Interpretation
\begin{itemize}
    \item \textbf{Input Format}: Popper consumes the VAE's output directly as a \textbf{relational structure}.
    \item \textbf{Mechanism}: The VAE's predictions are encoded into first-order logic facts, which serve as \textbf{Background Knowledge} (BK) for the solver (e.g., $\texttt{blue(obj1)}$).
\end{itemize}

\subsubsection*{Rule Interpretation (RQ3: Synthetic Rules)}
\begin{itemize}
    \item \textbf{Mechanism}: Popper operates through a Generate, Test, and Constrain loop, learning constraints to prune the hypothesis space and guarantee an optimal, minimally complex solution.
    \item \textbf{Performance}: ILP demonstrated \textbf{superior robustness} for combinatorial rules, achieving a perfect $\texttt{F1-score}$ of $1.0$ on the Simple Existential, Conjunctive, and Disjunctive rules, despite being fed noisy predicates from the $15\%$-supervised VAE checkpoint. This indicates that Popper's constraint-learning mechanism is entirely resilient to the perceptual noise for these fundamental first-order logic structures.
    \item \textbf{Failure Mode}: Its strict logical formalism caused it to \textbf{fail} on non-combinatorial rules like Cardinality and Universal Quantifier ($\texttt{F1}=0.0$).
\end{itemize}

\subsection{Tabular Statistical Reasoning: Decision Tree (DT) and Bayesian Network (BN)}
These classic statistical frameworks require an intermediate step to collapse the VAE's relational output into a fixed-dimensional vector.

\subsubsection*{VAE Output Interpretation}
\begin{itemize}
    \item \textbf{Input Format}: The relational output is transformed into a fixed-width \textbf{Bag-of-Properties (BoP)} vector.
    \item \textbf{Mechanism}: This flattens the scene description by counting the occurrence of specific predicate conjunctions (e.g., counting $[\texttt{red} \land \texttt{cube}]$).
\end{itemize}

\subsubsection*{Rule Interpretation (RQ3: Synthetic Rules)}
\begin{itemize}
    \item \textbf{Decision Tree (DT)}:
    \begin{itemize}
        \item \textbf{Mechanism}: Learns an explicit, hierarchical model based on splitting the BoP vector.
        \item \textbf{Performance}: DT was most effective for the \textbf{Cardinality} (counting) rule ($\texttt{F1}=0.571$), but struggled with other fundamental rules.
    \end{itemize}
    \item \textbf{Bayesian Network (BN)}:
    \begin{itemize}
        \item \textbf{Mechanism}: Models the conditional probability distributions between features in the BoP vector, deriving prediction from a reasoned calculation across all noisy inputs.
        \item \textbf{Performance}: BN excelled at the \textbf{Universal Quantifier} rule ($\texttt{F1}=0.760$). Its probabilistic bias is suited for inference under the uncertainty of VAE predicates. However in \textbf{RQ4}, the BN was often excluded from analysis due to poor performance ($\texttt{F1} \approx 0.0-0.2$).
    \end{itemize}
\end{itemize}

\subsection{Implicit Neural Reasoning: NS-CL (DeepSet Classifier)}
The Neural-Symbolic Concept Learner (NS-CL) framework performs set-based reasoning entirely within a specialized neural architecture, making its reasoning implicit.

\subsubsection*{VAE Output Interpretation}
\begin{itemize}
    \item \textbf{Input Format}: The NS-CL consumes the VAE's output as an \textbf{unordered set of object feature vectors}. This input is essentially the feature vectors $Z_{slot}^{(i)}$ outputted by the Slot Attention module.
    \item \textbf{Mechanism}: The architecture is based on the \textbf{Deep Sets} principle, ensuring \textbf{permutation invariance} (the output is independent of object order).
\end{itemize}

\subsubsection*{Rule Interpretation (RQ3/RQ4: Generalization)}
\begin{itemize}
    \item \textbf{Mechanism}: The model uses a shared encoder ($\phi$) applied to each object vector, followed by a permutation-invariant aggregation function (summation/pooling), and finally a decoder ($\rho$) to map the aggregated feature to classification logits.
    \item \textbf{Performance (RQ3)}: NS-CL also excelled at the \textbf{Universal Quantifier} rule ($\texttt{F1}=0.754$). Reasoning is \textbf{implicit}.
    \item \textbf{Performance (RQ4)}: When applied to the HAM skin lesion task (diagnostic rules), the NS-CL pipeline's performance was critically \textbf{bottlenecked by the low fidelity of the VAE predicates}. For instance, it failed ($\texttt{F1} \approx 0.0$) on the complex conjunctive rule ("is mel\_on\_back") across most supervision levels because the required predicate accuracy for the secondary localization concept was extremely low.
\end{itemize}

\section{CLEVR-2D Dataset: Design and Factor Space for Disentangled Learning}

\begin{figure}
    \centering
    \includegraphics[width=0.21\linewidth]{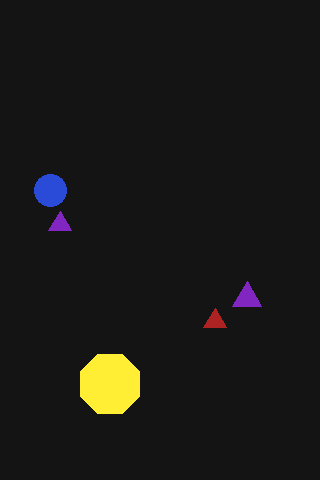}
    \includegraphics[width=0.21\linewidth]{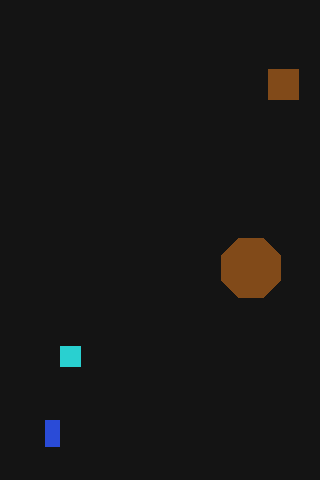}
    \includegraphics[width=0.21\linewidth]{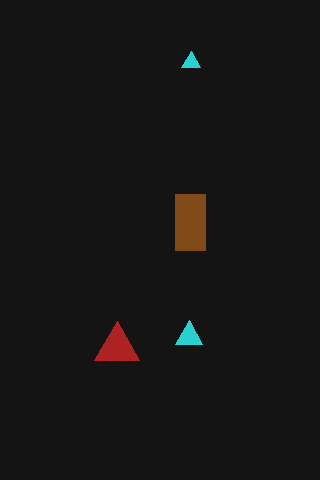}
    \includegraphics[width=0.21\linewidth]{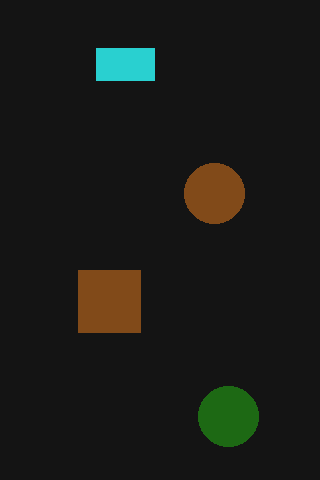}
    \caption{SAMPLES OF 2D Version of CLEVR}
    \label{fig:placeholder}
    \end{figure}
    
Dataset Properties :
The \textbf{CLEVR-2D Dataset} is a synthetically generated benchmark designed as a structurally simplified, two-dimensional analogue of the original CLEVR\cite{johnson2016clevrdiagnosticdatasetcompositional} environment. It serves as a controlled domain for evaluating the capacity of Slot VAE architectures to learn disentangled, object-centric representations, particularly in experiments assessing dimensional generalization (e.g., $2\texttt{D} \leftrightarrow 3\texttt{D}$ transfer). The synthetic generation process enforces a strong, clear inductive bias, crucial for guiding unsupervised representation learning.

\begin{figure}
    \centering
    \includegraphics[width=1.0\linewidth]{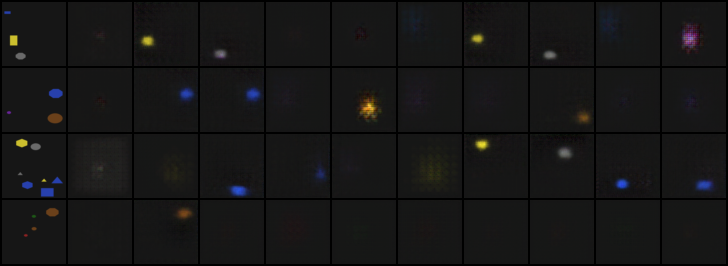}
    \includegraphics[width=0.49\linewidth]{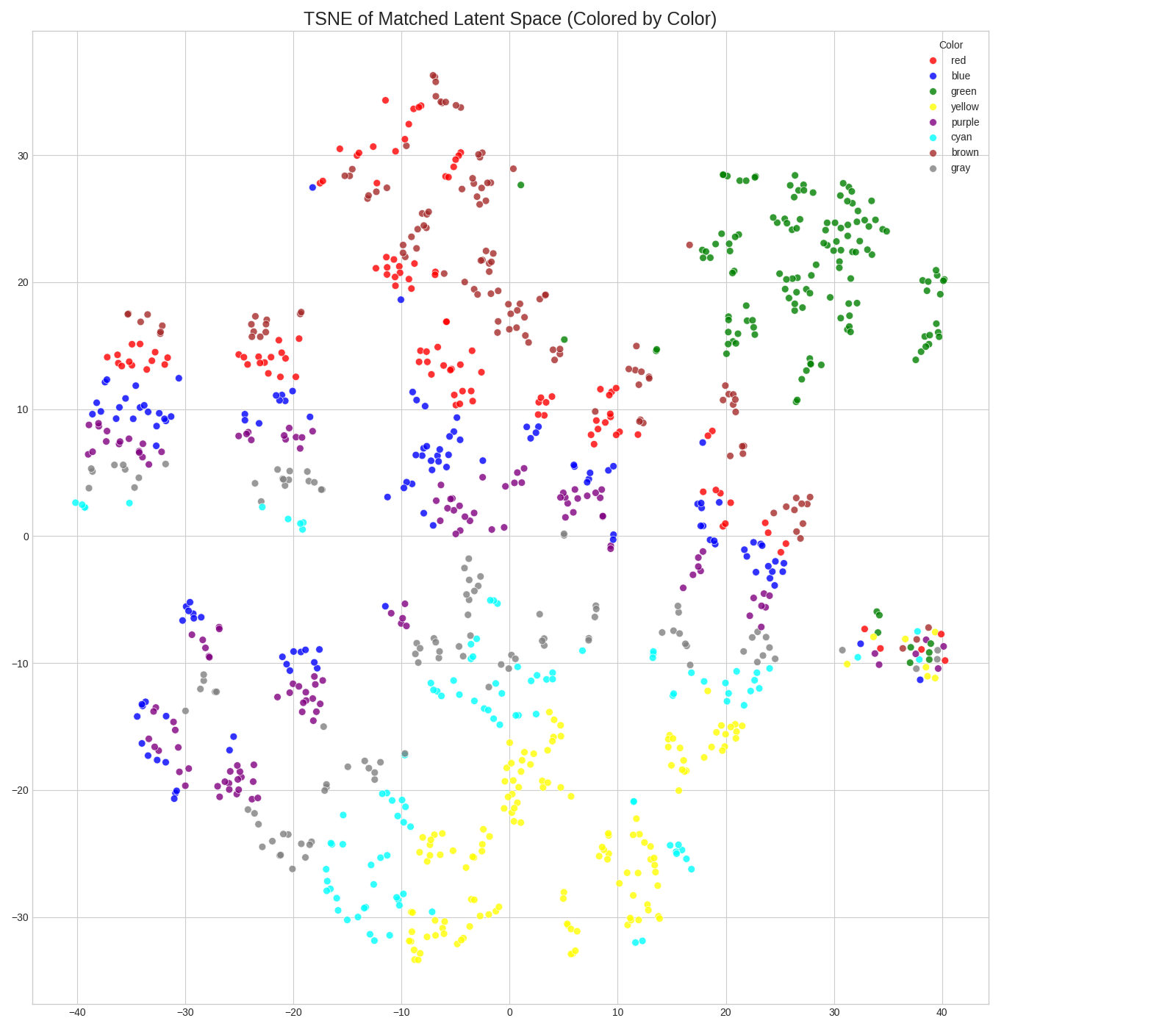}
    \includegraphics[width=0.49\linewidth]{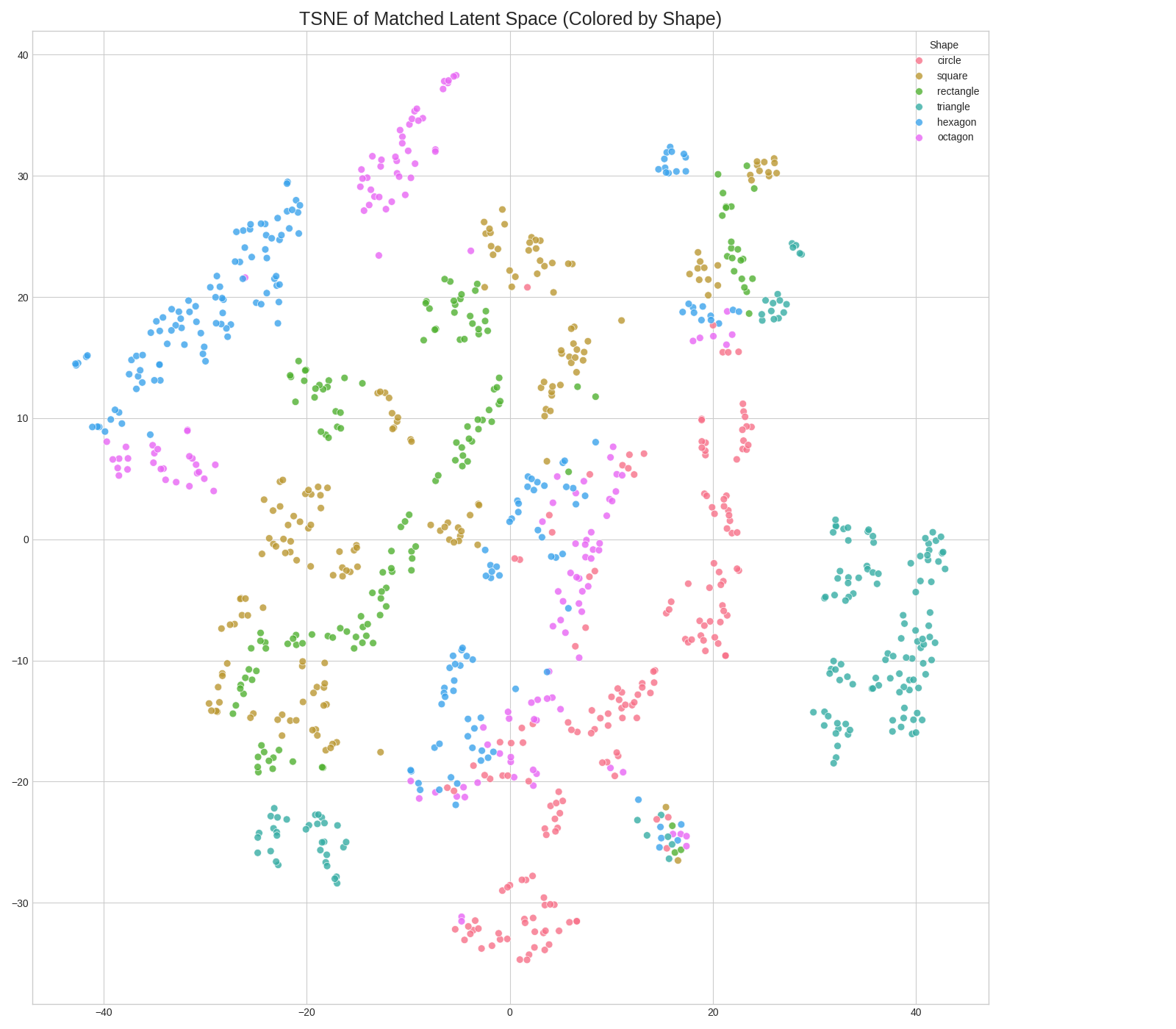}
    \caption{Representing Slot Reconstruction along with VAE latent space for color and shape for model trained on 15\% label supervision on 2D Clevr Dataset, k= 7(Max Objects in a scene)}
    \label{fig:Clevr2D_latent}
\end{figure}

\begin{itemize}
    \item \textbf{Factor Space and Concepts (Ground Truth):} The dataset features a factorially structured space governed by three primary conceptual variables.
    \begin{itemize}
        \item \texttt{SHAPE}: Six discrete primitives are used: $\texttt{circle}$, $\texttt{square}$, $\texttt{rectangle}$, $\texttt{triangle}$, $\texttt{hexagon}$, and $\texttt{octagon}$. The polygons are rendered using a regular polygon function.
        \item \texttt{COLOR}: Eight distinct, fixed RGB colors are defined (e.g., $\texttt{gray}$, $\texttt{red}$, $\texttt{blue}$, $\texttt{yellow}$).
        \item \texttt{SIZE}: Two discrete size classes, $\texttt{small}$  and $\texttt{large}$ , are mapped to controlled continuous ranges to maintain variance while ensuring clear categorical separation.
    \end{itemize}

    \item \textbf{Scene Composition:} Scenes are deliberately complex, containing a variable number of objects ranging from $\texttt{min\_objects}=   3$ to $\texttt{max\_objects}=7$. A non-overlap constraint is enforced during placement, ensuring that the primary challenge remains conceptual binding and segregation rather than handling severe occlusion. The background is fixed to a uniform dark gray.

    \item \textbf{Dimensionality Reduction:} By excluding 3D geometric cues (such as complex volumetric primitives, lighting, shading, and material textures) present in the original CLEVR domain, the dataset provides a low-fidelity, informationally sparse environment. This simplification is key to isolating the model's ability to capture basic structural concepts before transfer.
    
    \item \textbf{Output and Annotation:} Each generated sample includes a $320 \times 480$ pixel image (PNG) and a corresponding JSON scene graph. This metadata contains the full ground-truth conceptual factors (shape, size, color), center pixel coordinates , and a tight $\texttt{bbox}$ annotation for all objects in the scene, facilitating fully supervised concept loss application and downstream relational reasoning.
\end{itemize}

\section{Detailed Results for RQ1}
\begin{figure}
    \centering
    \includegraphics[width=0.75\linewidth]{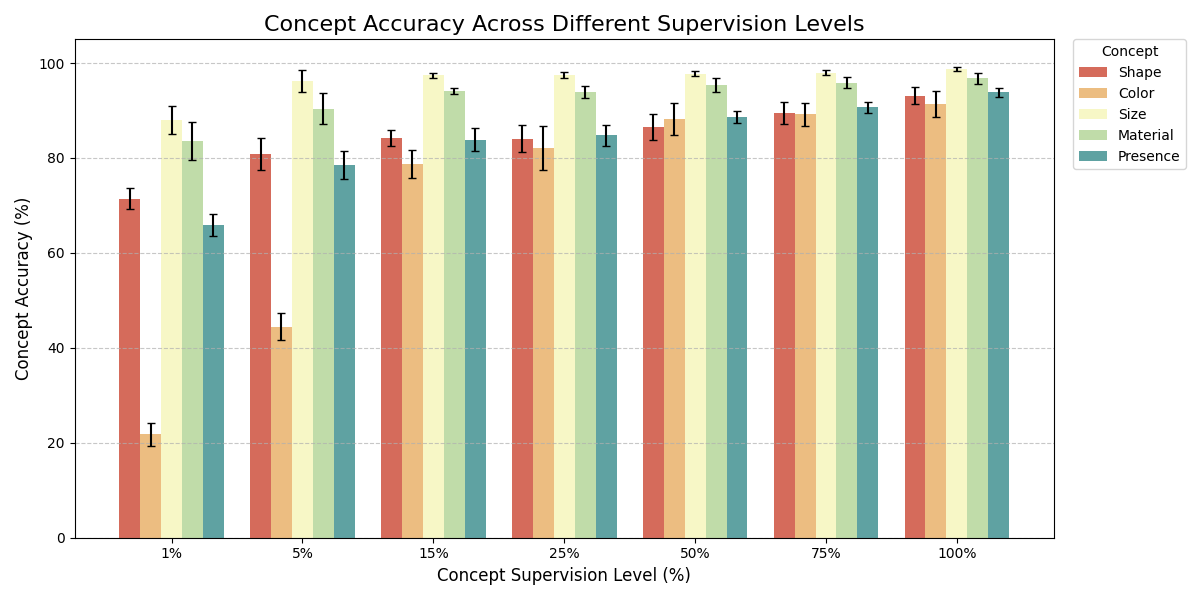}
    \caption{SlotVAE(k=10) Performance on various Label supervision on various Concepts in clevr dataset}
    \label{fig:concept_acc}
\end{figure}

Along with individual concept accuracy as shown in Figure \ref{fig:concept_acc} ,we performed an ablation study to determine the impact of the number of slots ($k$) on predicate accuracy within the Slot-VAE framework, specifically examining its interaction with weak concept supervision on the CLEVR dataset(5000 images). The CLEVR environment is multi-object, featuring up to ten objects per scene, which suggests a minimum required $k$ of 10 to successfully isolate all possible visual entities. We tested $k$ values of 10, 15, and 20.

\subsection{Impact of Number of Slots on CLEVR}

\begin{figure}
    \centering
    \includegraphics[width=0.49\linewidth]{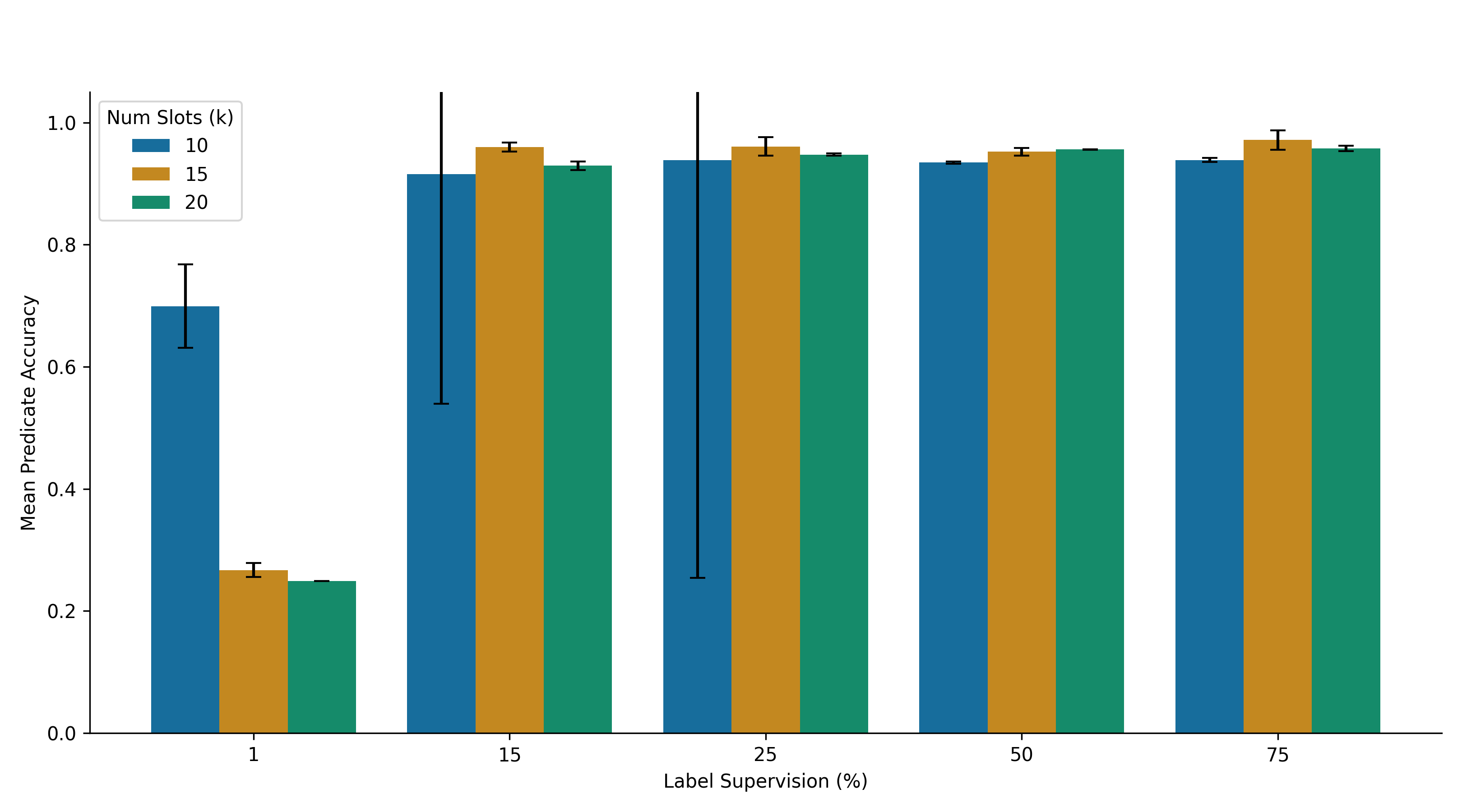}
    \includegraphics[width=0.49\linewidth]{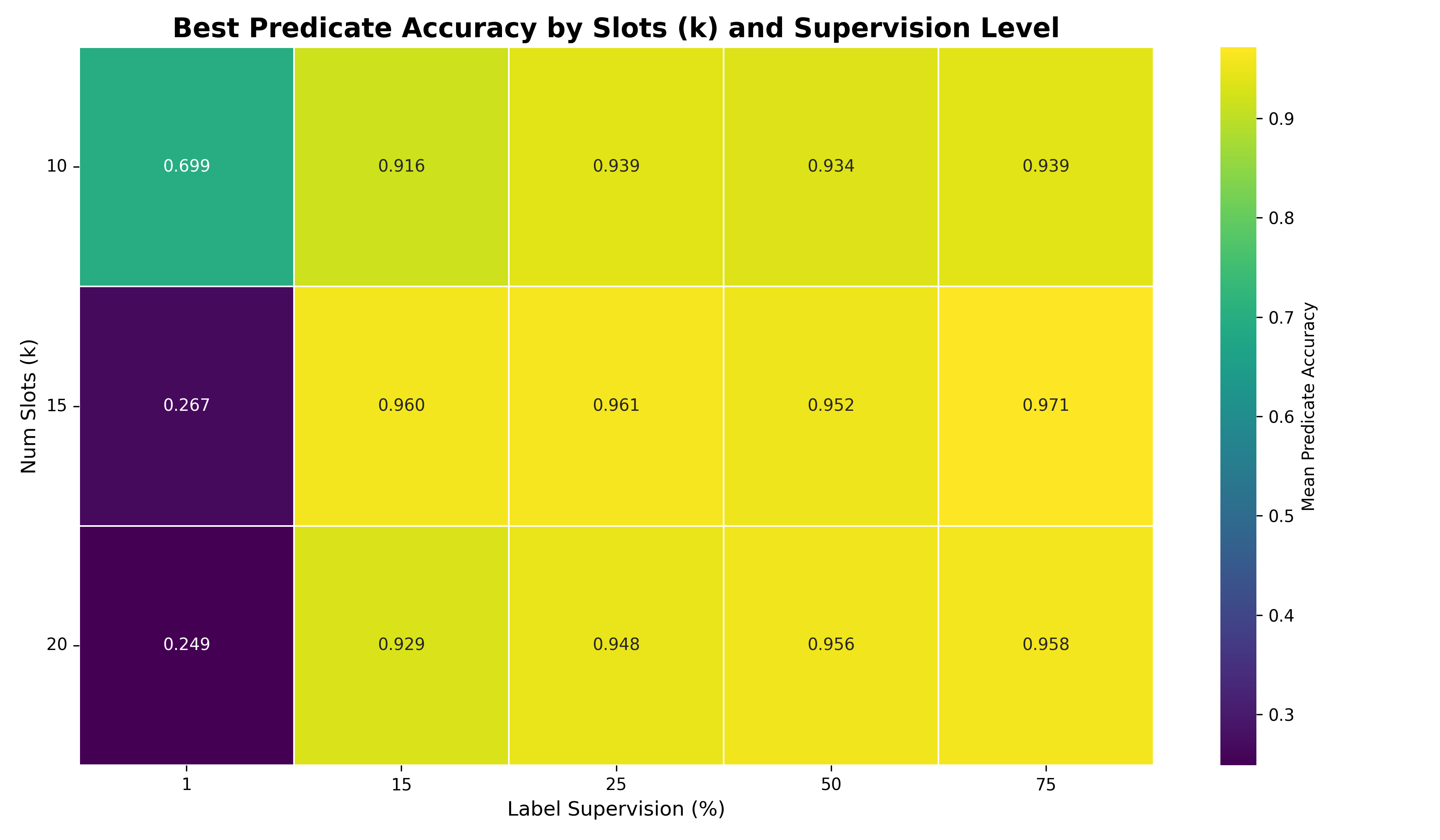}
    \caption{SLOTVAE ON CLEVR 5000 IMAGES vs No. of slots for 25 model in total 5 models for each label supervision}
    \label{fig:Clevr_SLOT_5000}
\end{figure}

\textbf{Sensitivity to Minimal Supervision (1\%):}
The number of slots proved to be most critical when the supervisory signal was weakest (1\% label percentage). At this minimal level, the model performance was highly dependent on the correct inductive bias provided by $k$. 
\begin{itemize}
    \item The configuration $k=10$ (which corresponds to the maximum number of objects in the scene) yielded the strongest performance, achieving an accuracy of approximately $69.9\%$ as seen in Figure \ref{fig:Clevr_SLOT_5000}.
    \item Over-segmentation models, such as $k=15$ and $k=20$, struggled significantly under low supervision, resulting in much lower mean accuracies (approx. $26.7\%$ and $24.9\%$, respectively) and dramatically increased standard deviations (up to $\pm 0.685$), indicating high instability and failure to converge reliably.
\end{itemize}
This demonstrates that excess capacity in the slot mechanism, which leads to highly fragmented attention or irrelevant object decomposition, cannot be corrected by a weak supervisory signal. The correct structural bias ($k=10$) is essential for the Slot-VAE to successfully decouple objects when data is scarce.

\textbf{Robustness under Increased Supervision ($15\%$ and $25\%$):}
As the supervision level increased to $15\%$ and $25\%$, the penalty for using an excessive number of slots ($k > 10$) was largely mitigated by the stronger supervisory signal.

\begin{itemize}
    \item \textbf{Optimal Performance at $k=15$:} At $15\%$ and $25\%$ supervision, the $k=15$ configuration achieved the highest predicate accuracy (up to $96.1\%$ at $25\%$). This suggests that while $k=10$ provides the minimum necessary capacity, a slight increase in $k$ beyond the maximum object count can provide beneficial redundancy or flexibility in complex scenes, provided sufficient ground-truth signal is available to guide the object segmentation as can be seen in Figure \ref{fig:Clevr2D_latent}.
    % \item \textbf{High Stability:} All slot configurations ($k=10, 15, 20$) achieved highly stable and competitive performance at these higher supervision levels (all greater than $91\%$). The standard deviations dropped to minimal values (e.g., $k=10$ at $15\%$ supervision showed $\pm 0.002$), confirming that the model has successfully decoupled the objects and learned robust, meaningful predicates.
\end{itemize}

\begin{figure}
    \centering
    \includegraphics[width=0.49\linewidth]{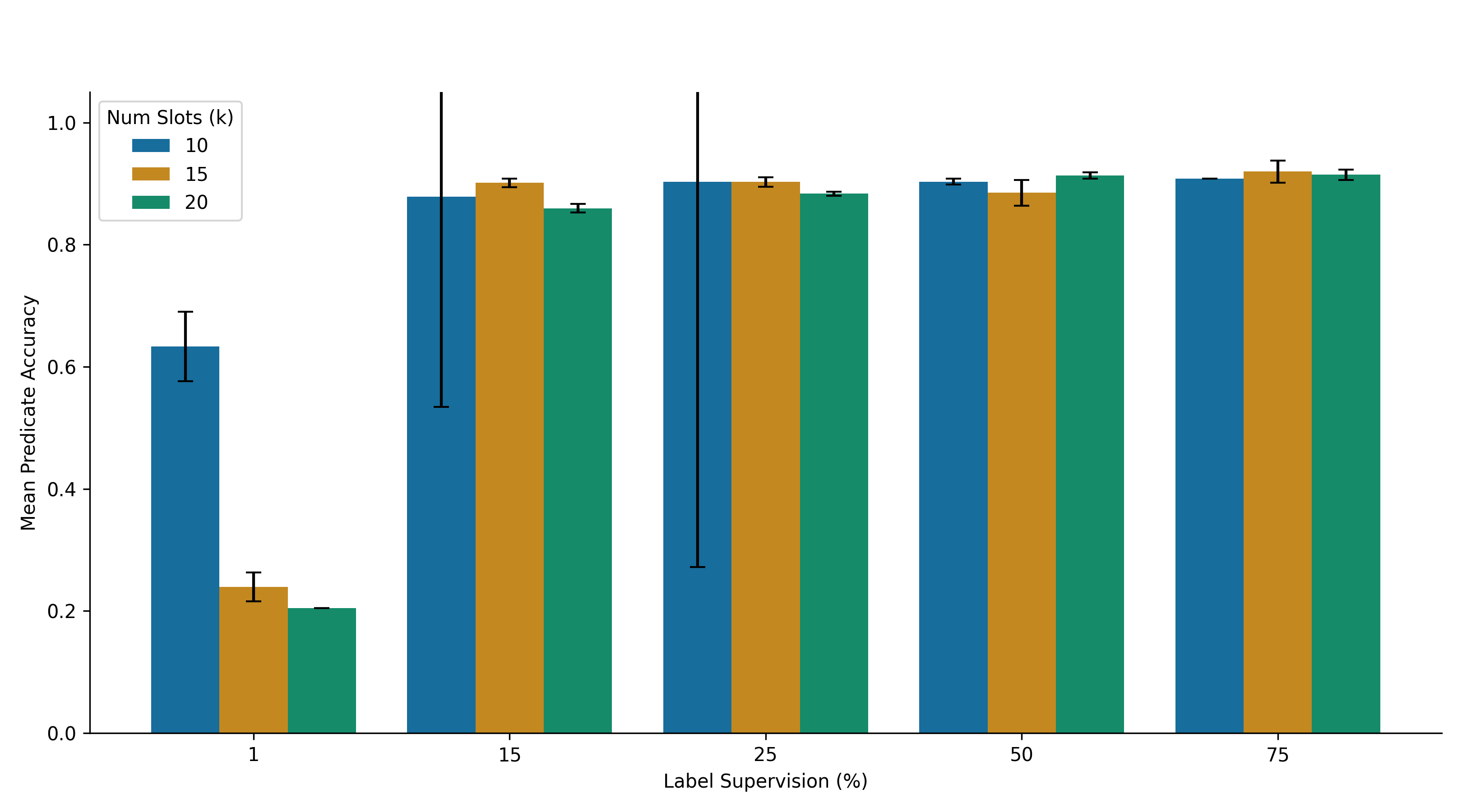}
    \includegraphics[width=0.49\linewidth]{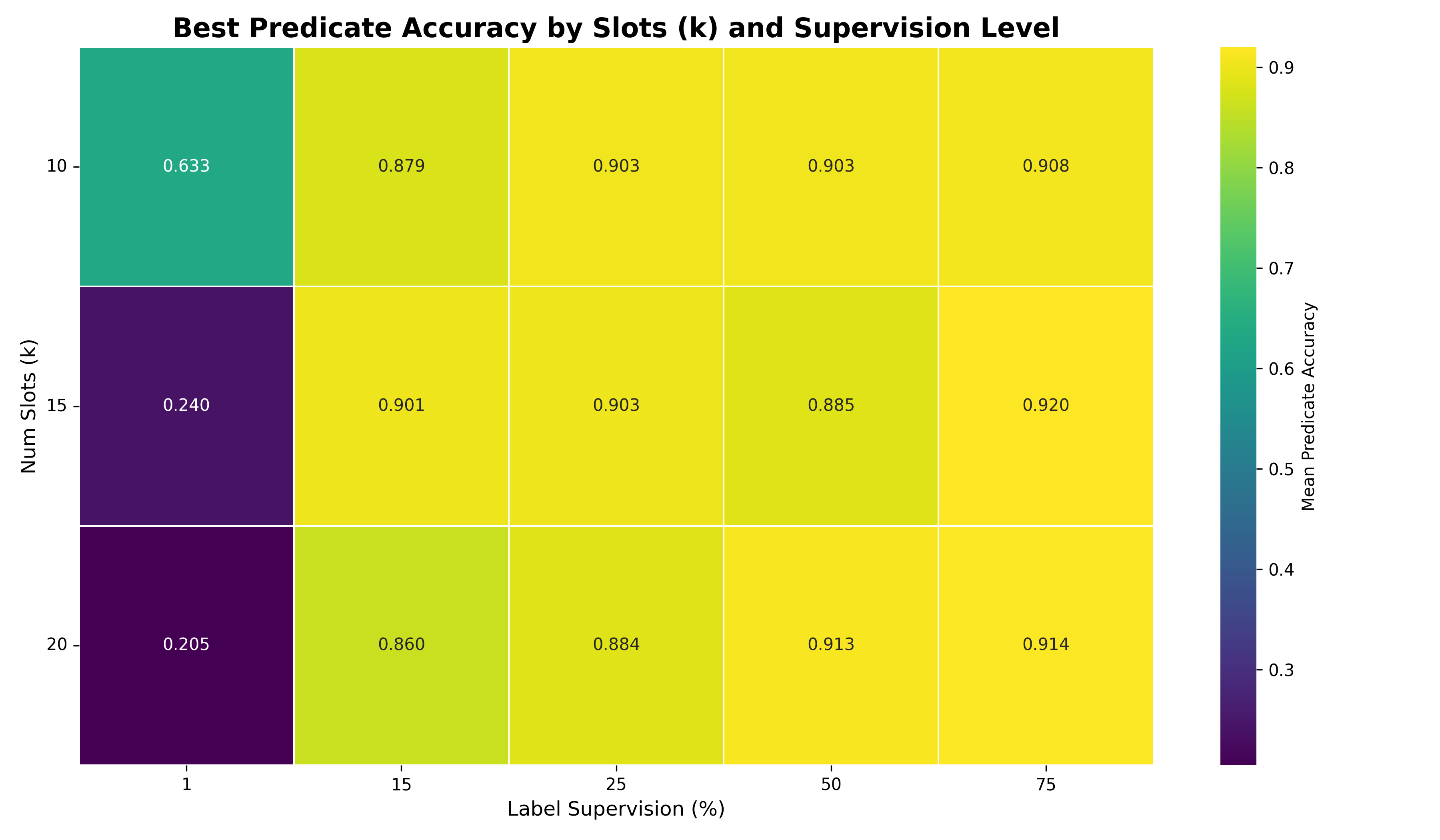}
    \caption{SLOTVAE ON CLEVR 100 IMAGES vs No. of slots ,for 25 model in total 5 models for each label supervision  }
    \label{fig:Clevr_SLOT_100}
\end{figure}

\textbf{Summary:}
For the multi-object CLEVR environment, providing the minimum correct structural inductive bias ($k=10$) is critical for stability and performance under \textbf{minimal supervision}. However, as the supervisory signal is strengthened (e.g., $15\%$ and above), increasing the slot number slightly beyond the scene maximum (e.g., $k=15$) results in the highest overall accuracy, demonstrating the trade-off between strict inductive bias and model capacity as a function of available supervision, and same can be said for 100 images as seen in Figure \ref{fig:Clevr_SLOT_100}.

\subsection{Weak Supervision Task on ILP Framework}
Here is the table \ref{tab:ilp_reasoning_performance_corrected} for RQ1   using 500 images divided in set of 100 pos and 400 neg examples 
\begin{table}[tb]
\centering
\caption{Impact of VAE Supervision Level on Downstream ILP Task Performance using 500 images [neg: 400, pos: 100].}
\label{tab:ilp_reasoning_performance_corrected}
\scriptsize % Use a smaller font
\resizebox{\columnwidth}{!}{%
\begin{tabular}{@{}llcccc@{}}
\toprule
\textbf{Rule} & \textbf{Difficulty} & \textbf{Supervision} & \textbf{Relevant Concept Acc.} & \textbf{ILP Iterations} & \textbf{Outcome} \\
\midrule
\multirow{3}{*}{\begin{tabular}[c]{@{}l@{}}"Contains a \\ yellow sphere"\end{tabular}} & \multirow{3}{*}{Easy} & 1\% & Shape: 71.4\%, Color: 21.8\% & 115 & Success \\
& & 25\% & Shape: 84.1\%, Color: 82.1\% & 115 & Success \\
& & 75\% & Shape: 89.4\%, Color: 89.2\% & 115 & Success \\
\midrule
\multirow{3}{*}{\begin{tabular}[c]{@{}l@{}}"Large blue sphere \& \\ small yellow sphere"\end{tabular}} & \multirow{3}{*}{Medium} & 1\% & Size: 88.1\%, Color: 21.8\% & 10,000+ & Fail (Timeout) \\
& & 25\% & Size: 97.5\%, Color: 82.1\% & 7,800 & Success \\
& & 75\% & Size: 97.9\%, Color: 89.2\% & 7,800 & Success \\
\midrule
\multirow{3}{*}{\begin{tabular}[c]{@{}l@{}}"Blue sphere in \\ front of yellow cube"\end{tabular}} & \multirow{3}{*}{Hard} & 1\% & Color: 21.8\%, Coord. MAE: 0.265 & $>$10,000+ & Fail (Timeout) \\
& & 25\% & Color: 82.1\%, Coord. MAE: 0.087 & $>$10,000+ & Fail (Timeout) \\
& & 75\% & Color: 89.4\%, Coord. MAE: 0.057 & 12,600 & Success \\
\bottomrule
\end{tabular}
}
\end{table}

\section{RQ2: Detailed Results on Domain Shift Robustness}

\subsection{Dimensionality Transfer: dSprites $\longleftrightarrow$ 3DShapes}
This bidirectional transfer assesses the generalization of fundamental geometric concepts across rendering styles and dimensional complexity ($2\texttt{D} \longleftrightarrow 3\texttt{D}$). While zero-shot transfer failed, full fine-tuning yielded effective bidirectional adaptation.

$\text{dSprites\cite{dsprites17}} \longrightarrow \text{3DShapes\cite{3dshapes18}}$ (2D to 3D)
The model trained on simple $2\texttt{D}$ sprites (dSprites) successfully adapted to the texture-rich $3\texttt{D}$ environment (3DShapes).

\begin{figure}[h!]
\centering
\includegraphics[width=0.5\linewidth]{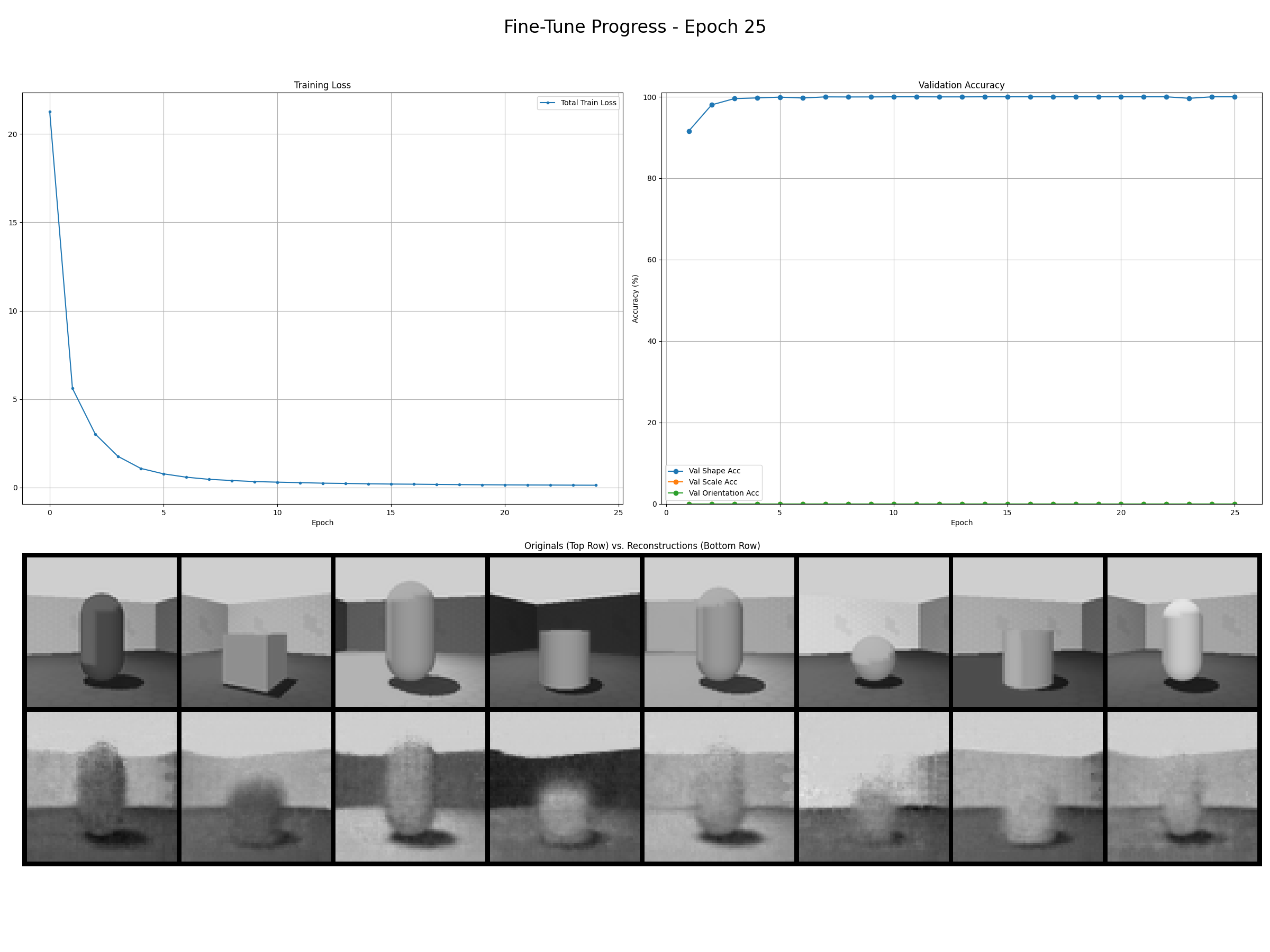}
\includegraphics[width=0.4\linewidth]{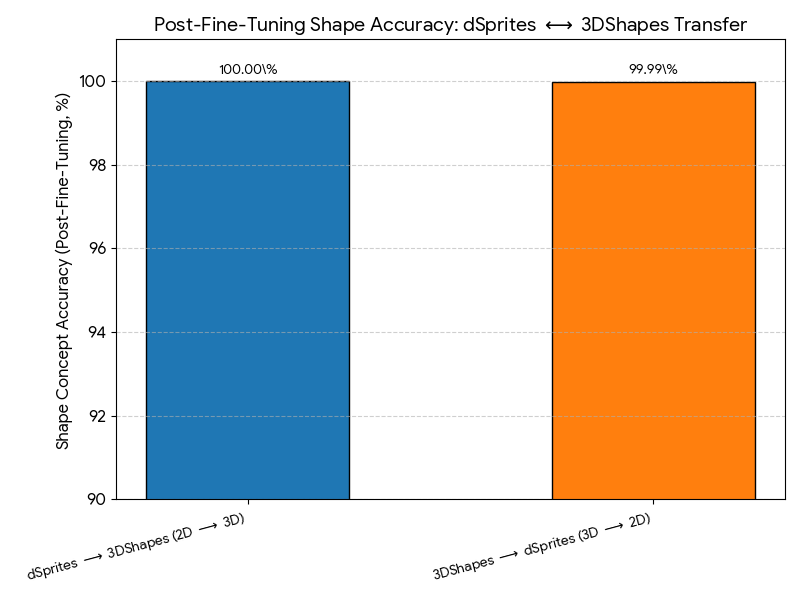}
\caption{Reconstruction of 3DShapes scenes by the fine-tuned dSprites model(left). Fine Tuned Shape Accuracy for Dsprites and 3Dshapes Domain transfer}
\label{fig:dsprites_to_3dshapes_finetune}
\end{figure}

The fine-tuned dSprites model achieved $\approx 99\%$ shape accuracy on 3DShapes. The successful reconstruction of $3\texttt{D}$ objects, including their shading (Figure \ref{fig:dsprites_to_3dshapes_finetune}), demonstrates that the foundational $2\texttt{D}$ geometric features learned are sufficiently abstract to represent simple $3\texttt{D}$ objects, validating the successful adaptation of structural concepts.

$\text{3DShapes} \longrightarrow \text{dSprites}$ (3D to 2D)
The model trained on $3\texttt{D}$ data successfully adapted to the simpler $2\texttt{D}$ domain.

\begin{figure}[h!]
\centering
\includegraphics[width=0.25\linewidth]{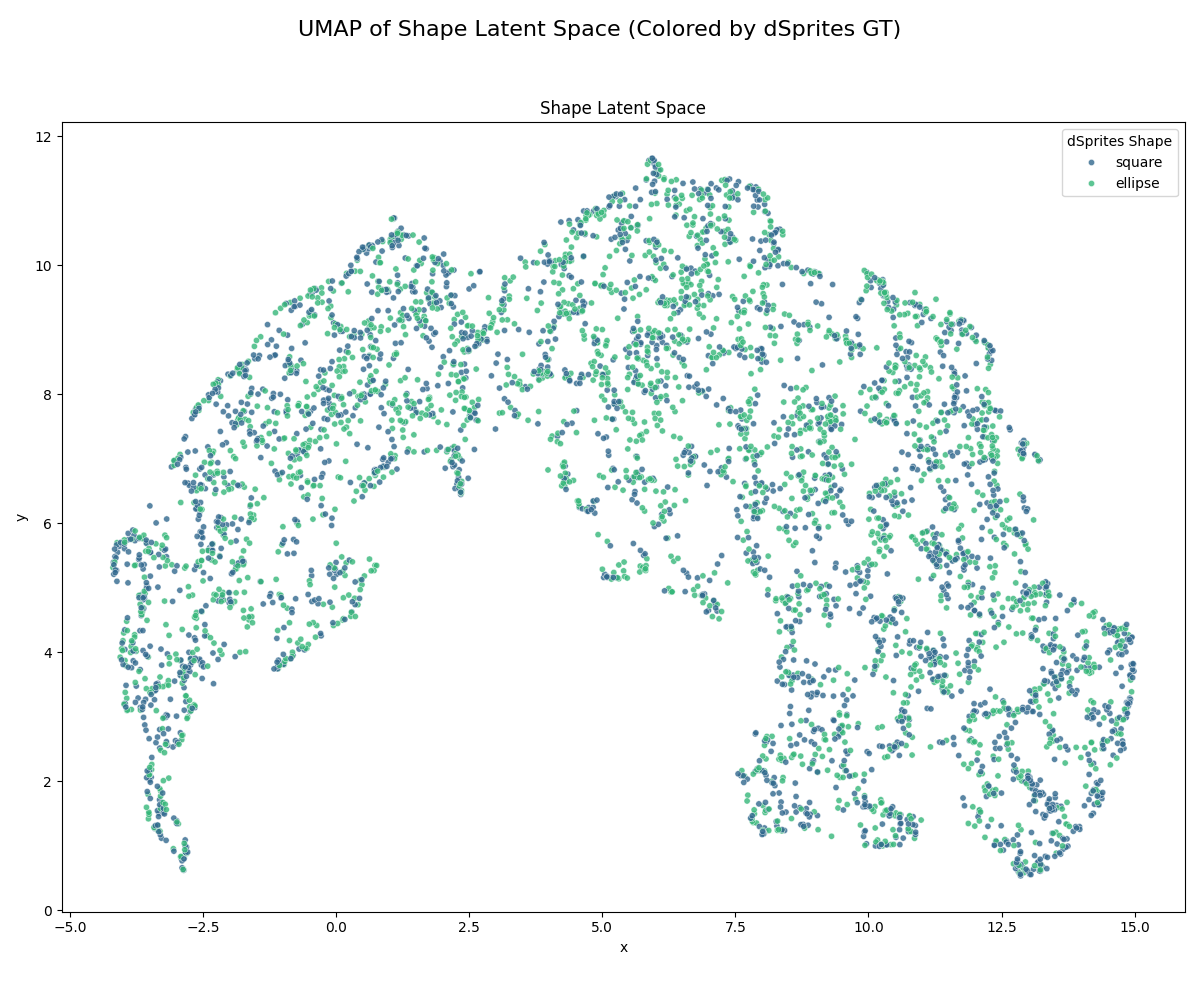}
\includegraphics[width=0.3\linewidth]{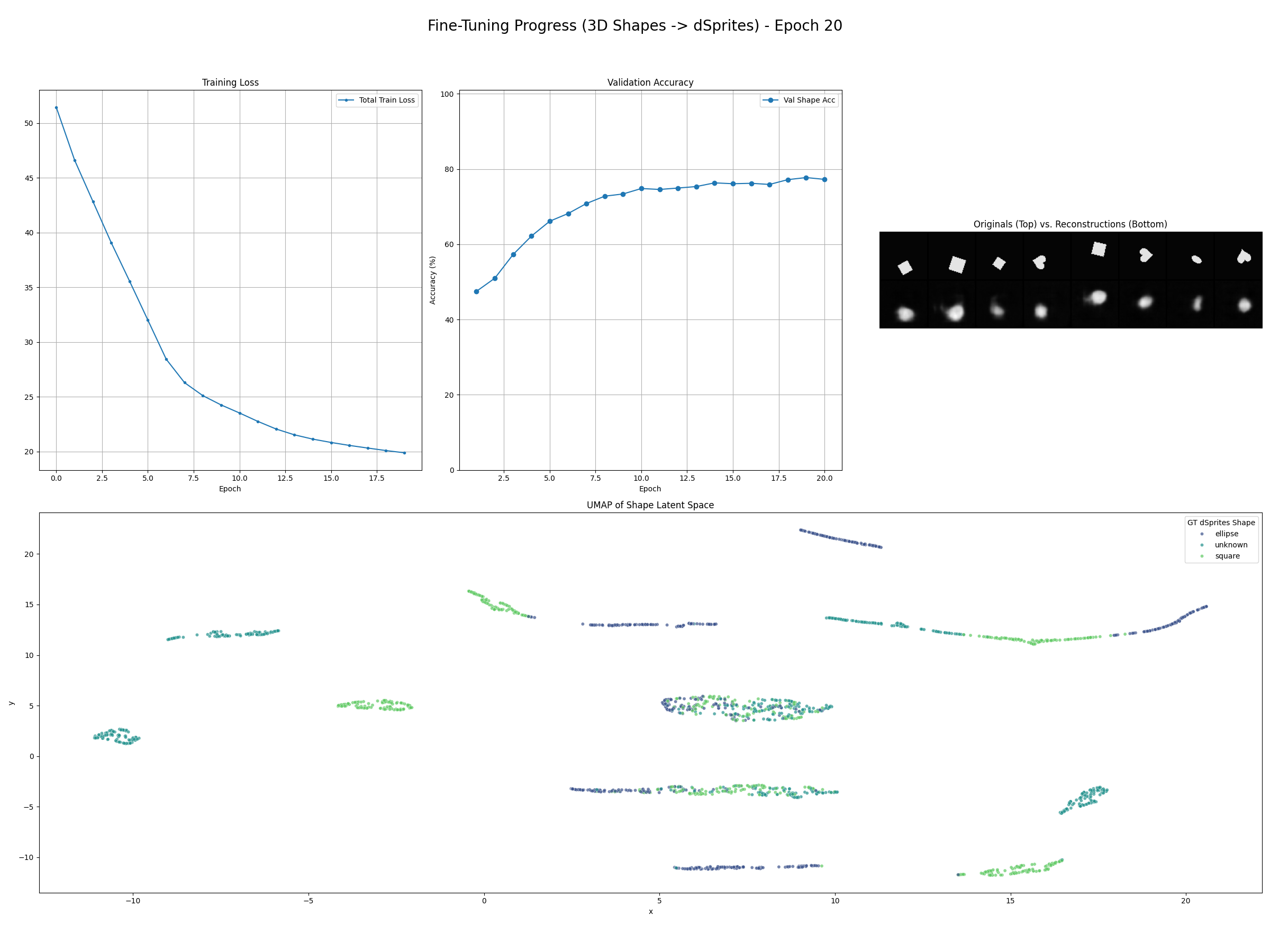}
\caption{Latent Space Disentanglement for $\text{3DShapes} \longrightarrow \text{dSprites}$. \textbf{Left}: Shape latent plot under zero-shot transfer. \textbf{Right}: Reconstruction plot after fine-tuning. Full fine-tuning resulted in a well-disentangled latent space for shape, confirming robust feature generalization across dimensional shifts.}
\label{fig:3dshapes_to_dsprites_finetune}
\end{figure}

The reverse transfer also produced a well-disentangled latent space for shape after fine-tuning (Figure \ref{fig:3dshapes_to_dsprites_finetune}, Right), confirming that representations for foundational concepts like geometry are robust to dimensionality and rendering style changes.

\subsection{Information Loss Transfer: CLEVR $\longleftrightarrow$ CLEVR-2D}
This shift, which involves transferring between a complex $3\texttt{D}$ domain (CLEVR) and a flat $2\texttt{D}$ version (CLEVR-2D), highlights the fragility of the perception module against the loss of salient visual information.

\subsubsection*{$\text{CLEVR} \longrightarrow \text{CLEVR-2D}$ (3D to 2D)}
The transfer from the information-rich $3\texttt{D}$ domain to the information-poor $2\texttt{D}$ domain proved destructive.

\begin{figure}[h!]
    \centering
    \includegraphics[width=0.45\linewidth]{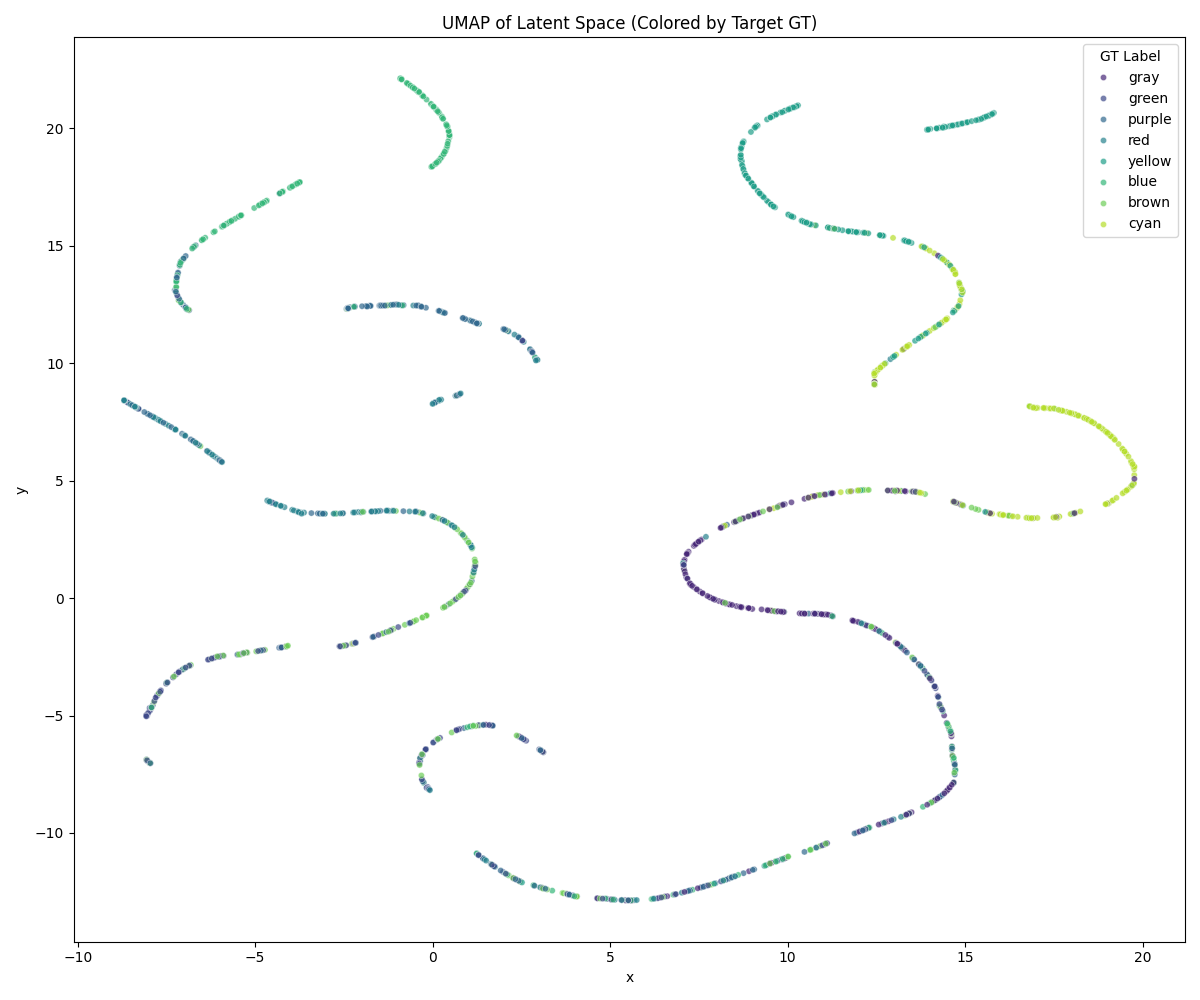}
    \includegraphics[width=0.45\linewidth]{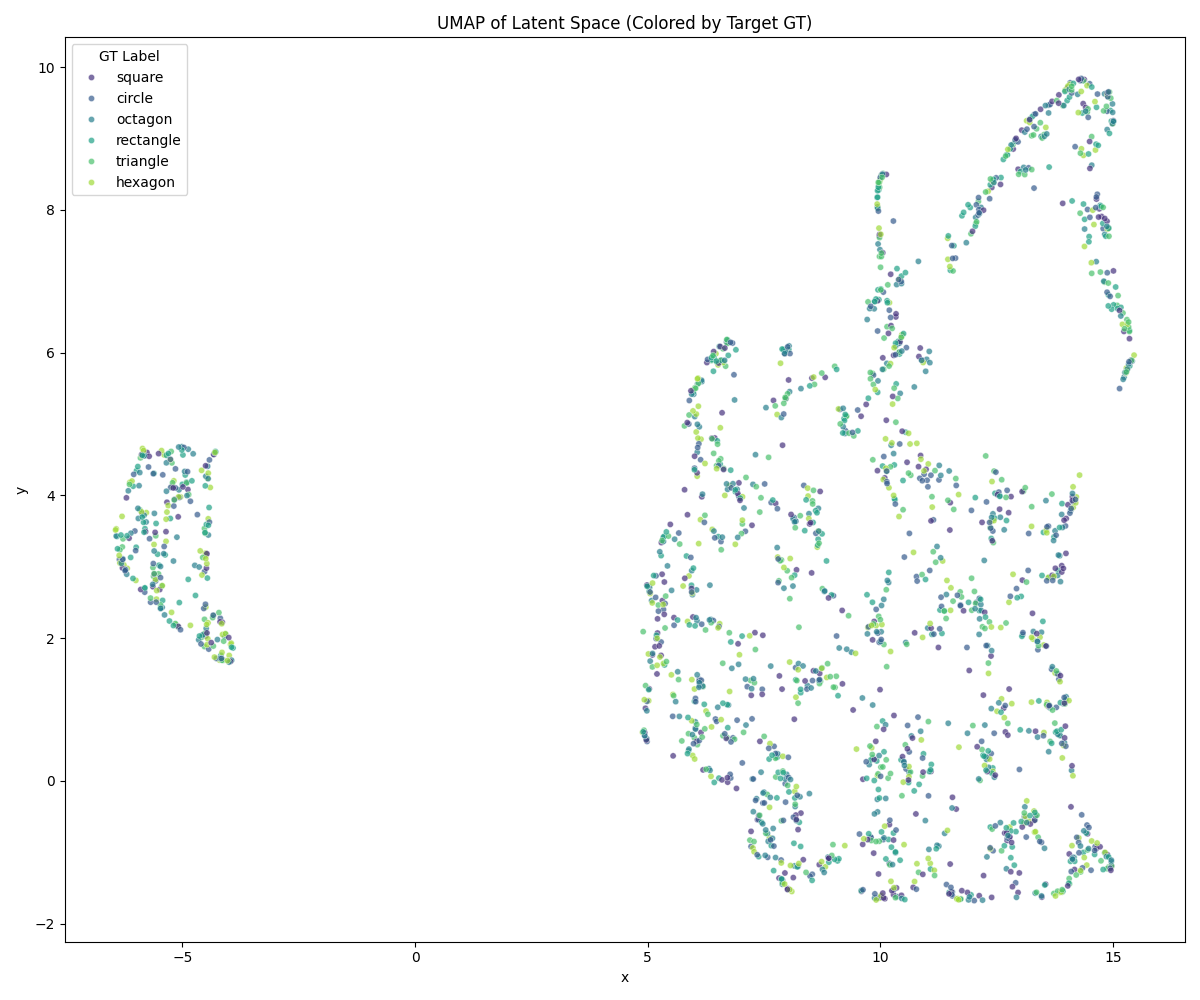}
    \includegraphics[width=0.45\linewidth]{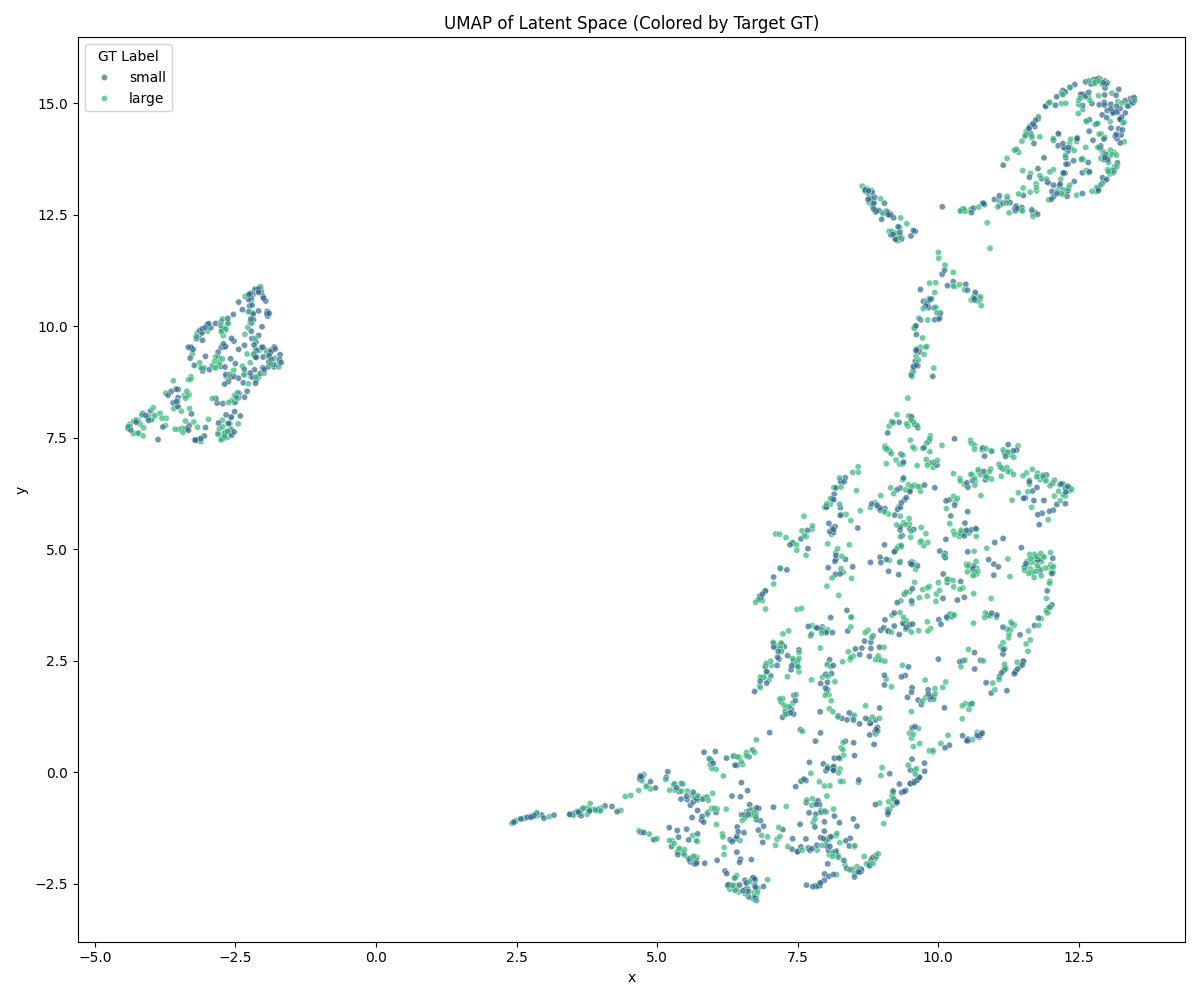}
    \caption{Zero-Shot Latent Space for $\text{CLEVR} \longrightarrow \text{CLEVR-2D}$. UMAP plots of latent dimensions colored by \textbf{Color, Shape, and Size} (Left to Right). The encoder, over-specialized to $3\texttt{D}$ features like shading, suffers \textbf{complete model collapse}.}
    \label{fig:clevr_to_clevr2d_zeroshot}
\end{figure}

\begin{figure}
    \centering
    \includegraphics[width=0.45\linewidth]{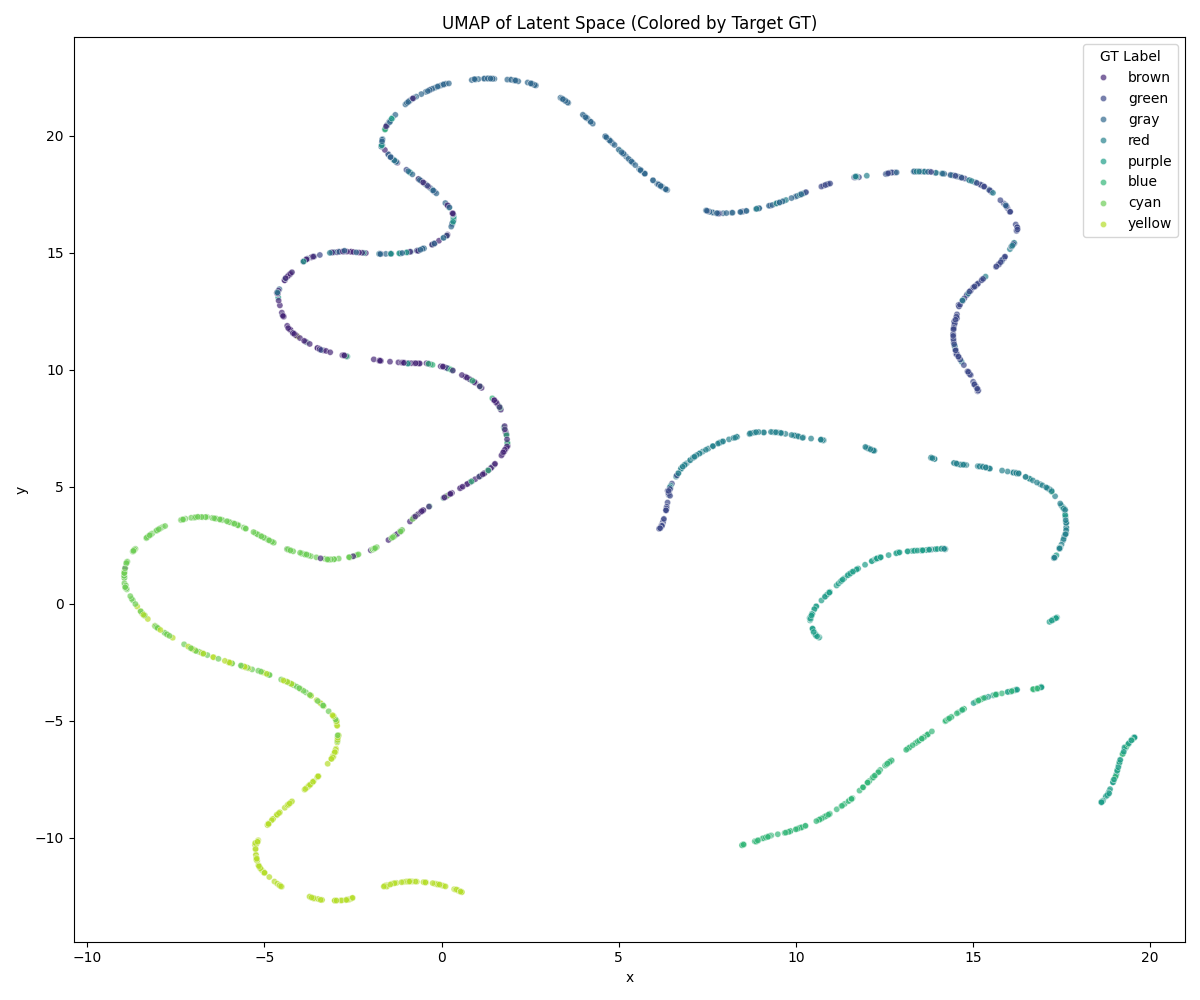}
    \includegraphics[width=0.45\linewidth]{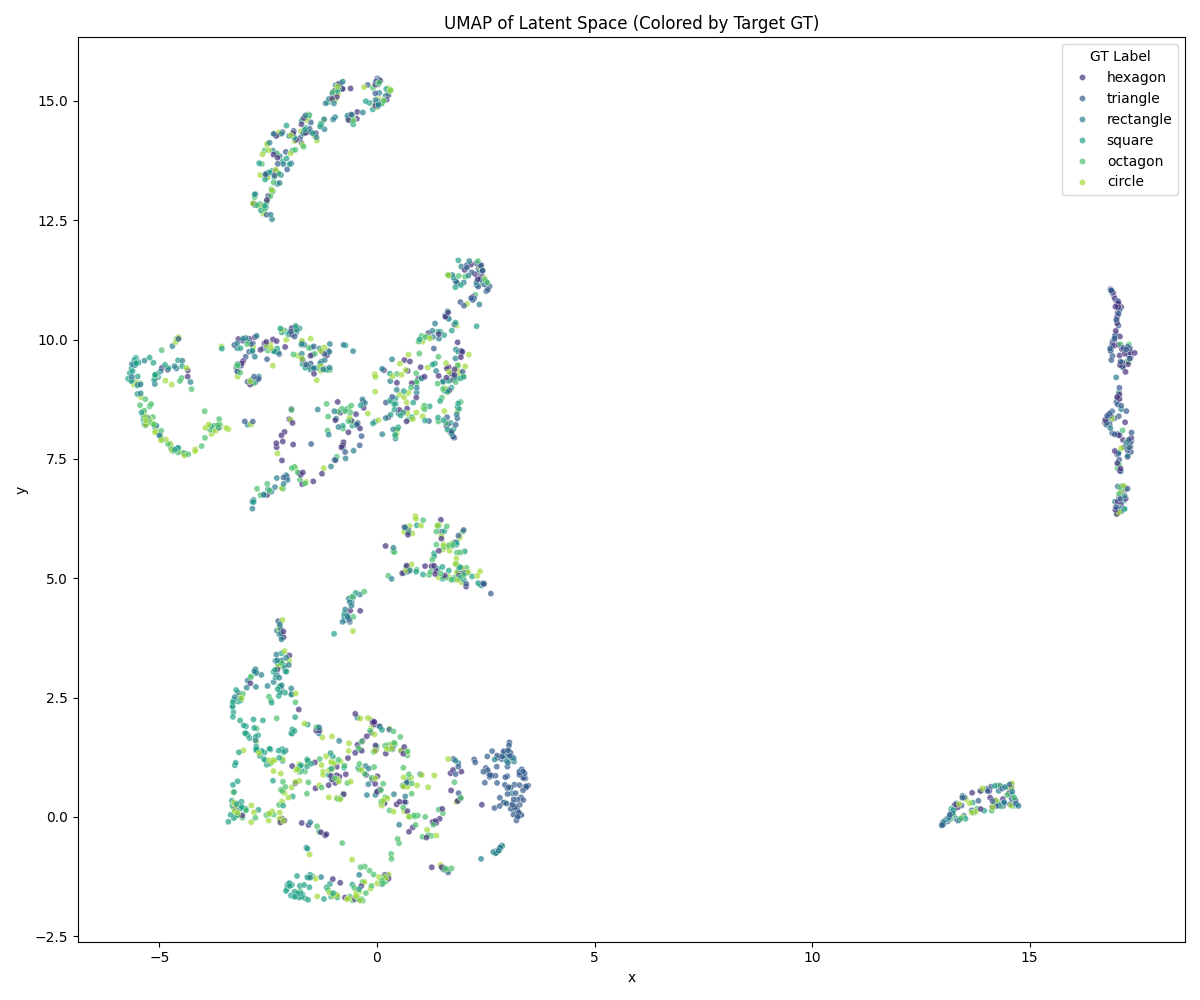}
    \includegraphics[width=0.45\linewidth]{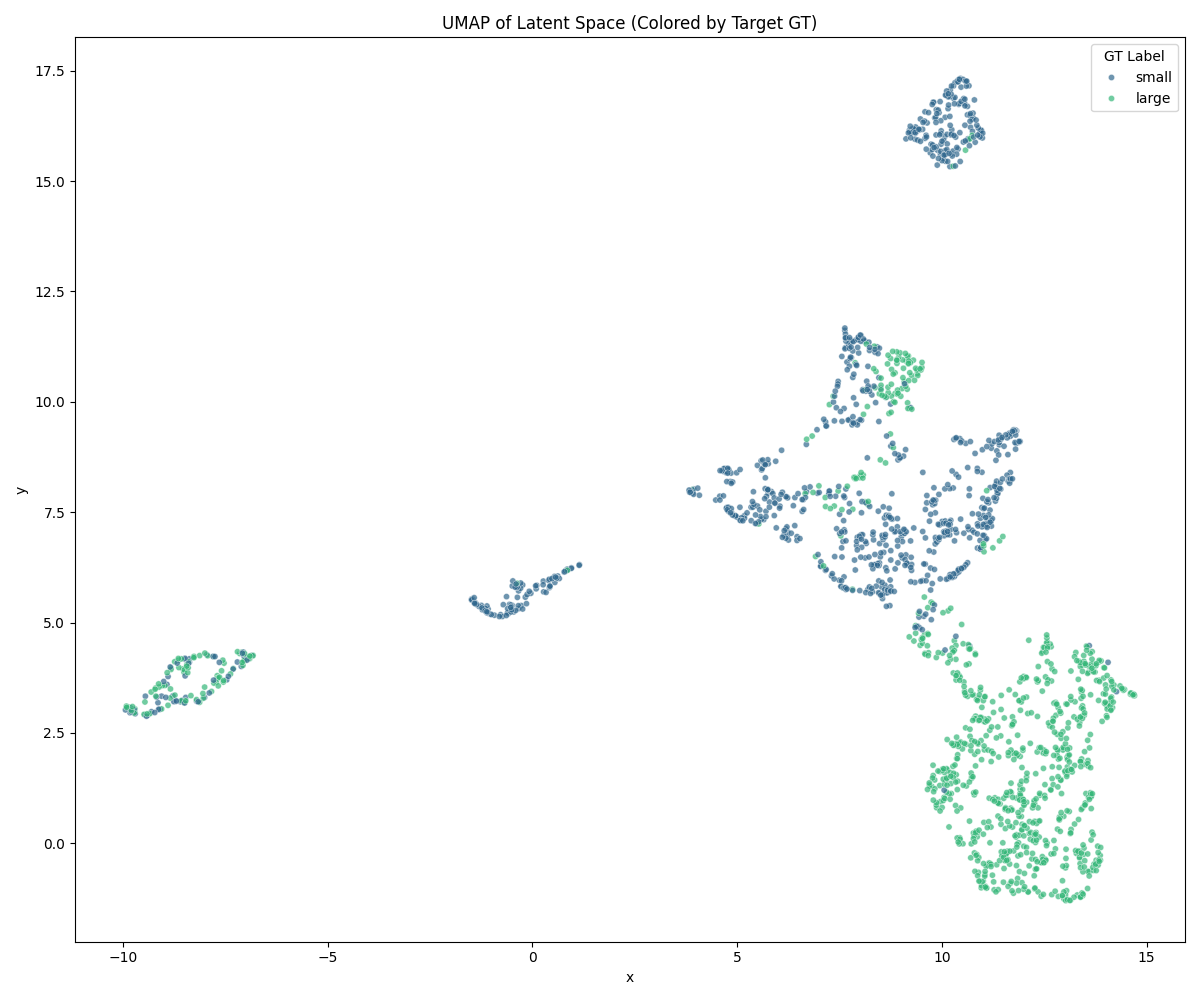}
    \includegraphics[width=0.75\linewidth]{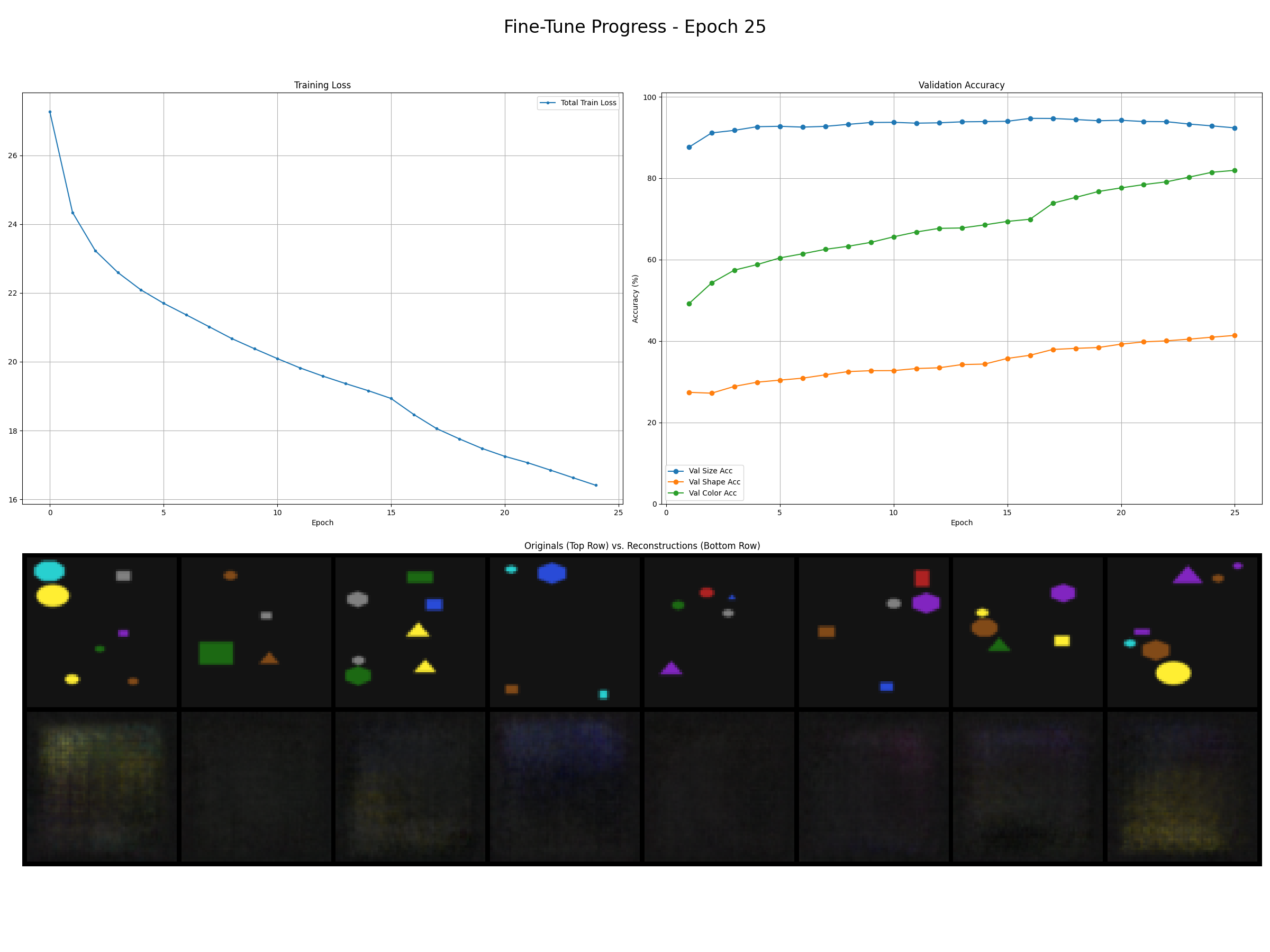}
    \caption{Fine-Tuned Latent Space for $\text{CLEVR} \longrightarrow \text{CLEVR-2D}$ (Color, Shape, Size (Left to right)), along with the reconstruction results .}
    \label{fig:clevr_to_clevr2d_fine_tune}
\end{figure}

The $3\texttt{D} \longrightarrow 2\texttt{D}$ transfer resulted in complete model collapse (Figure \ref{fig:clevr_to_clevr2d_zeroshot}), as the VAE's encoder was over-specialized to $3\texttt{D}$ features that were entirely absent in the target $2\texttt{D}$ images. Even full fine-tuning (Figure \ref{fig:clevr_to_clevr2d_fine_tune}  did not salvage the representation.

\subsubsection*{$\text{CLEVR-2D} \longrightarrow \text{CLEVR}$ (2D to 3D)}
The reverse transfer showed clear adaptation success.

\begin{figure}[h!]
\centering
\includegraphics[width=0.33\linewidth]{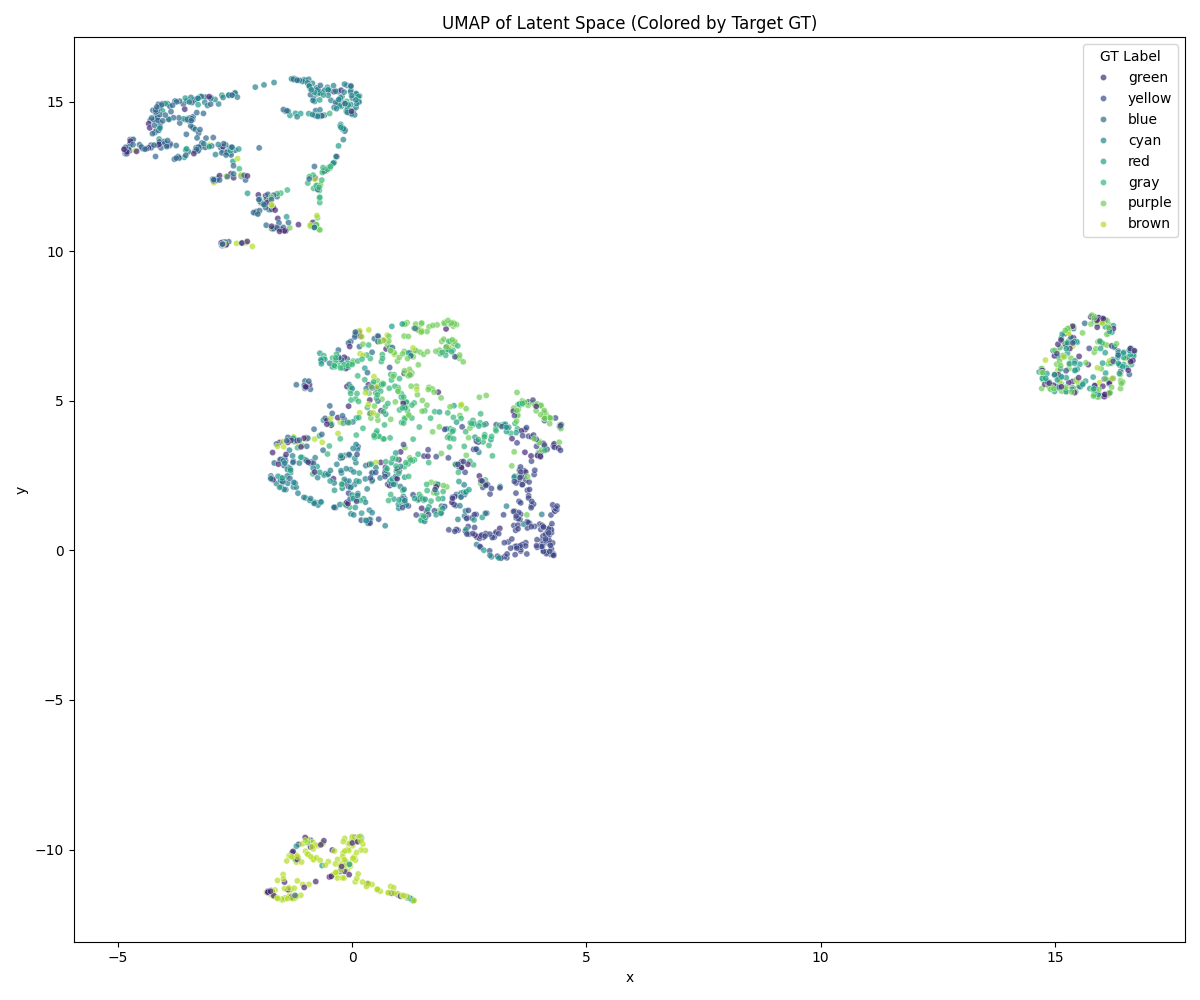}
\includegraphics[width=0.33\linewidth]{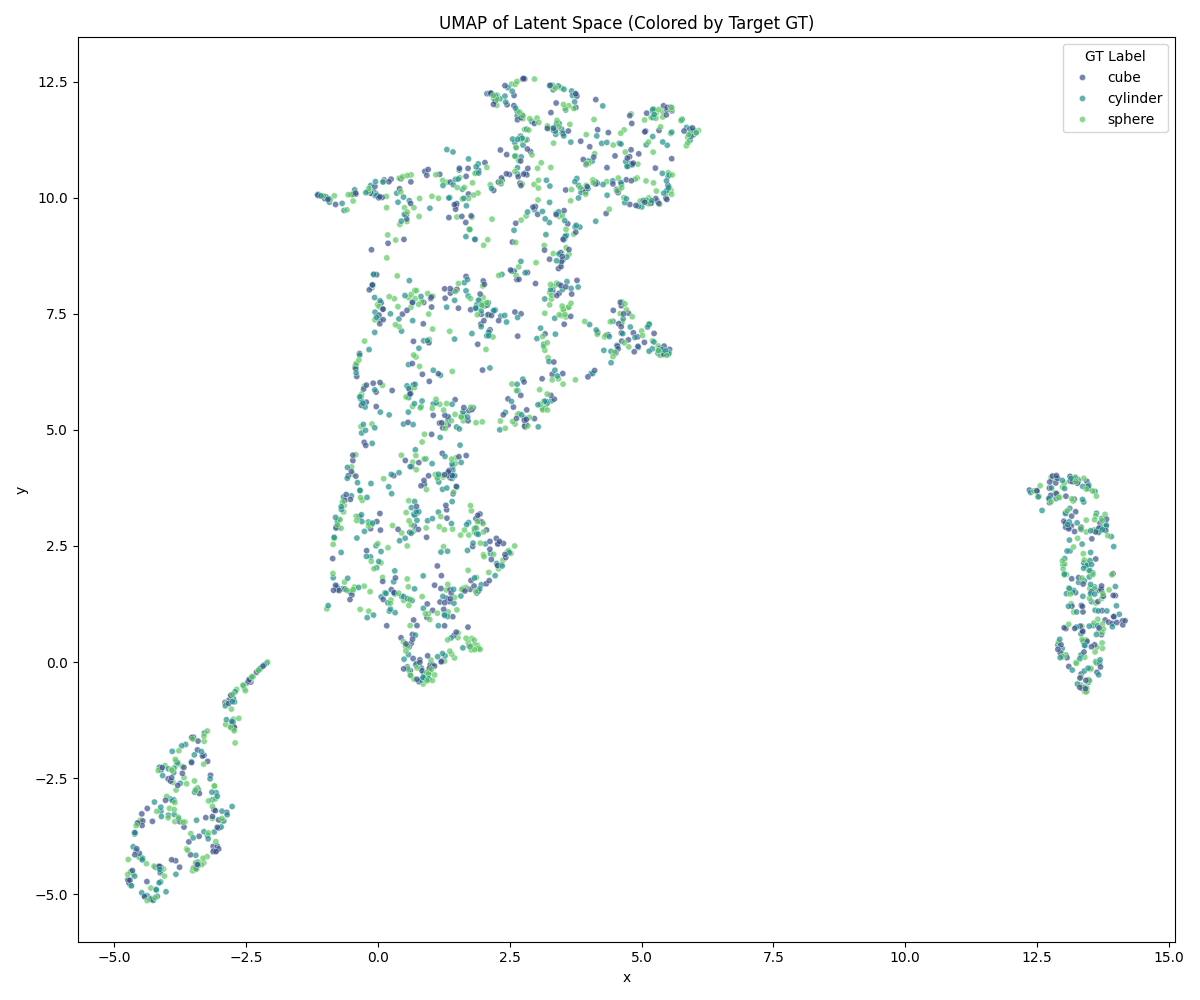}
\includegraphics[width=0.33\linewidth]{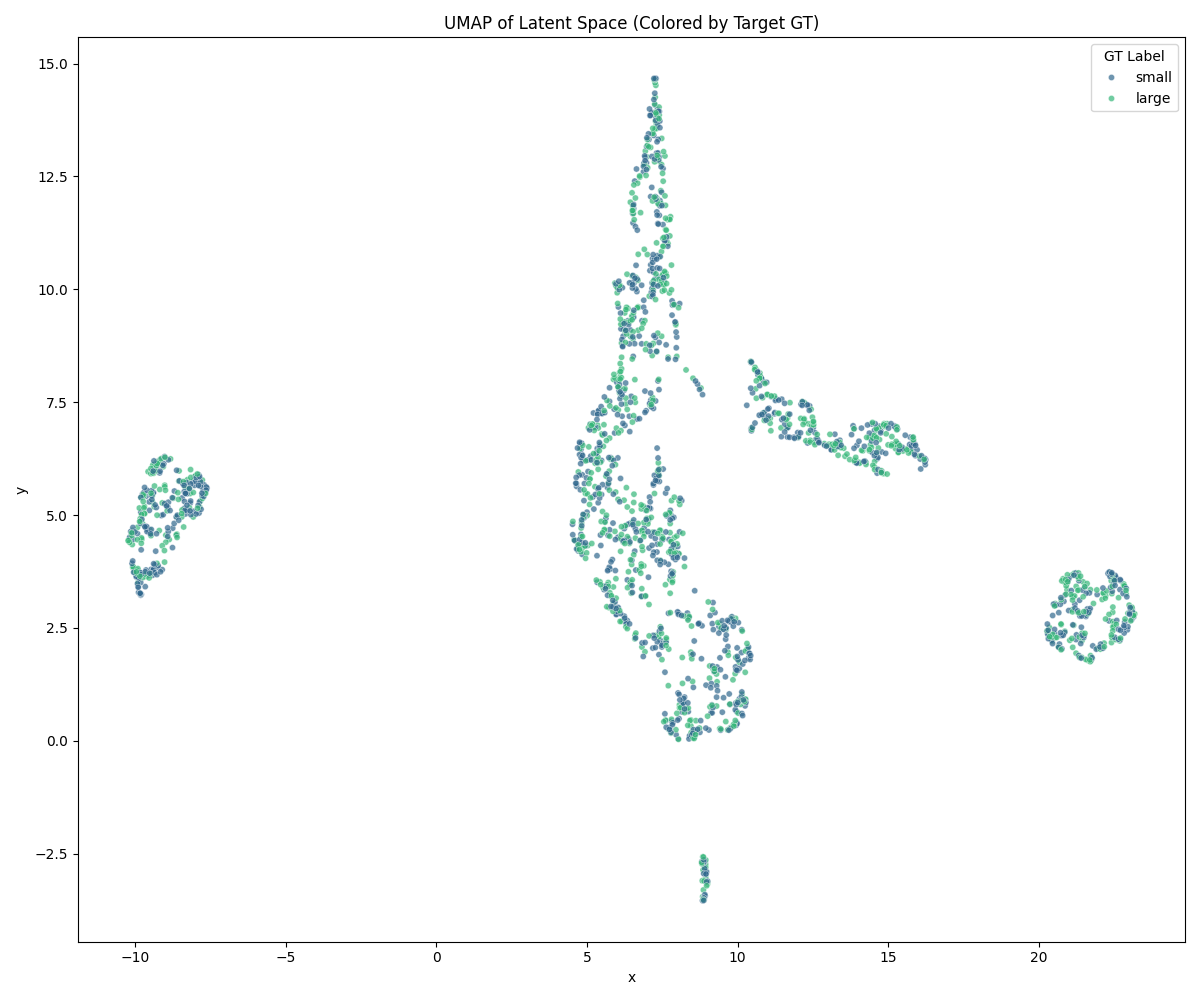}
\caption{Zero-Shot Latent Space for $\text{CLEVR-2D} \longrightarrow \text{CLEVR}$ (Color, Shape, Size). The latent spaces show poor separation under zero-shot transfer especially for concepts like shape and size.}
\label{fig:clevr2d_to_clevr_zeroshot}
\end{figure}

While the initial zero-shot performance was ineffective (Figure \ref{fig:clevr2d_to_clevr_zeroshot}), full fine-tuning allowed the model to learn robust representations.

\begin{figure}[h!]
\centering
\includegraphics[width=0.33\linewidth]{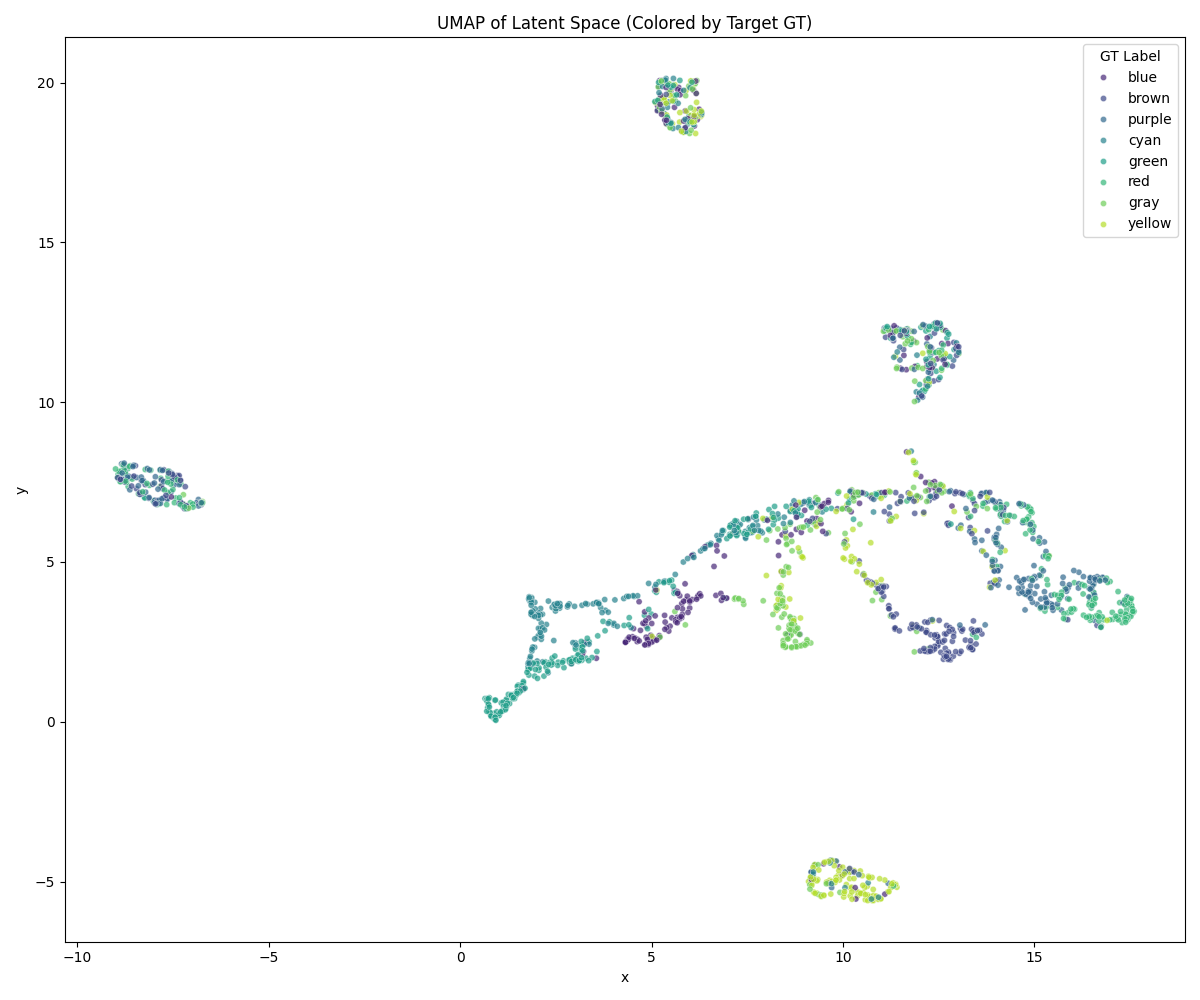}
\includegraphics[width=0.33\linewidth]{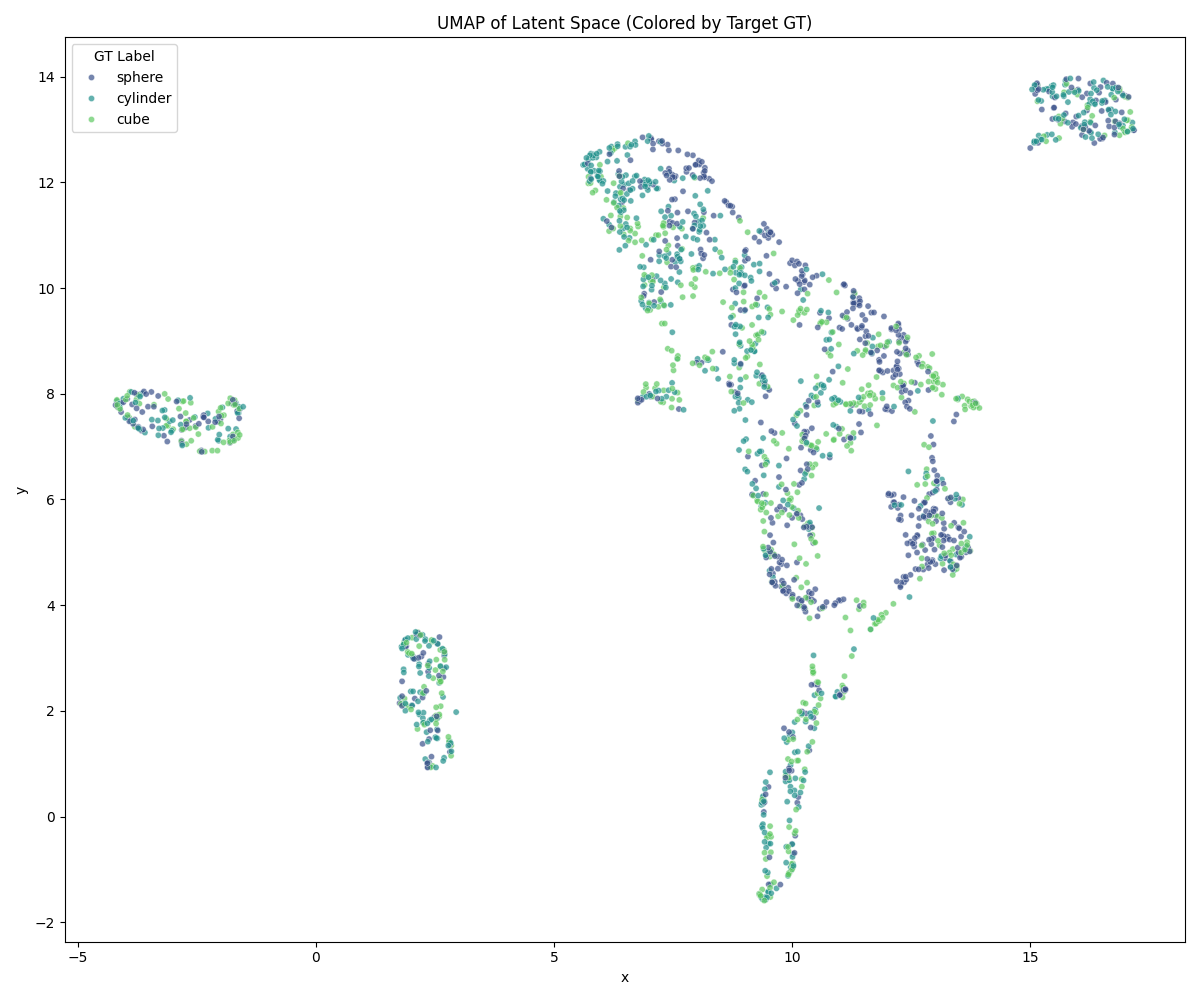}
\includegraphics[width=0.33\linewidth]{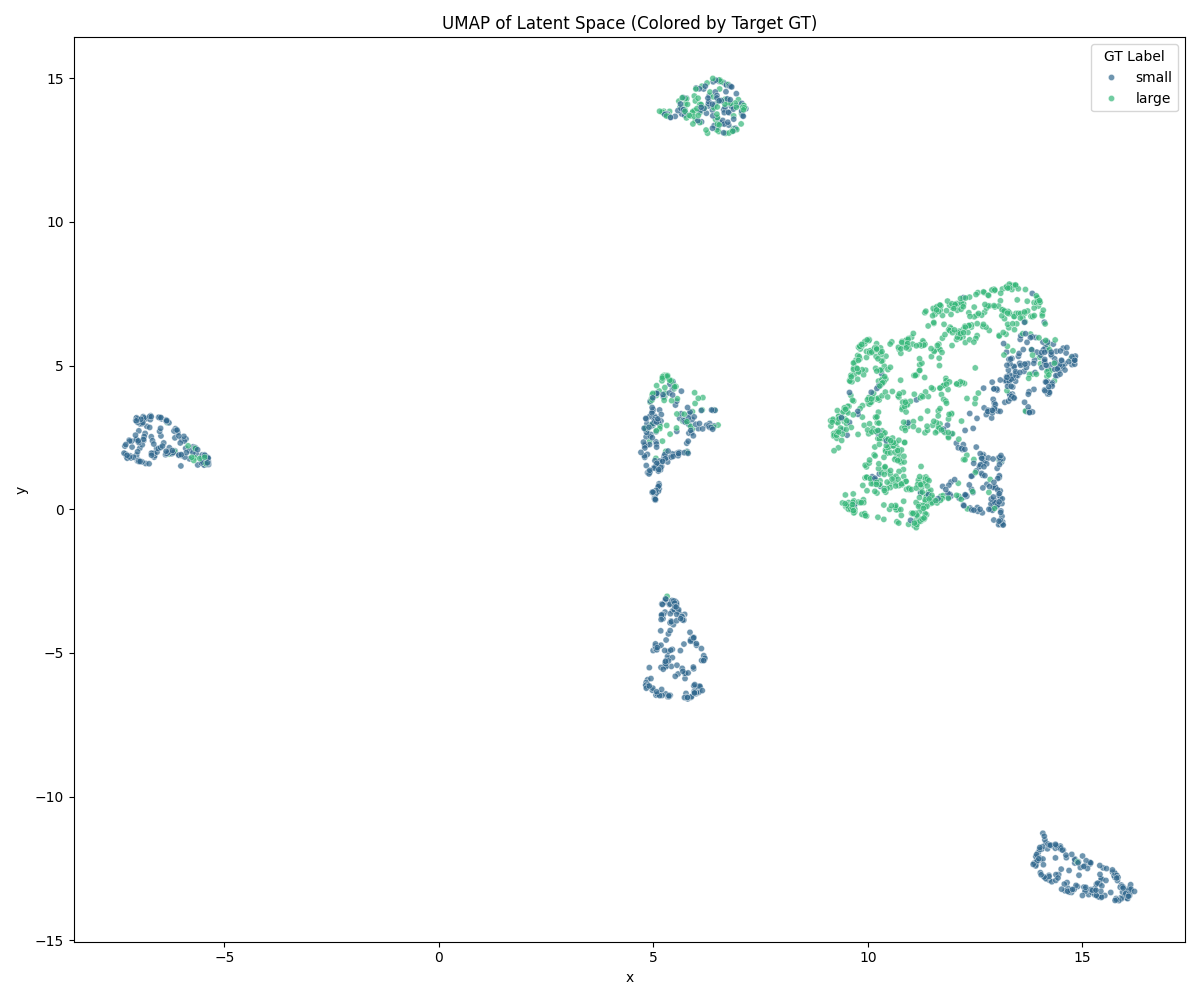}
\caption{Fine-Tuned Latent Space for $\text{CLEVR-2D} \longrightarrow \text{CLEVR}$ (Color, Shape, Size). After fine-tuning, the latent space shows clear, well-separated clusters for $\textbf{shape}$ and $\textbf{size}$, confirming successful disentanglement of foundational structural concepts.}
\label{fig:clevr2d_to_clevr_finetune}
\end{figure}

\begin{figure}
    \centering
    \includegraphics[width=0.49\linewidth]{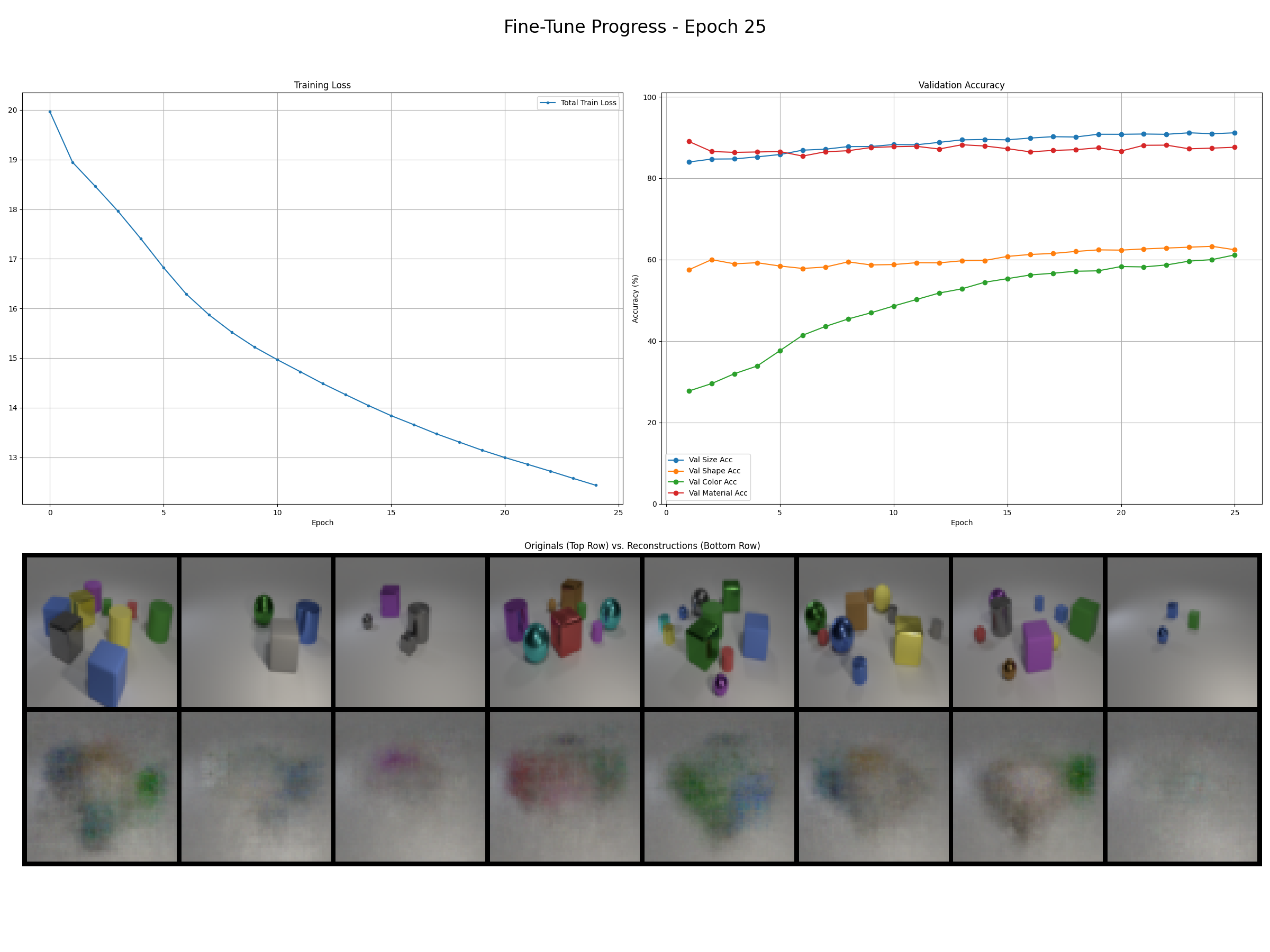}
    \includegraphics[width=0.49\linewidth]{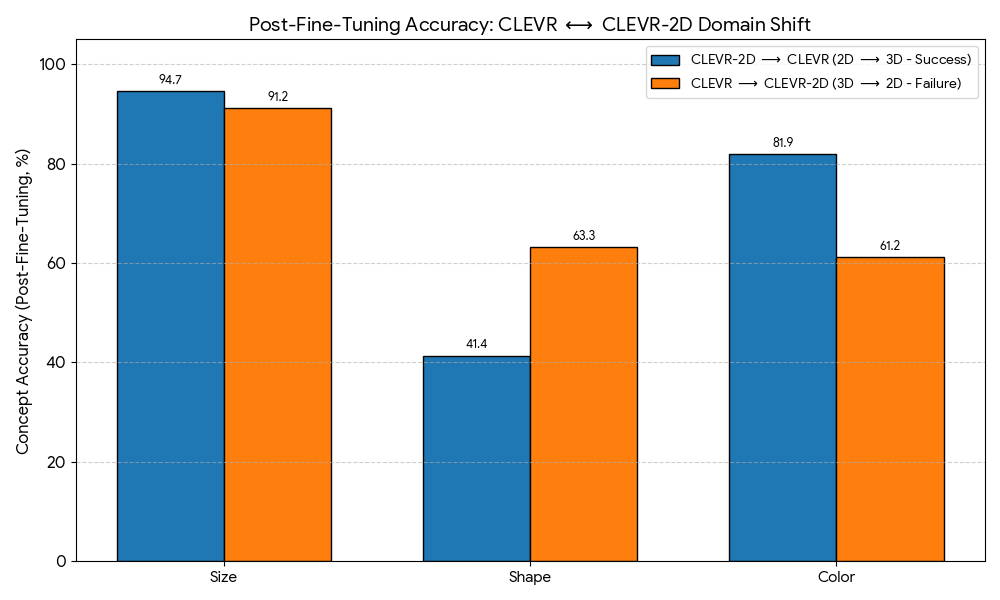}
    \caption{FINE TUNE for 25 EPOCHS of CLEVR2D  on CLEVR (left), concept accuracy across the domains(right)}
    \label{fig:clevr2d_to_clevr_finetune_1}
\end{figure}

Features for structural concepts ($\textbf{shape}$, $\textbf{size}$) learned in the information-poor ($2\texttt{D}$) domain are transferable and successfully adapted to the more complex ($3\texttt{D}$) domain (Figure \ref{fig:clevr2d_to_clevr_finetune}  and  \ref{fig:clevr2d_to_clevr_finetune_1}). The color concept remained more challenging to separate, showing partially entangled clusters.

\subsection{Surface Appearance Transfer: CLEVR $\longleftrightarrow$ CLEVR-Tex}
This shift investigates the model's robustness against changes in surface appearance, specifically the introduction of visually complex textures.

\subsubsection*{$\text{CLEVR} \longrightarrow \text{CLEVR-Tex\cite{karazija2021clevrtextexturerichbenchmarkunsupervised}}$ (Simple to Complex)}
\begin{figure}[h!]
\centering
\includegraphics[width=0.65\linewidth]{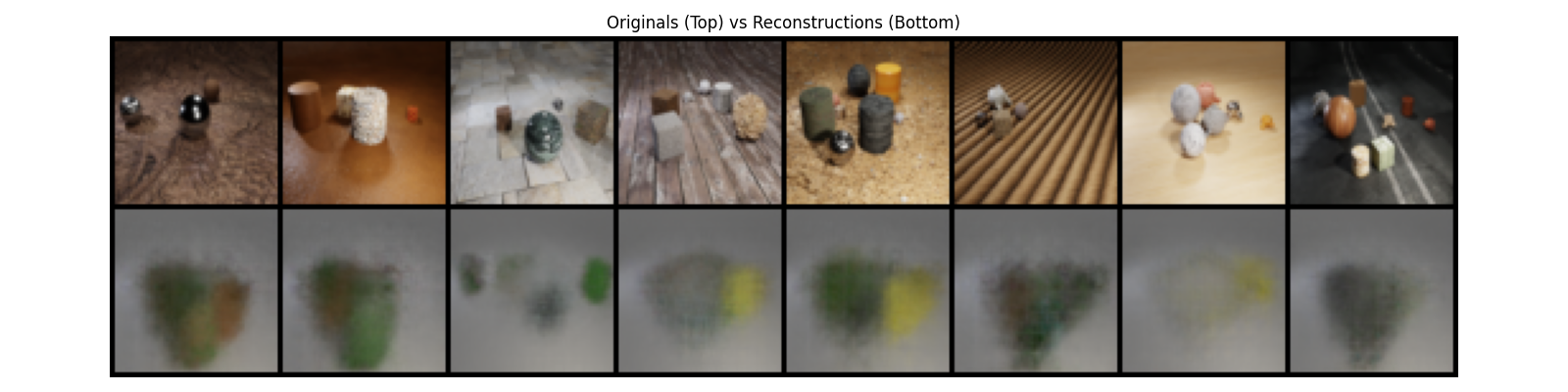}
\caption{Zero-Shot Reconstruction for $\text{CLEVR} \longrightarrow \text{CLEVR-Tex}$. The model attempts reconstruction on textured data.}
\label{fig:clevr_to_tex_zeroshot_viz}
\end{figure}

This transfer (uniform color to complex texture) as seen in Figure \ref{fig:clevr_to_tex_zeroshot_viz} revealed a conceptual dichotomy.

\begin{figure}[h!]
\centering
\includegraphics[width=0.45\linewidth]{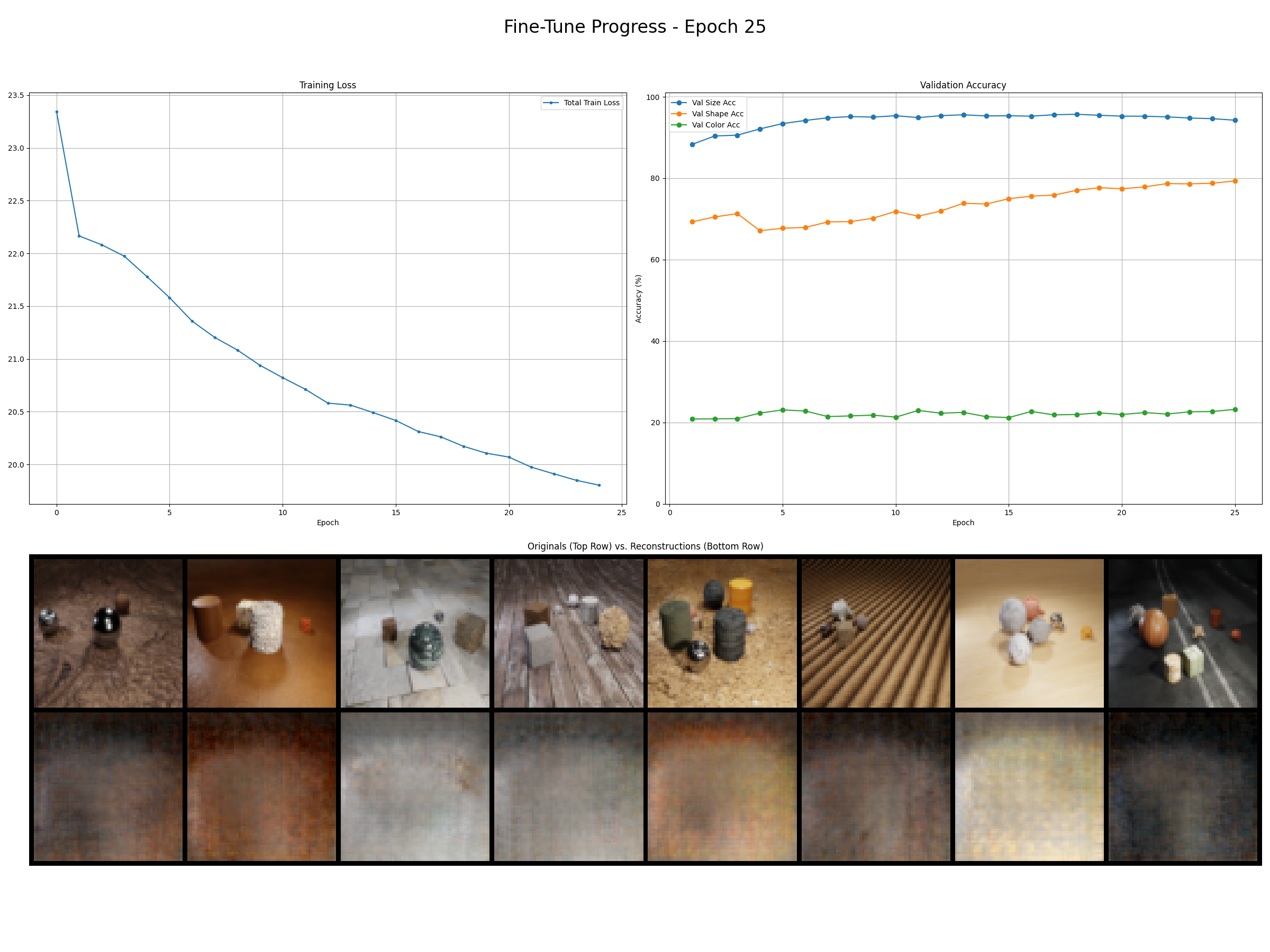}
\caption{Fine-Tuning Reconstruction for $\text{CLEVR} \longrightarrow \text{CLEVR-Tex}$. Even a fine-tuned model unable to adapt its reconstruction to the textured domain.}
\label{fig:clevr_to_tex_finetune_viz}
\end{figure}

Post-finetuning, structural concepts ($\textbf{shape}: 91\%$; $\textbf{size}: 98.5\%$) adapted successfully, but surface-level concepts like color failed drastically, achieving only $22\%$ accuracy. This indicates that the uniform color bias in the source domain was insufficient for robust color generalization as seen in \ref{fig:clevr_to_tex_finetune_viz}

\subsubsection*{$\text{CLEVR-Tex} \longrightarrow \text{CLEVR}$ (Complex to Simple)}
\begin{figure}[h!]
\centering
\includegraphics[width=0.3\linewidth]{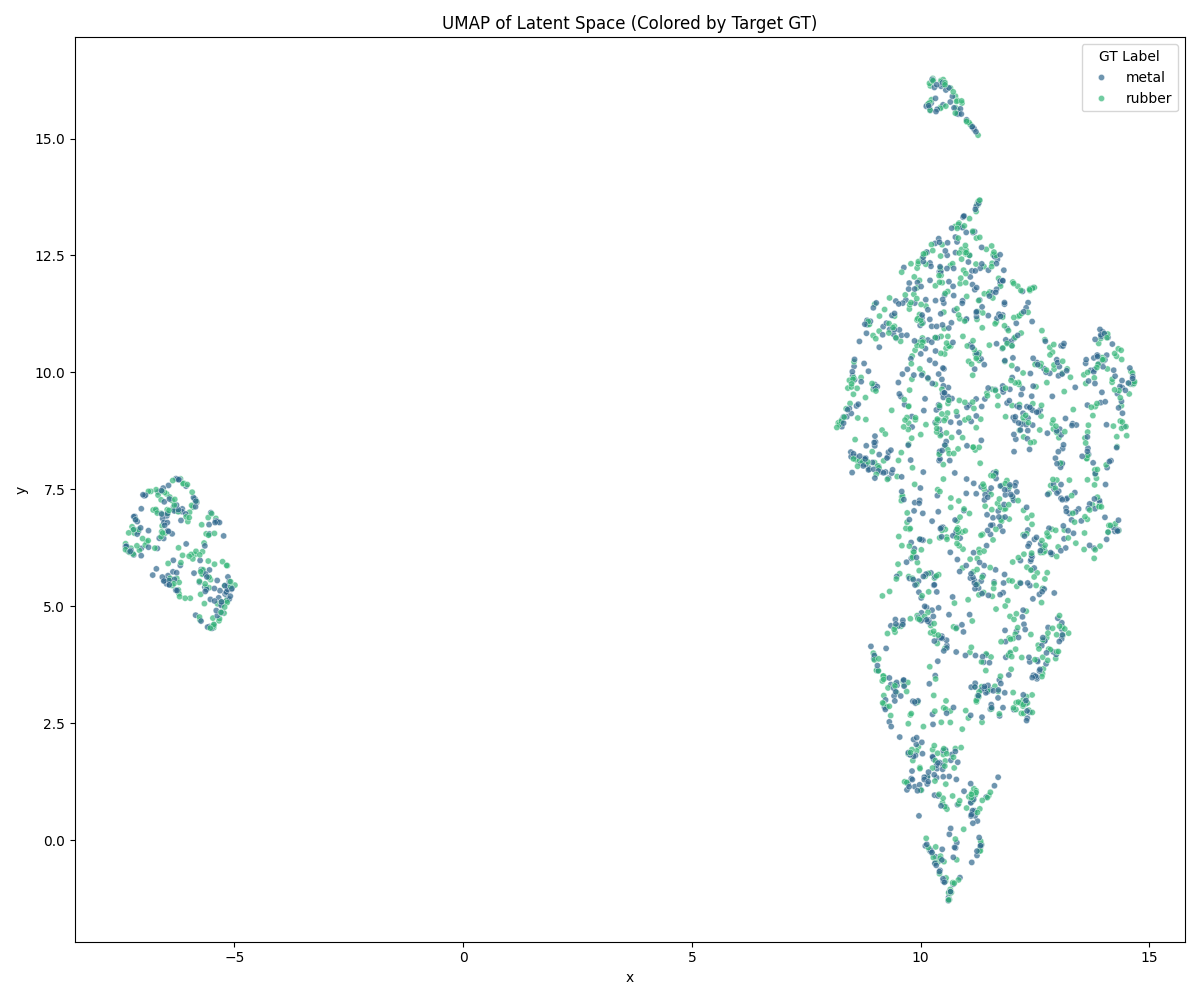}
\includegraphics[width=0.3\linewidth]{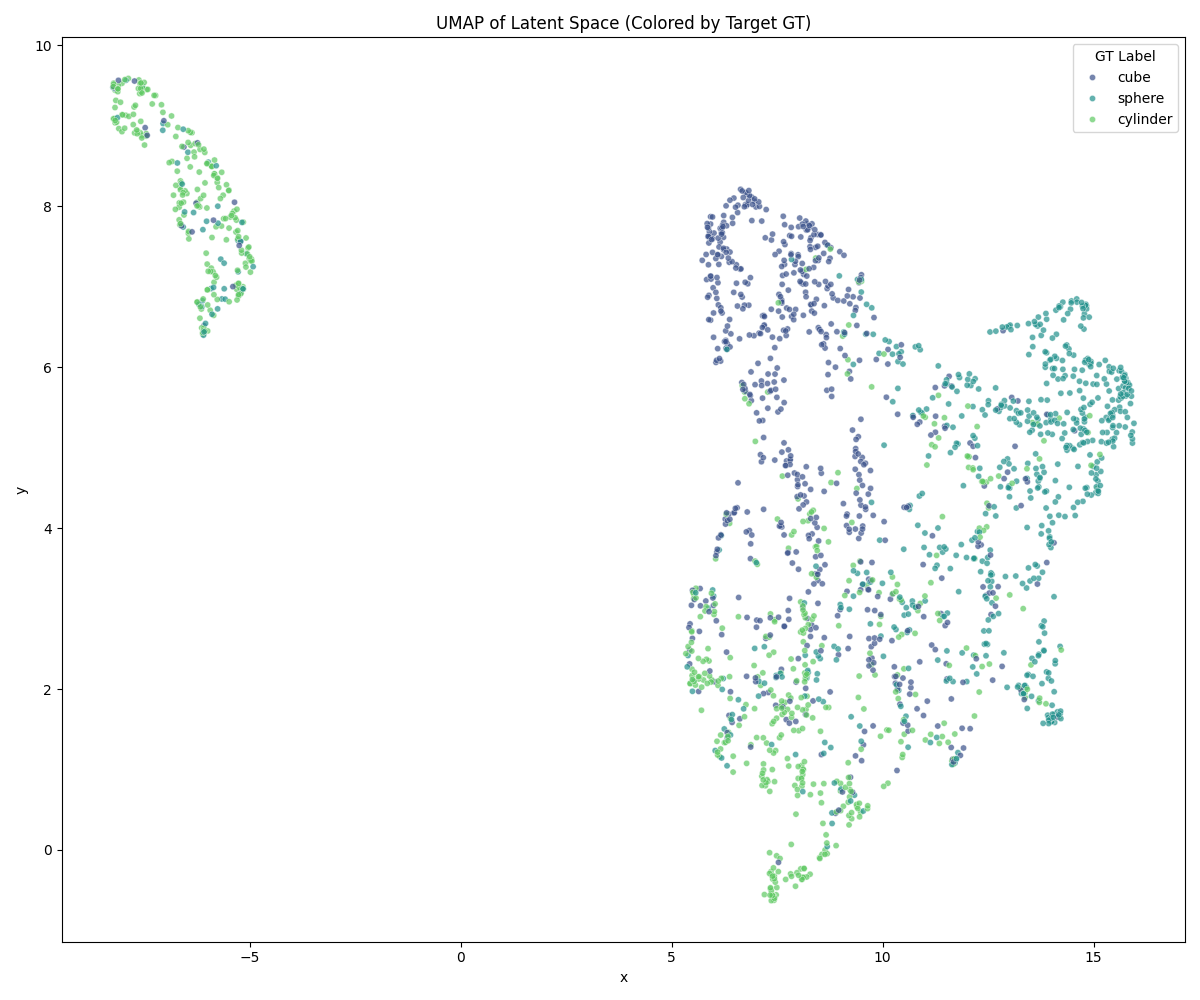}
\includegraphics[width=0.3\linewidth]{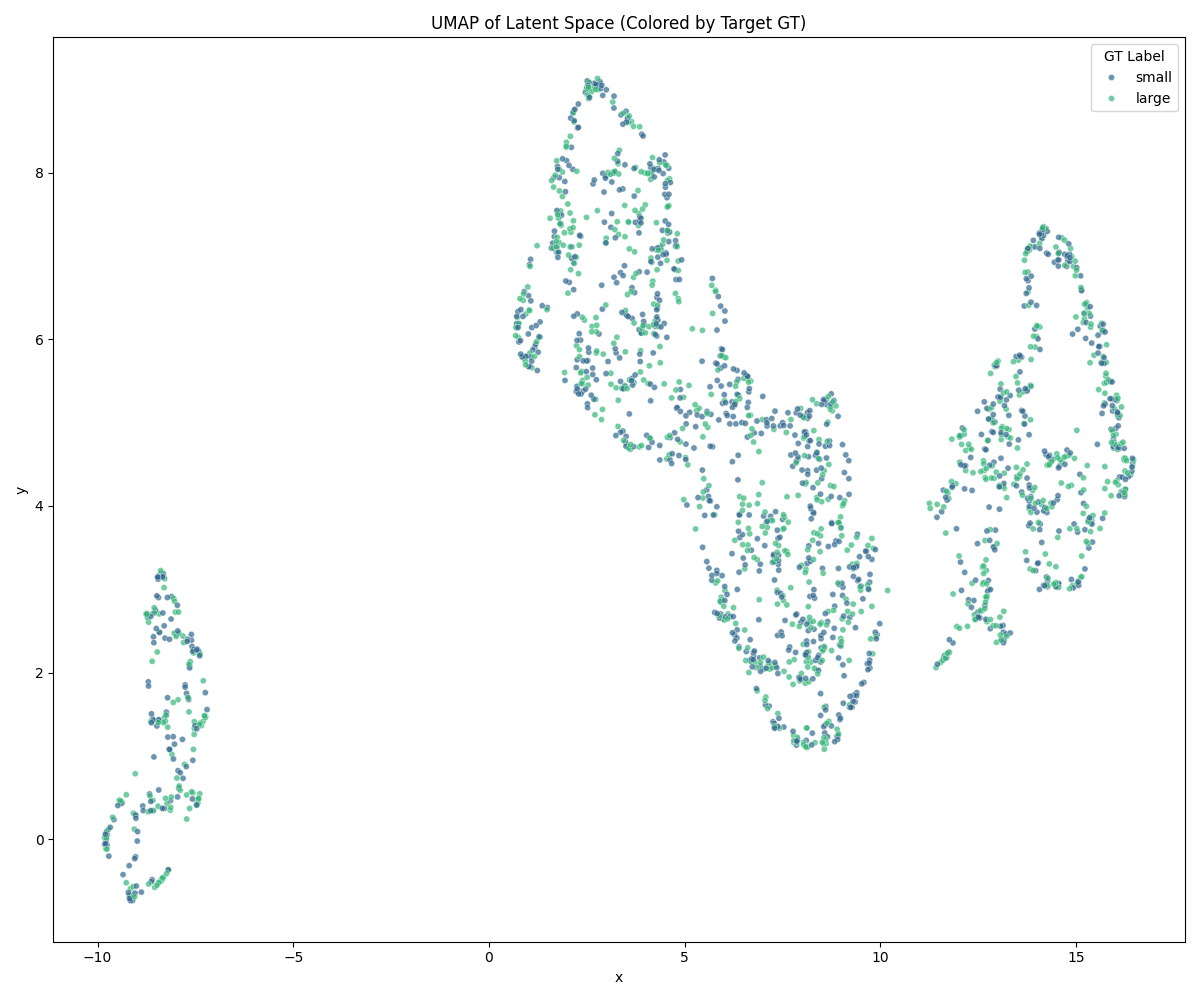}
\caption{Zero-Shot Latent Space for $\text{CLEVR-Tex} \longrightarrow \text{CLEVR}$ (Material, Shape, Size). The model, trained on diverse textures, retains robust features that are highly transferable to the simpler target domain.}
\label{fig:tex_to_clevr_zeroshot_latent}
\end{figure}

The reverse transfer from complex, textured data to simple, uniform data was highly effective, demonstrating robust generalization across all concepts, including color (accuracy $\approx 90\%$) as seen in Figure \ref{fig:tex_to_clevr_zeroshot_latent} . 

\begin{figure}
    \centering
    \includegraphics[width=0.35\linewidth]{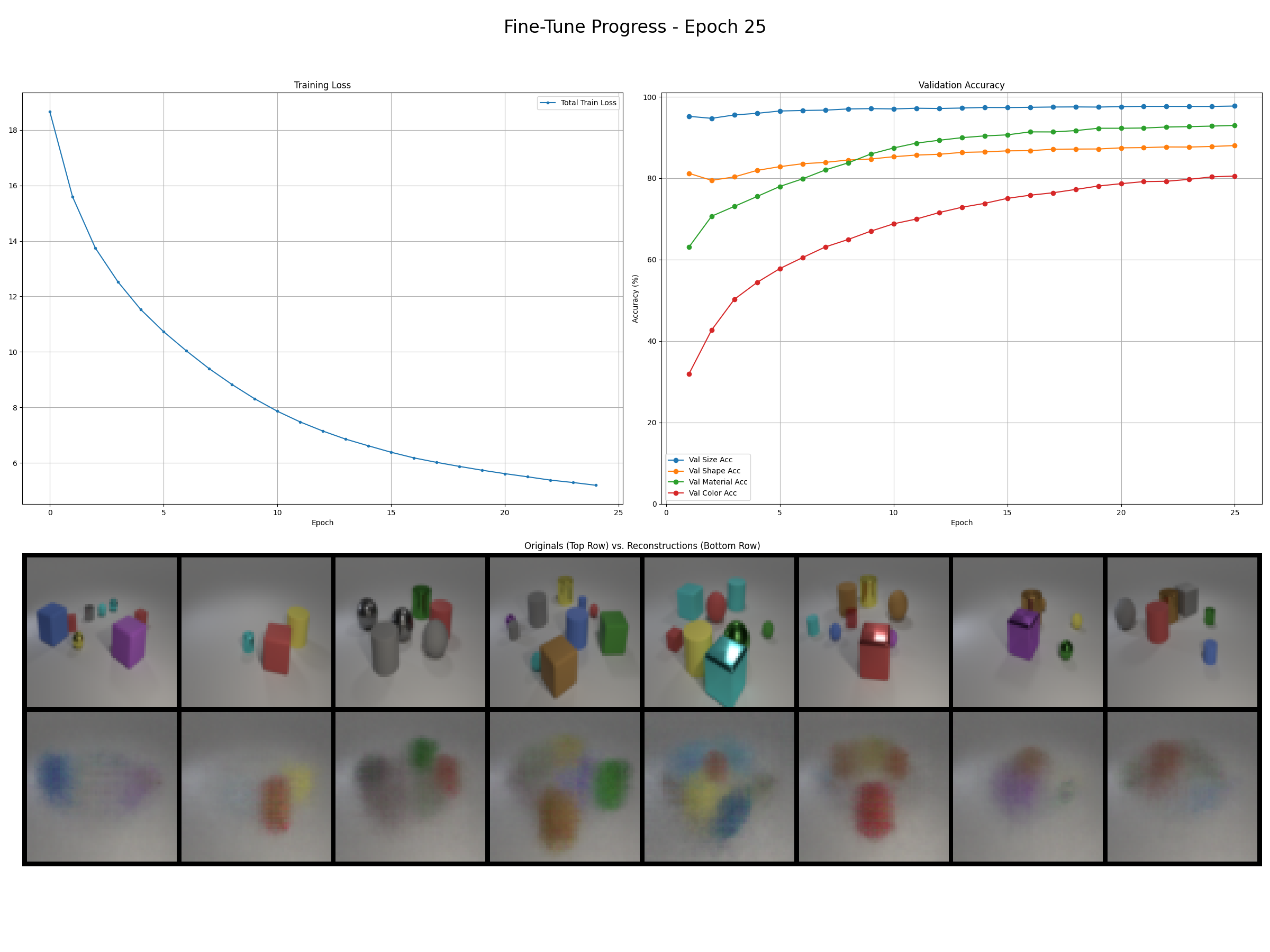}
    \includegraphics[width=0.5\linewidth]{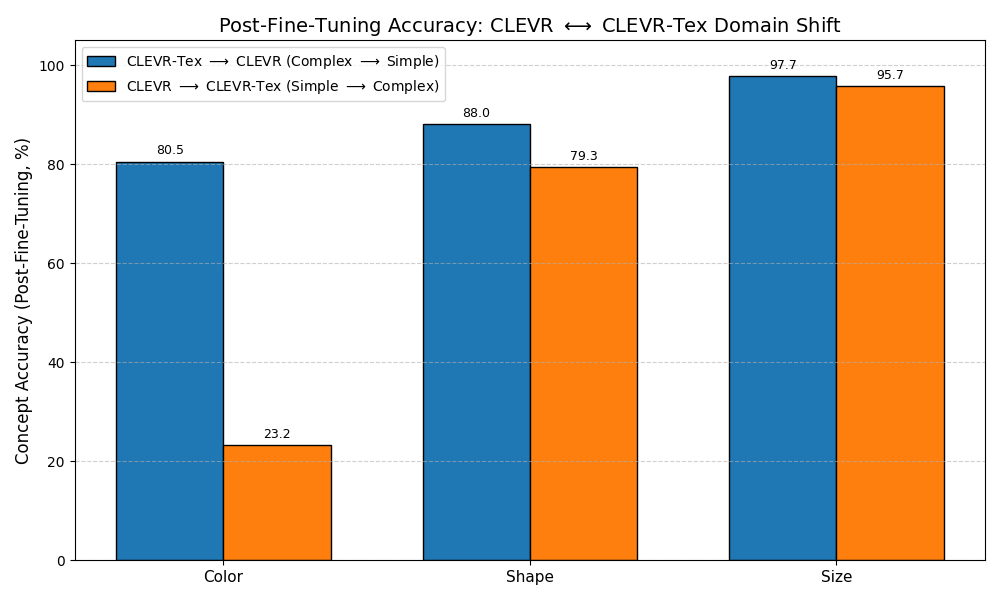}
    \caption{FINE TUNE RESULTS and RECONSTRUCTION(left), concept accuracy (right)}
    \label{fig:tex_to_clevr_finetune_latent}
\end{figure}

This finding implies as based on (Figure \ref{fig:tex_to_clevr_finetune_latent}) that exposure to visually diverse, textured data is a necessary inductive bias for learning fundamental and robust representations of surface-level properties.

% \begin{figure}[h!]
%     \centering
%     \includegraphics[width=0.24\linewidth]{RQ2/finetune_report_epoch_20 (7).png}
%     \includegraphics[width=0.24\linewidth]{RQ2/finetune_report_epoch_20 (8).png}
%     \includegraphics[width=0.24\linewidth]{RQ2/finetune_report_epoch_20 (9).png}
    
%     \caption{UMAP visualizations of the latent space after fine-tuning on different domain shifts. From left to right: Shape clusters after Dsprites $\rightarrow$ 3DShapes transfer; Shape clusters after CLEVR $\rightarrow$ CLEVR-TEX transfer; entangled Shape and Color clusters after CLEVR $\rightarrow$ CLEVR-2D transfer.}
%     \label{fig:domain_shift_umaps}
% \end{figure}

% \begin{figure}[h!]
%     \centering
%     \includegraphics[width=0.24\linewidth]{RQ2/finetune_report_epoch_30 (4).png}
%     \includegraphics[width=0.24\linewidth]{RQ2/finetune_report_epoch_30 (5).png}
%     \includegraphics[width=0.24\linewidth]{RQ2/monitoring_report_epoch_20 (4).png}
    
%     \caption{UMAP visualizations of the latent space after fine-tuning on different domain shifts. From left to right: Shape clusters after 3DShapes $\rightarrow$ dSprites transfer; Shape clusters after CLEVR-Tex $\rightarrow$ CLEVR transfer; entangled Shape and Color clusters after CLEVR-2D $\rightarrow$ CLEVR transfer.}
%     \label{fig:domain_shift_umaps_2}
% \end{figure}

\section{RQ3}

\begin{figure}
    \centering
    \includegraphics[width=0.95\linewidth]{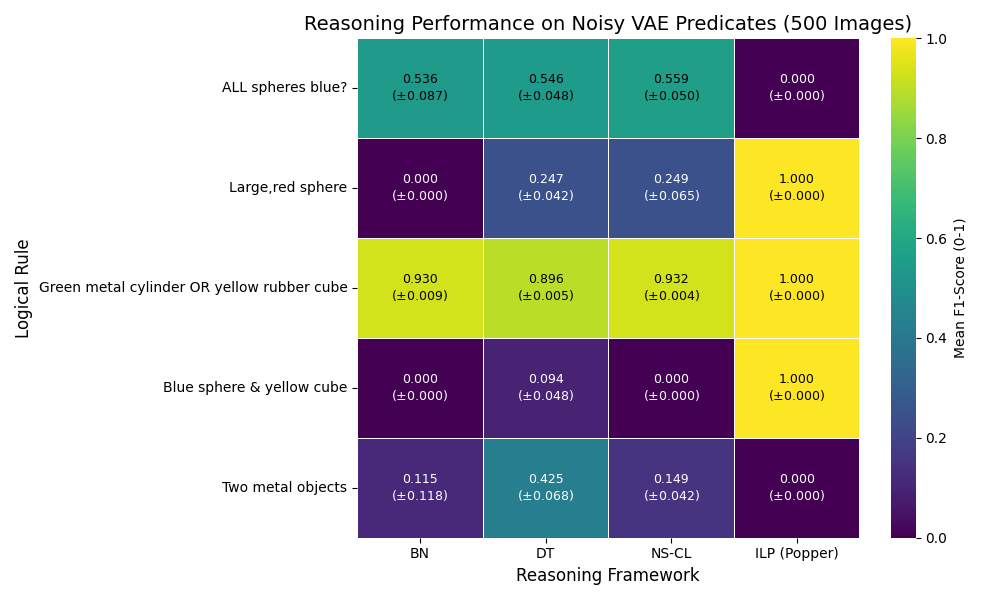}
    \caption{Performance of varioius Reasoniong framworks on 500 image set}
    \label{fig:500_reasoning}
\end{figure}

This section details the performance of the downstream reasoning frameworks when supplied with conceptual predicates generated by the Slot-VAE, trained using $15\%$ label supervision on a set of 500 CLEVR images. The quantitative F1-scores for the synthetic rules are presented in Table \ref{tab:gap_500}, with visual comparison available in Figure \ref{fig:500_reasoning}.

\begin{table}[tb]
\centering
\caption{Performance on \textbf{500 Images} (15\% VAE Supervision). Values are F1-Score (Mean $\pm$ Std. Dev.) over 5 runs.}
\label{tab:gap_500}
\tiny % Keeping the smallest standard font size
\begin{tabular}{@{}llcc@{}}
\toprule
\textbf{Rule} & \textbf{Framework} & \textbf{GT F1 (Mean)} & \textbf{VAE F1 (Mean $\pm$ Std. Dev.)} \\
\midrule
\multirow{4}{*}{\begin{tabular}[c]{@{}l@{}}"Exactly two \\ metal objects?"\end{tabular}} & Bayesian Network (BN) & 0.400 & 0.115 $\pm$ 0.118 \\
& \textbf{Decision Tree (DT)} & \textbf{1.000} & \textbf{0.425 $\pm$ 0.068} \\
& NS-CL & 0.161 & 0.149 $\pm$ 0.042 \\
& ILP (Popper) & 0.000 & 0.000 $\pm$ 0.000 \\
\midrule
\multirow{4}{*}{\begin{tabular}[c]{@{}l@{}}"Blue sphere AND \\ yellow cube?"\end{tabular}} & Bayesian Network (BN) & 0.000 & 0.000 $\pm$ 0.000 \\
& Decision Tree (DT) & \textbf{0.106} & \textbf{0.094 $\pm$ 0.048} \\
& NS-CL & 0.000 & 0.000 $\pm$ 0.000 \\
& ILP (Popper) & 1.000 & 1.000 $\pm$ 0.000 \\
\midrule
\multirow{4}{*}{\begin{tabular}[c]{@{}l@{}}"Green metal cylinder \\ OR yellow rubber cube?"\end{tabular}} & Bayesian Network (BN) & 0.944 & 0.930 $\pm$ 0.009 \\
& Decision Tree (DT) & 0.901 & 0.896 $\pm$ 0.005 \\
& \textbf{NS-CL} & \textbf{0.935} & \textbf{0.932 $\pm$ 0.004} \\
& ILP (Popper) & 1.000 & 1.000 $\pm$ 0.000 \\
\midrule
\multirow{4}{*}{\begin{tabular}[c]{@{}l@{}}"Large, red \\ sphere?"\end{tabular}} & Bayesian Network (BN) & 0.000 & 0.000 $\pm$ 0.000 \\
& \textbf{Decision Tree (DT)} & 0.259 & 0.247 $\pm$ 0.042 \\
& \textbf{NS-CL} & \textbf{0.313} & \textbf{0.249 $\pm$ 0.065} \\
& ILP (Popper) & 1.000 & 1.000 $\pm$ 0.000 \\
\midrule
\multirow{4}{*}{\begin{tabular}[c]{@{}l@{}}"Are ALL \\ spheres blue?"\end{tabular}} & Bayesian Network (BN) & 0.493 & 0.536 $\pm$ 0.087 \\
& Decision Tree (DT) & 0.779 & 0.546 $\pm$ 0.048 \\
& \textbf{NS-CL} & \textbf{0.805} & \textbf{0.559 $\pm$ 0.050} \\
& ILP (Popper) & 0.148 & 0.045 $\pm$ 0.036 \\
\bottomrule
\end{tabular}
\end{table}

\section{RQ4}
This section provides a detailed breakdown of the visualizations and quantitative results supporting the findings of Research Question 4 (RQ4) in the main paper, focusing on the trade-off between in-domain specialization and zero-shot generalization under sparse supervision.

\subsection{Latent Space Analysis of Concept Grounding}

To better understand how different architectures and supervision levels affect the disentanglement of high-level diagnostic features, we visualize the latent spaces (using UMAP or t-SNE) under various conditions.

\begin{figure}[htbp]
    \centering
    \includegraphics[width=0.3\linewidth]{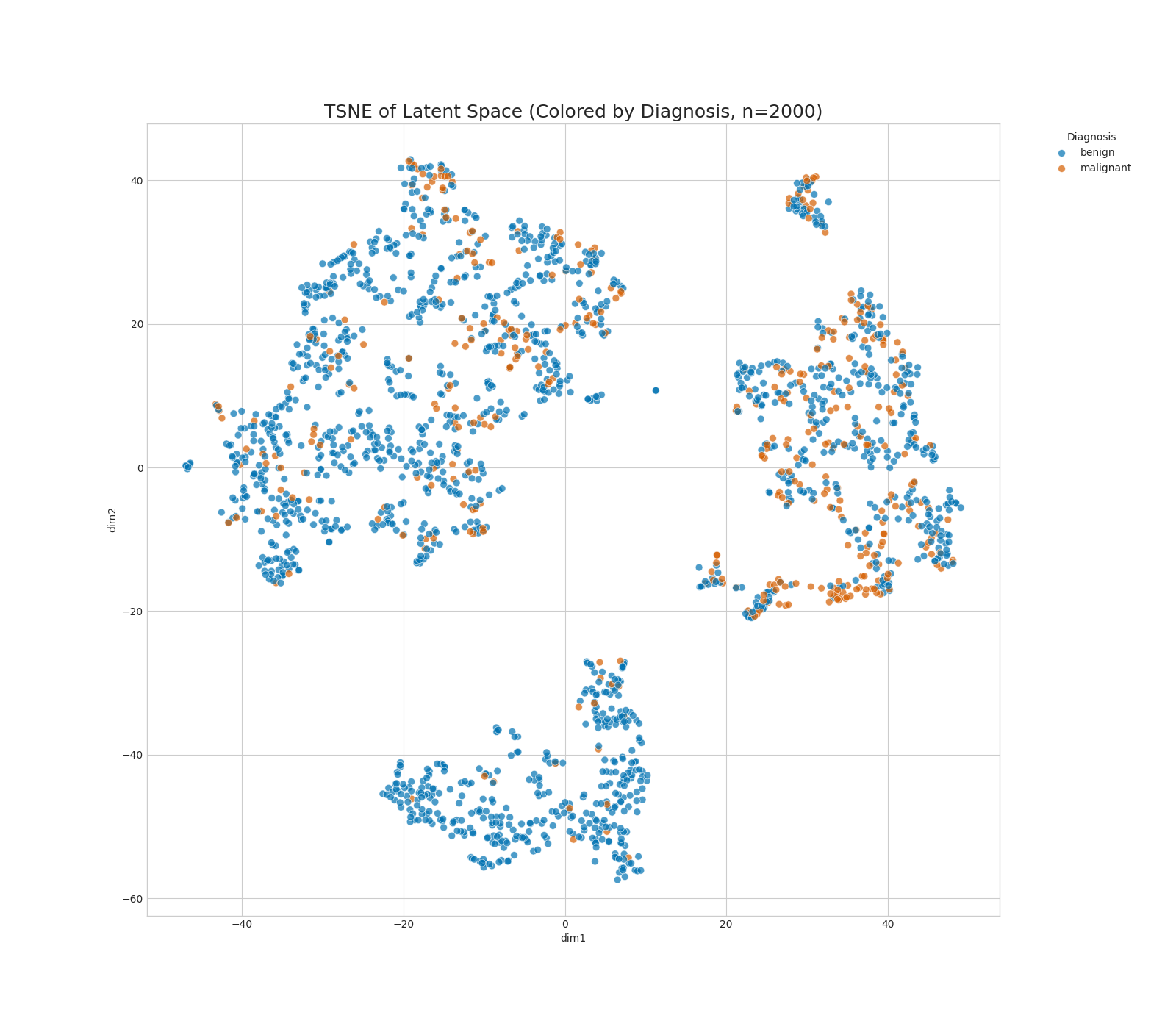}
    \includegraphics[width=0.3\linewidth]{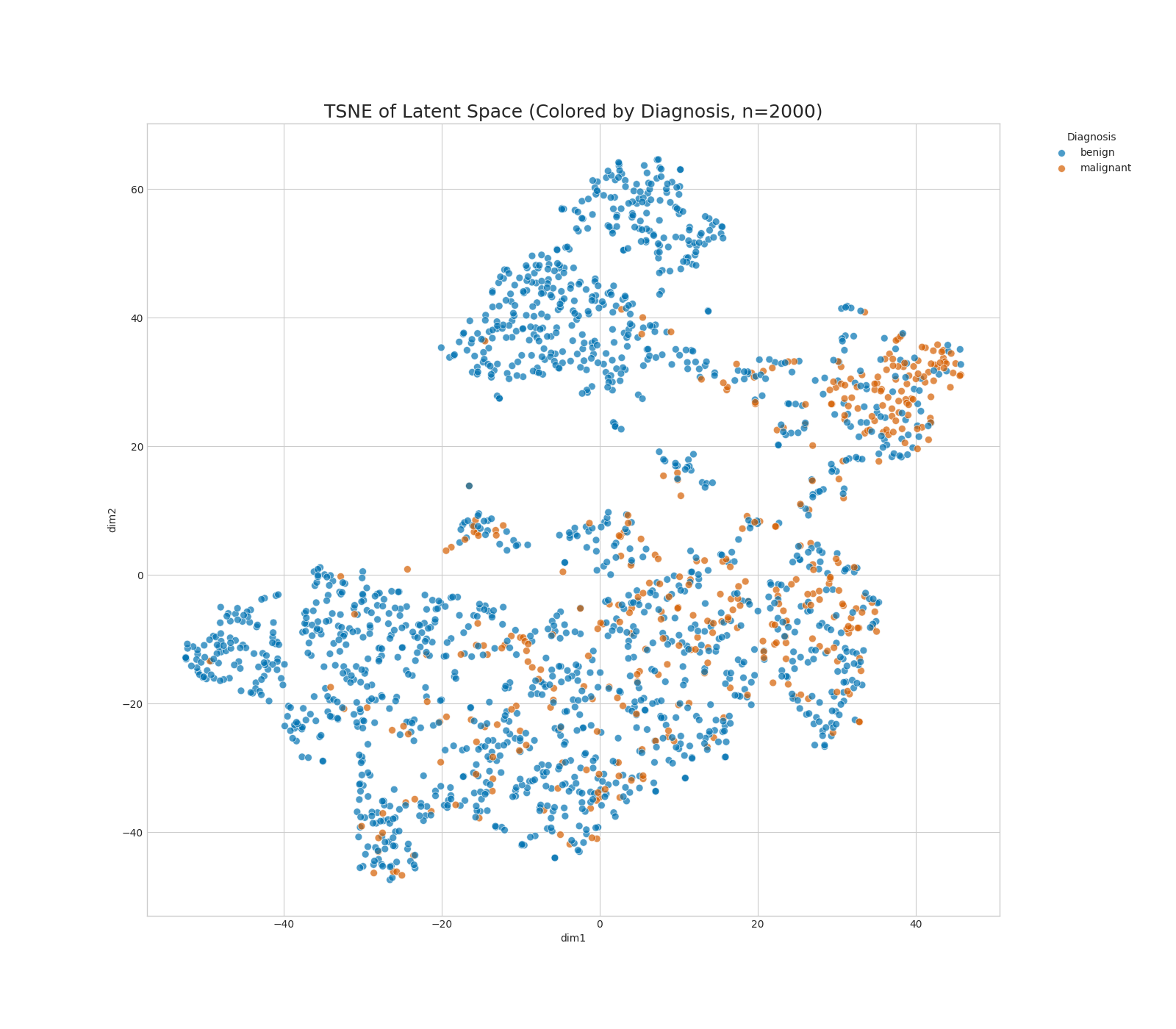}
    \includegraphics[width=0.3\linewidth]{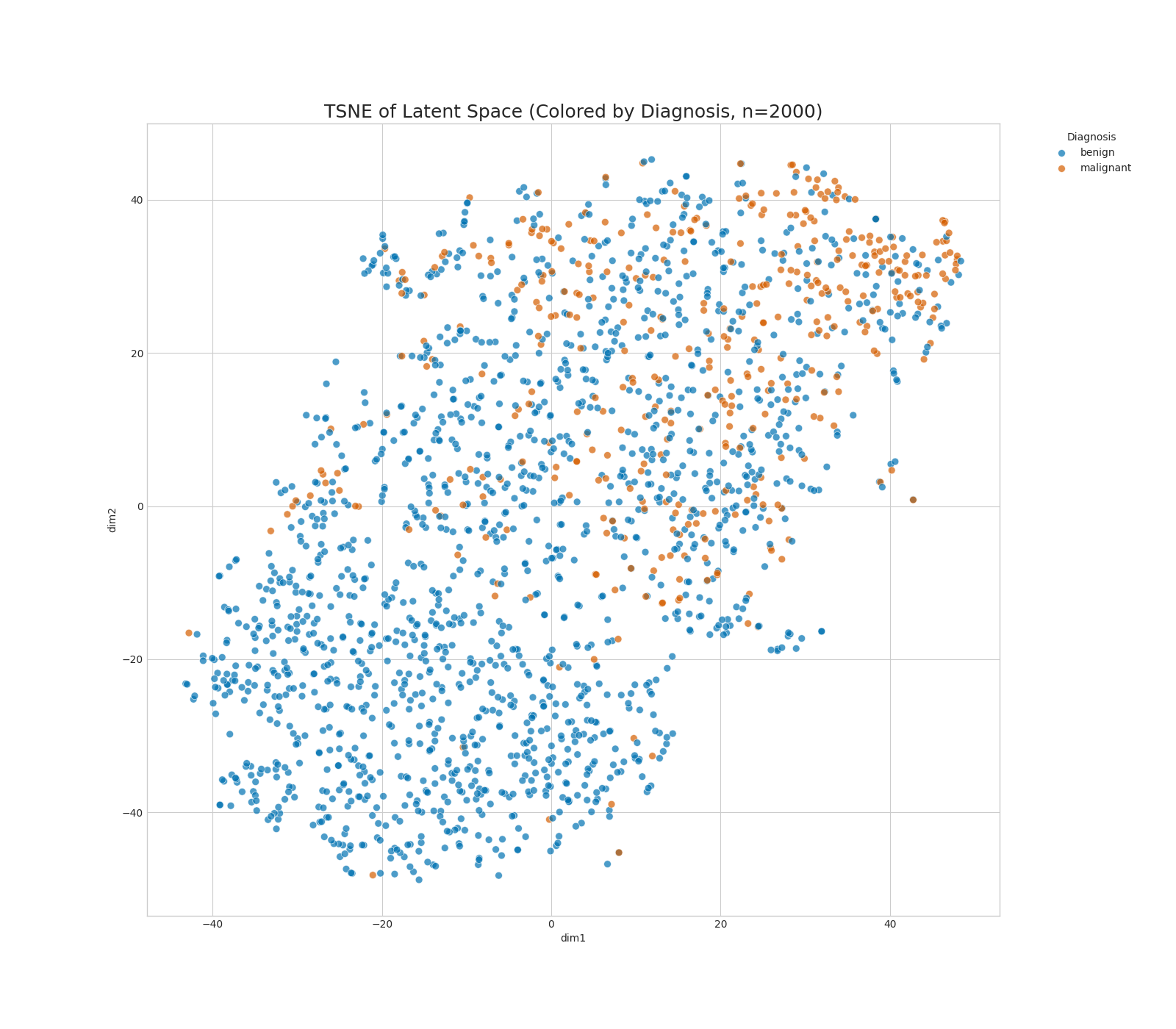}
    \includegraphics[width=0.3\linewidth]{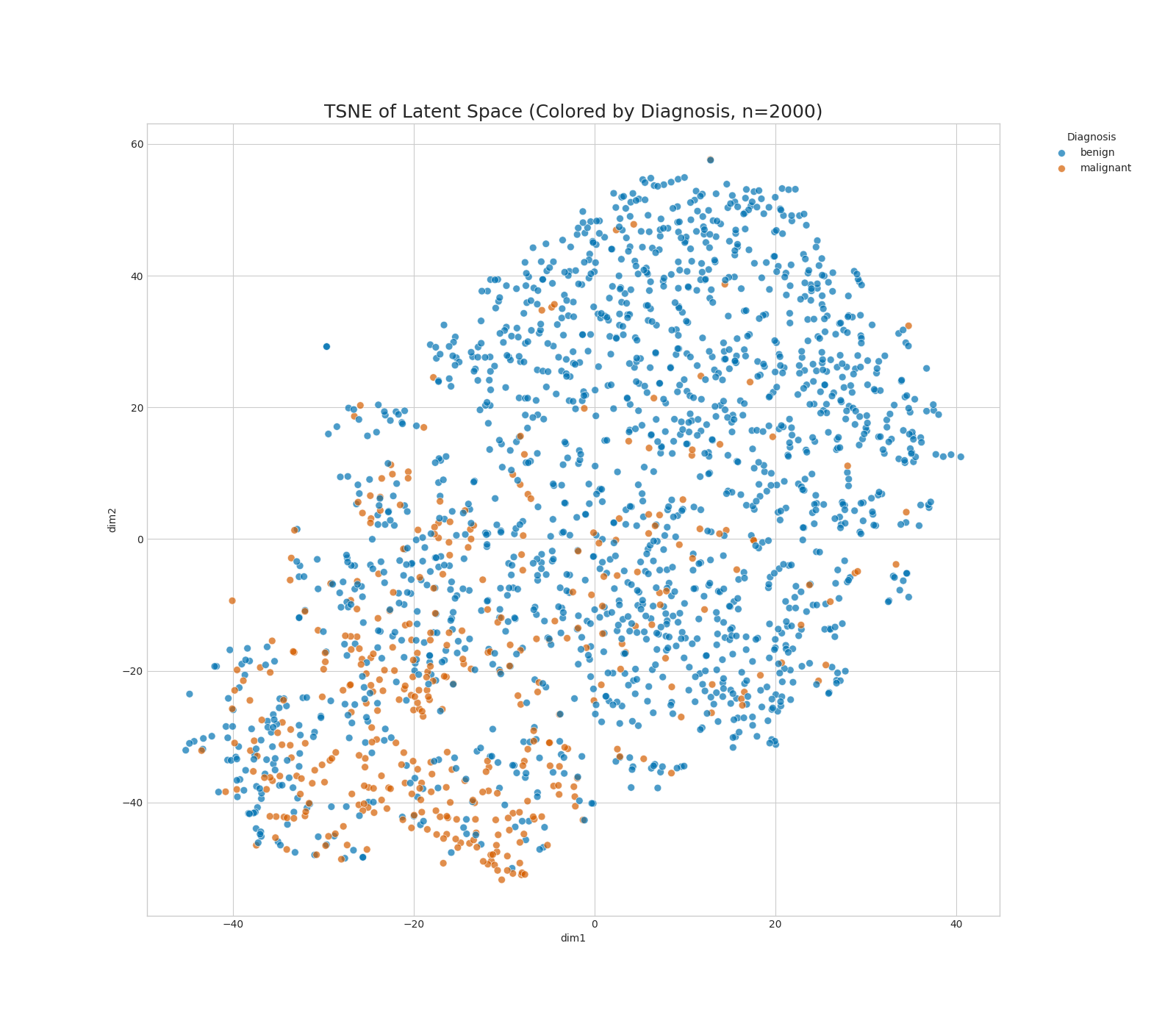}
    \includegraphics[width=0.3\linewidth]{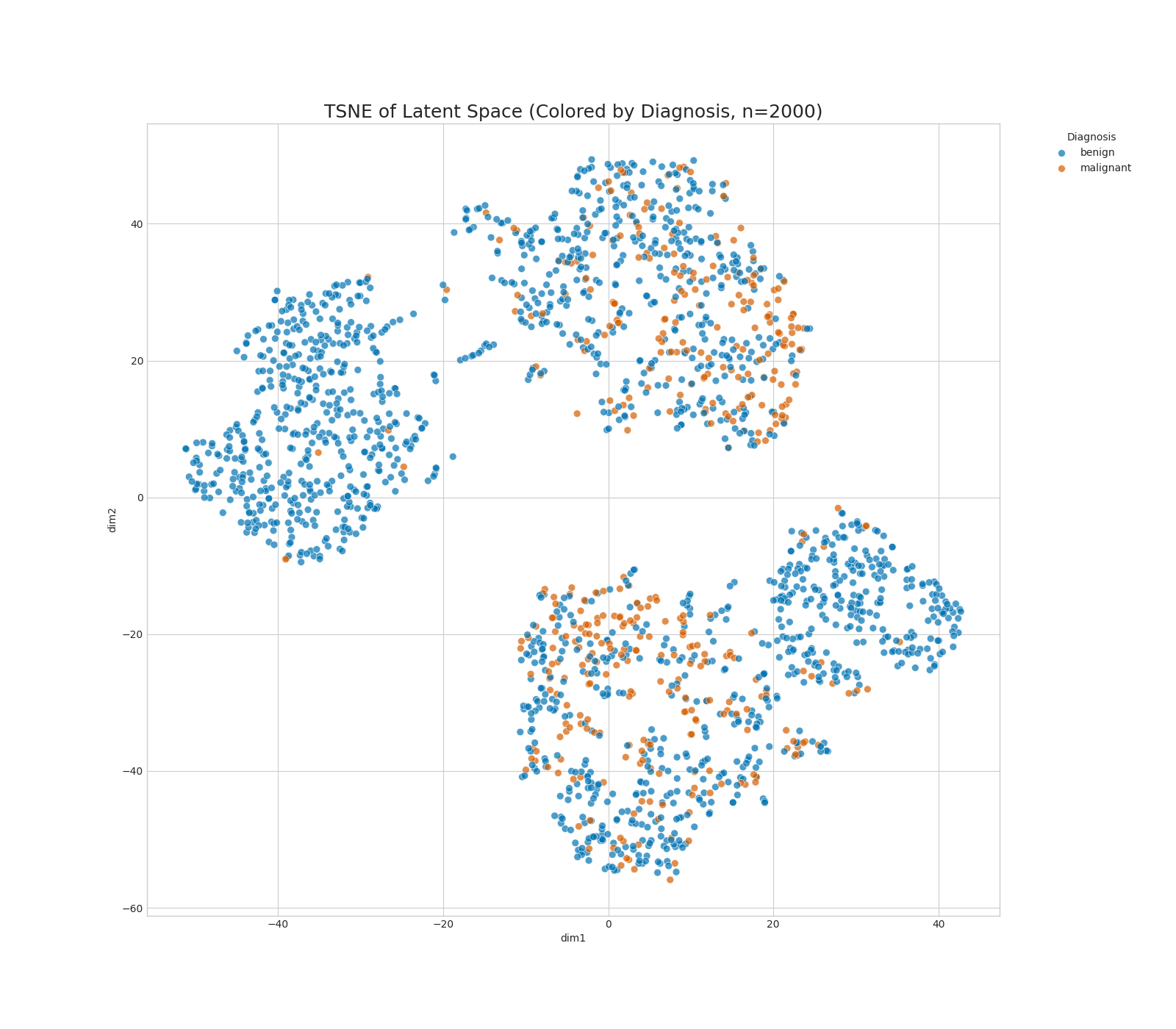}
    \includegraphics[width=0.3\linewidth]{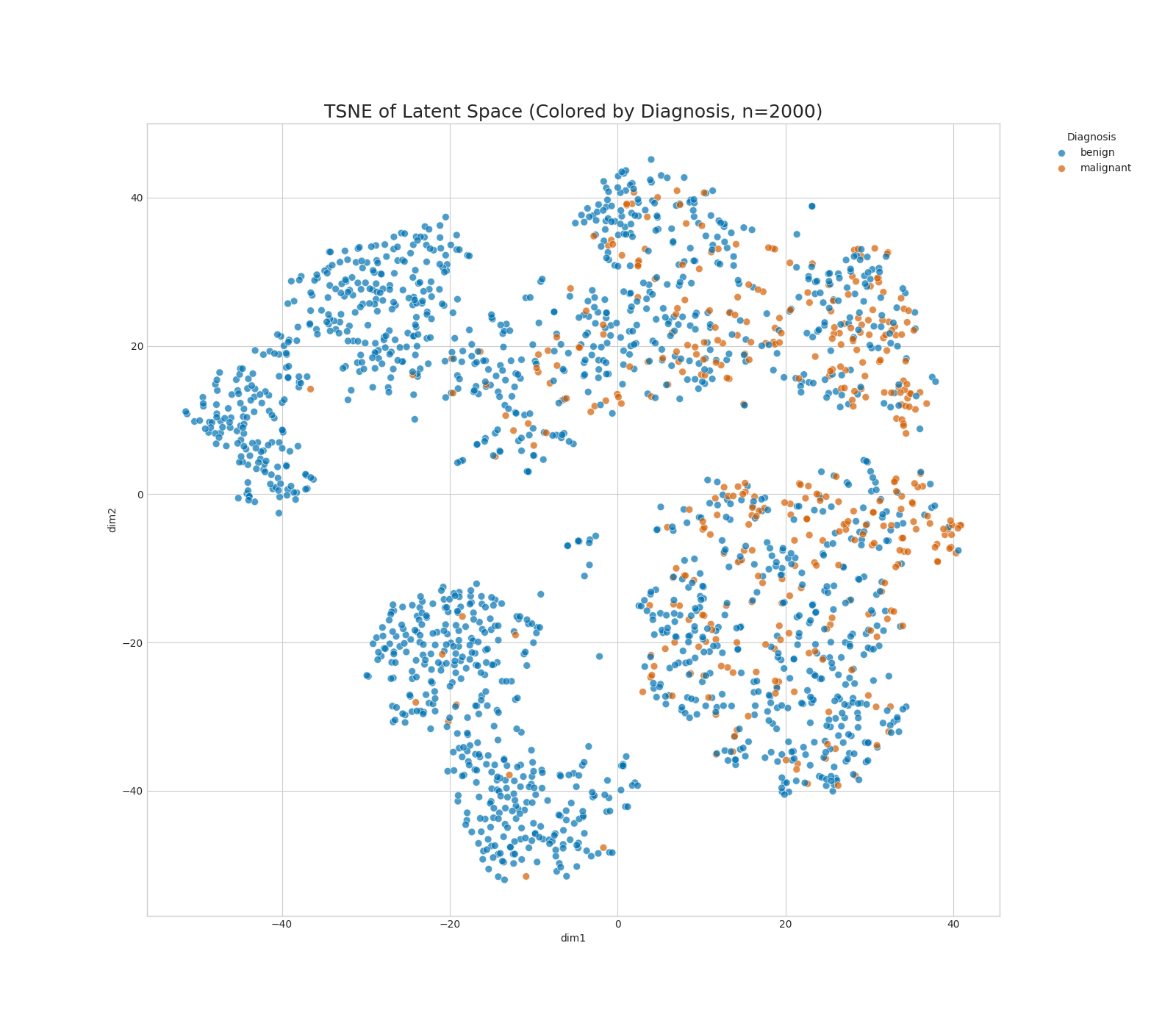}
    \caption{Malignant vs Benign Classification in latent space vs different label supervision (1\% to 100\%, left to right) on SLOTVAE ($k=2$).}
    \label{fig:slotvae_latent_separation}
\end{figure}

\begin{figure}[htbp]
    \centering
     \includegraphics[width=0.33\linewidth]{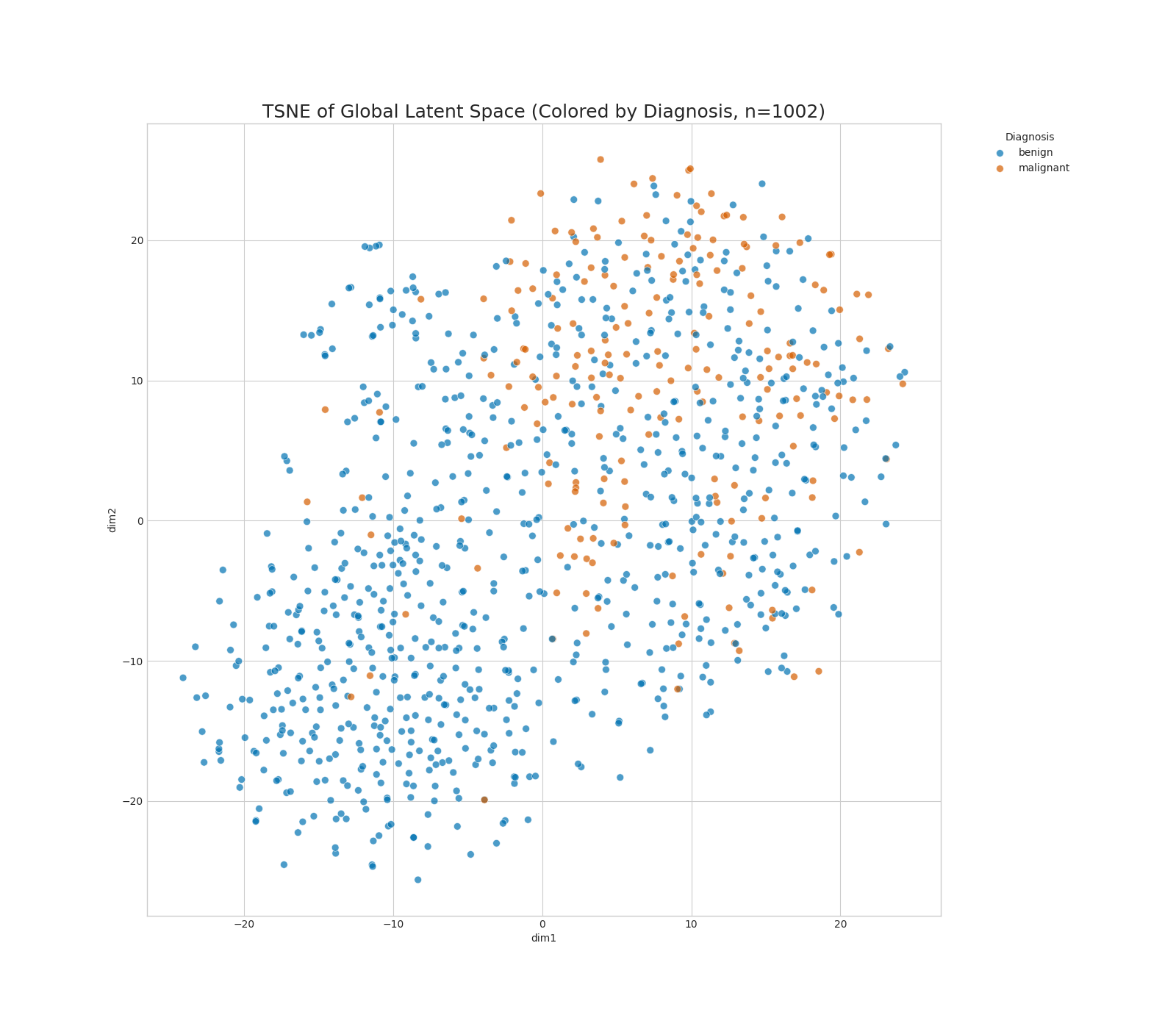}
    \includegraphics[width=0.33\linewidth]{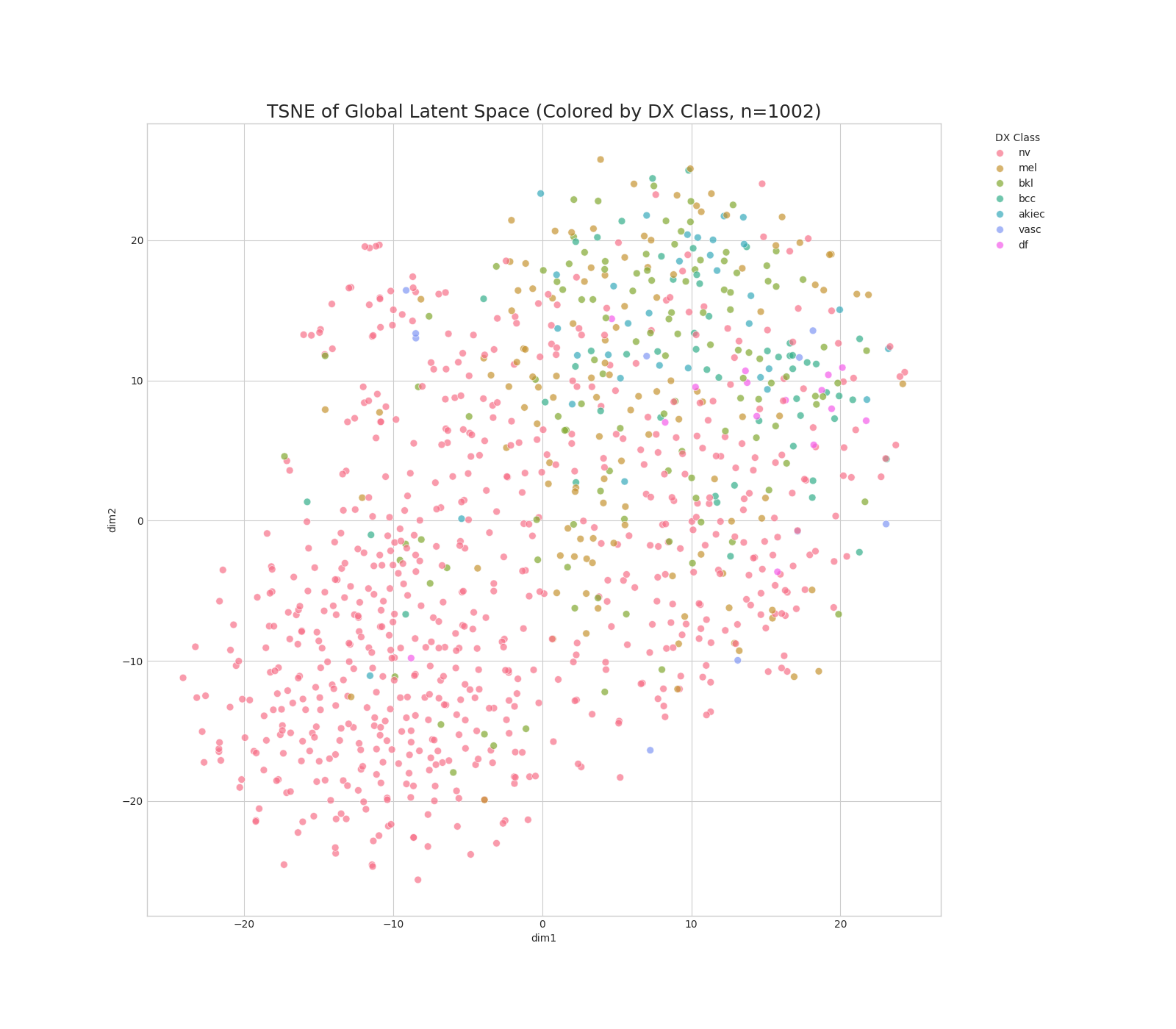}
    \includegraphics[width=0.33\linewidth]{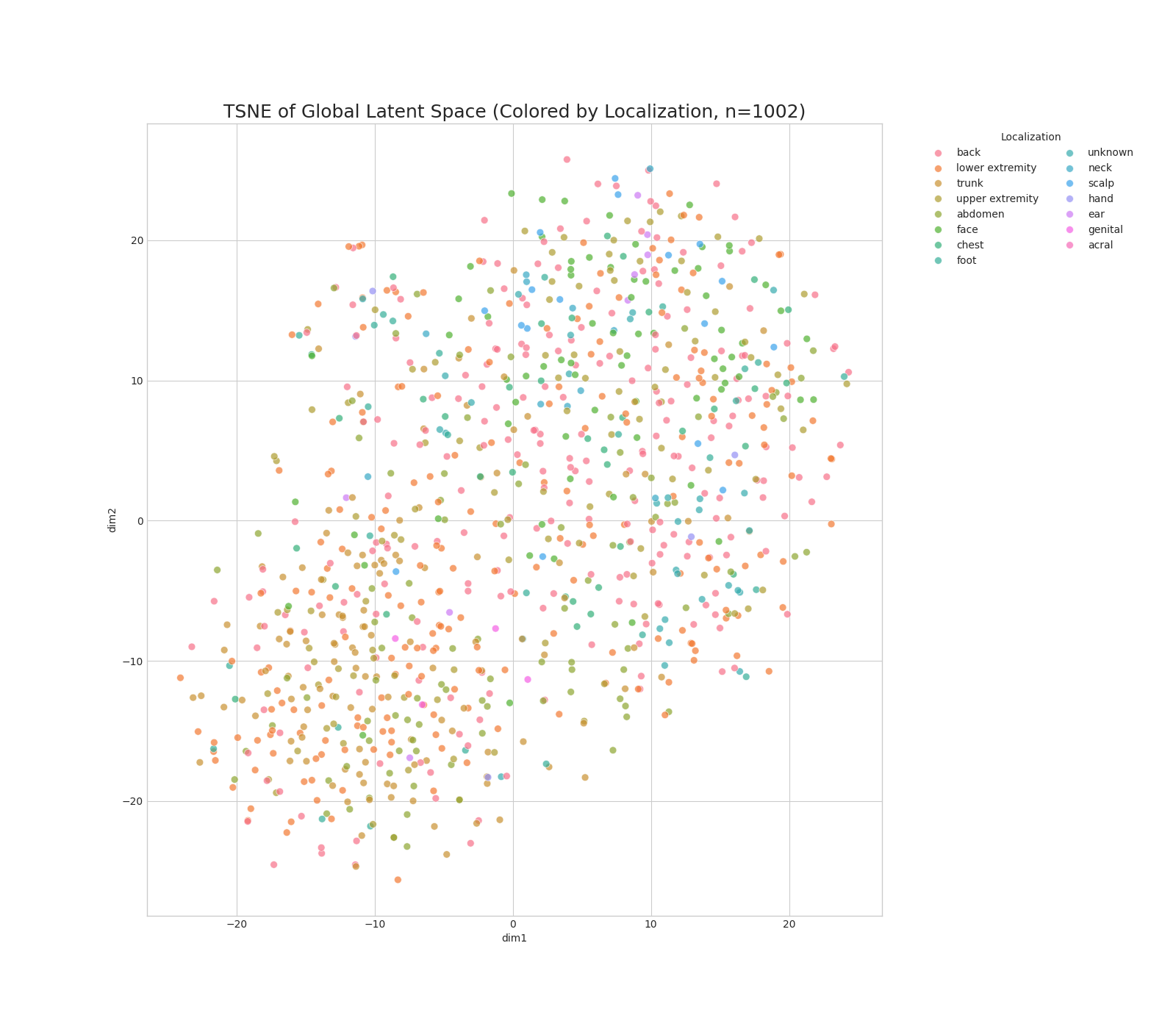}
    
    \caption{GLOBAL VAE latent space visualizations for the HAM dataset at 75\% Label supervision, showing feature separation by 7-class diagnosis, binary malignancy status (\textit{dx}), and localization.}
    \label{fig:globalvae_75_latent}
\end{figure}

\begin{figure}[htbp]
    \centering
    \includegraphics[width=0.30\linewidth]{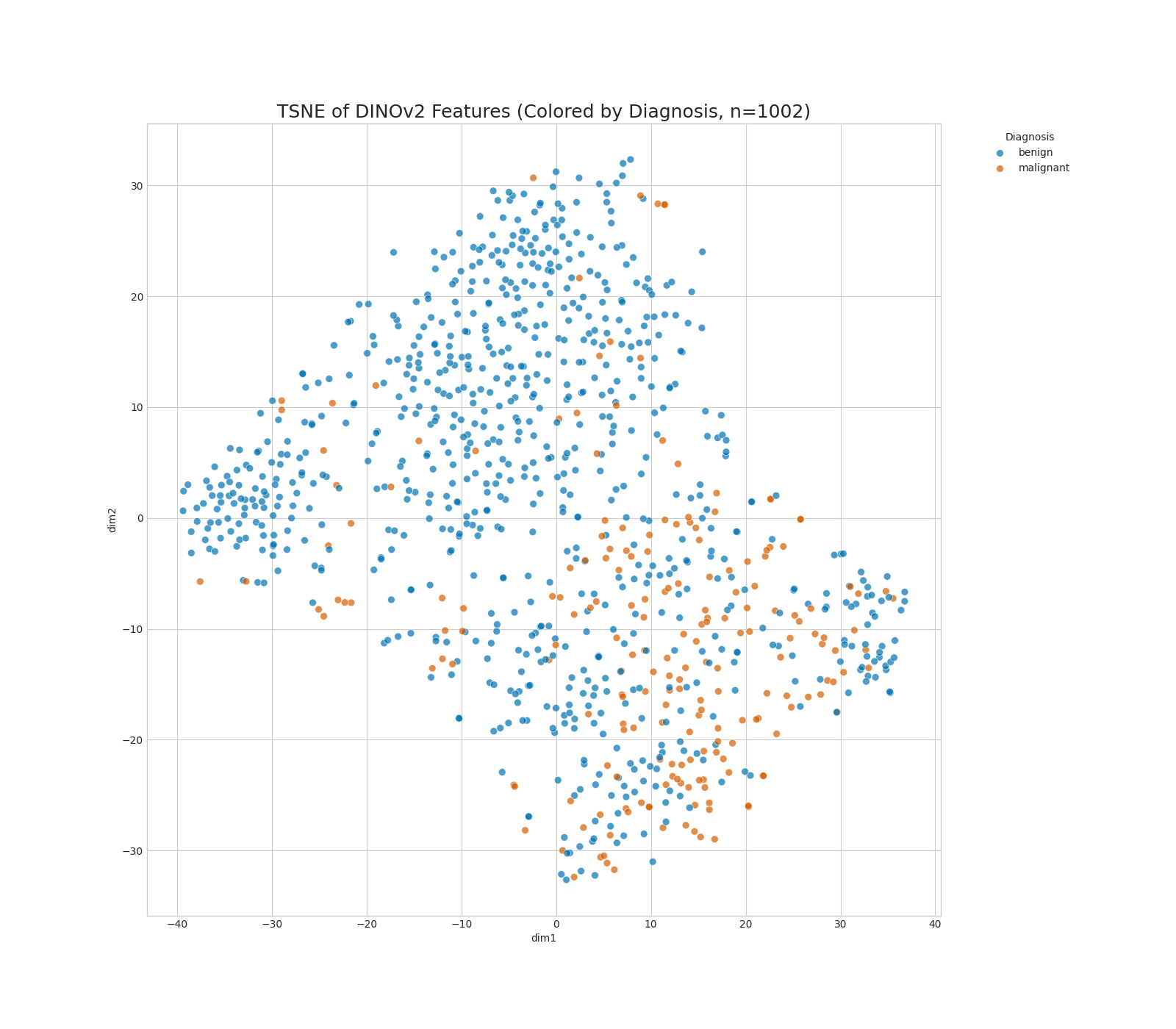}
    \includegraphics[width=0.30\linewidth]{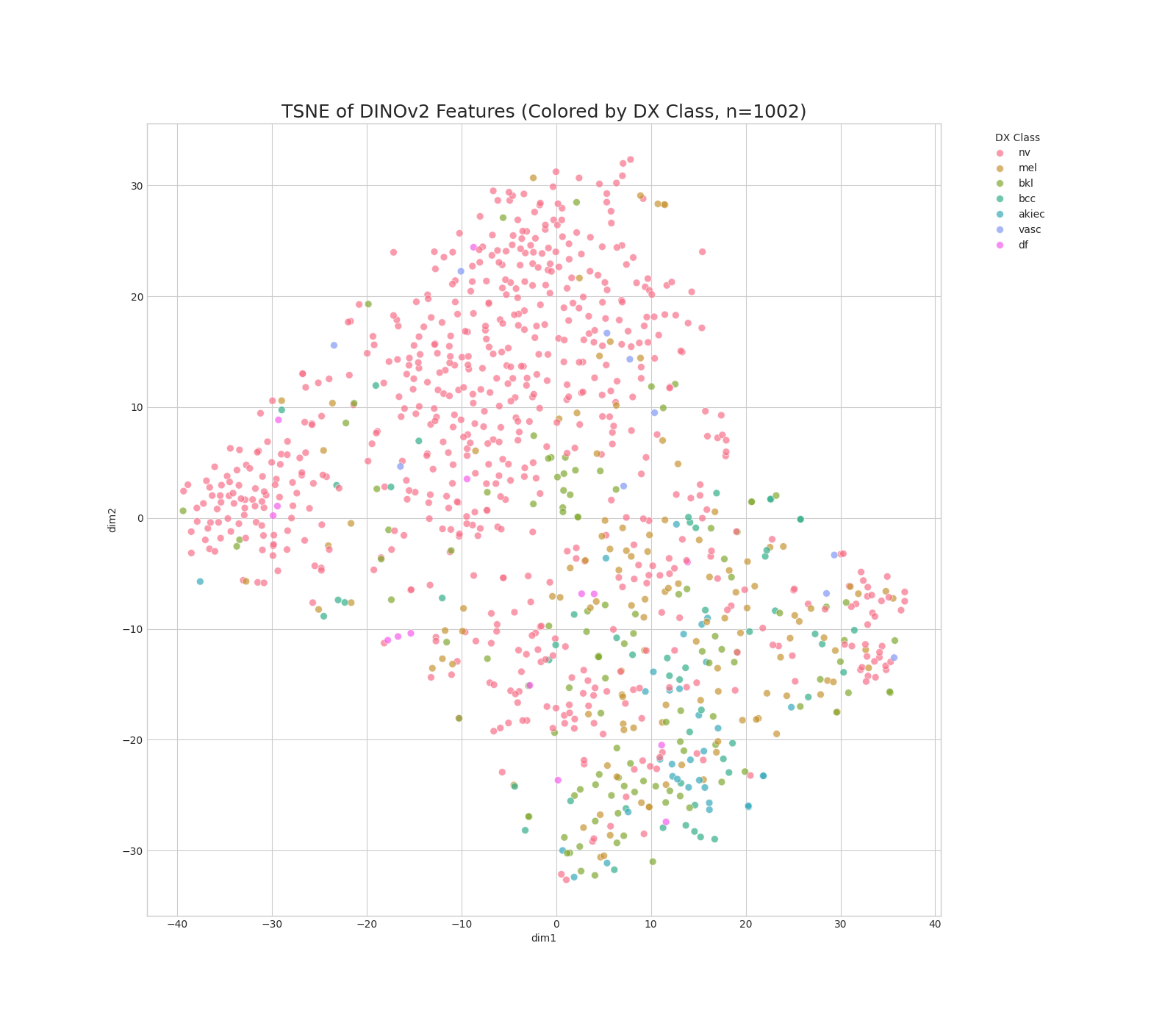}
    \includegraphics[width=0.30\linewidth]{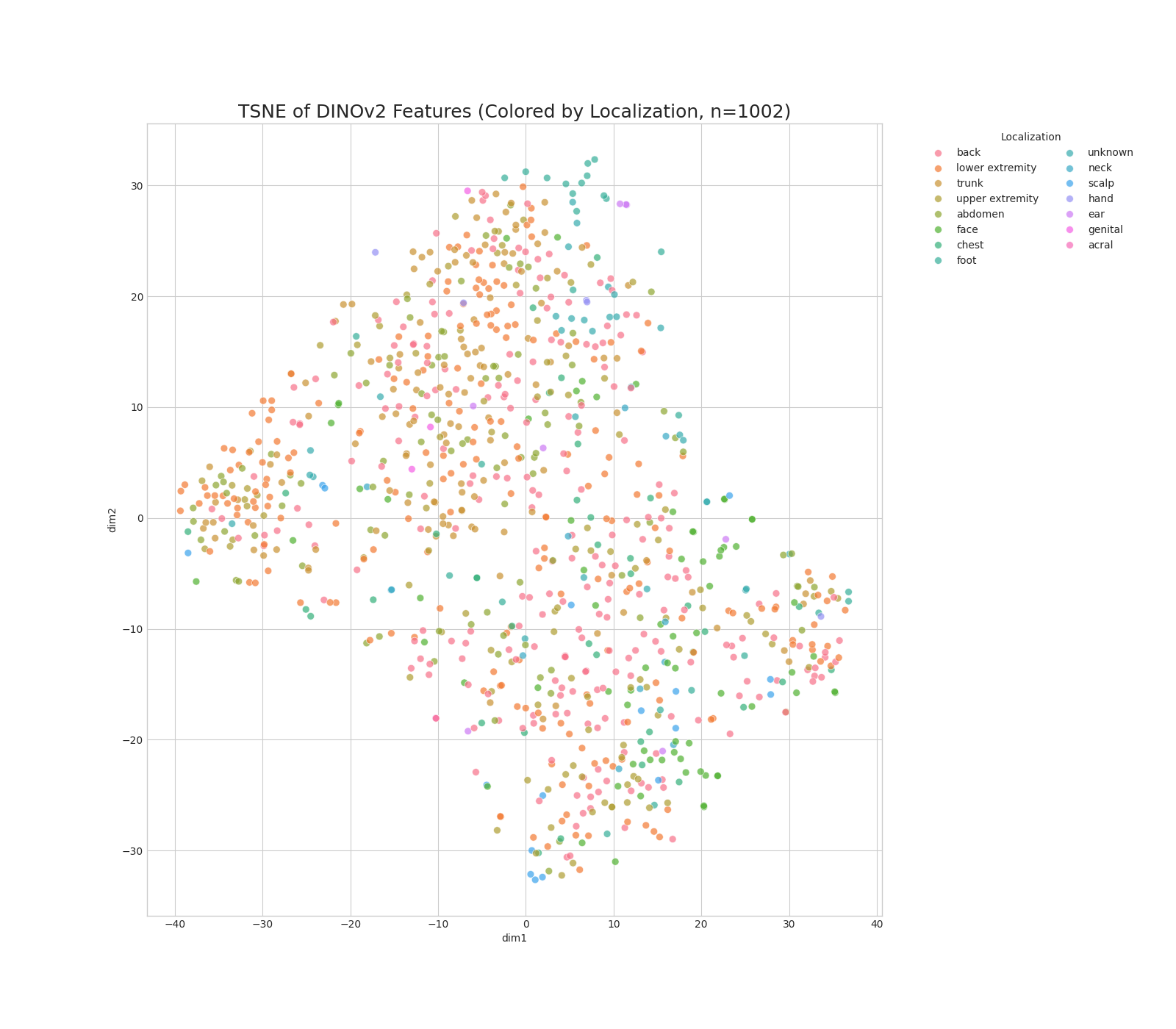}

    \caption{DINOV2 latent space visualizations  for different concepts (diagnosis/\textit{dx}, and localization) for 75 \% label supervision levels.}
    \label{fig:dinov2_latent_concepts}
\end{figure}

\textbf{Analysis of \autoref{fig:slotvae_latent_separation}:}
illustrates the malignant vs. benign classification performance of our SlotVAE ($k=2$) across the latent space for HAM. We observe a clear progression: at 1\% supervision, the classes are highly entangled, corresponding to the model's low initial zero-shot performance. As supervision increases, the separation becomes progressively cleaner, demonstrating that the weak supervision effectively guides the object-centric latent space to disentangle the critical diagnostic concept. \autoref{fig:dinov2_latent_concepts} presents a similar view for the DINOv2\cite{oquab2024dinov2learningrobustvisual} baseline, showing the feature representation separated by three concepts: the 7-class diagnosis ($\textit{diagnosis}$), the binary malignancy status ($\textit{dx}$), and the $\textit{localization}$ of the lesion, for 75\% supervision. While DINOv2 shows robust separation for $\textit{diagnosis}$ and $\textit{dx}$ even at lower supervision levels, the $\textit{localization}$ concept remains poorly clustered, reflecting the main paper's finding that models with object-centric or fixed-region inductive biases struggle to capture holistic, scene-centric properties.

\textbf{Analysis of \autoref{fig:globalvae_75_latent}:}
As noted in the main paper, the GlobalVAE\cite{watters2019spatialbroadcastdecodersimple}, which uses a single, holistic latent vector, was the only model to successfully learn the complex conjunctive Rule 3 (requiring $\textit{dx}$ AND $\textit{localization}$) at 75\% supervision, aligning with the paper's quantitative results in Table 5. This finding is visually reinforced by \autoref{fig:globalvae_75_latent}. Specifically, the  GlobalVAE at 75\% supervision  shows notably improved clustering and separation for the $\textit{localization}$ feature (right plot) compared to the SlotVAE and DINOv2 baselines. This visualization supports the conclusion that the inductive bias of a global, holistic encoder is more suitable for scene-centric concepts (like localization on the body) than spatially partitioned or slot-based architectures.

\subsection{Reasoning Frameworks Performance Comparison}

\begin{figure}[htbp]
    \centering
     \includegraphics[width=1.00\linewidth]{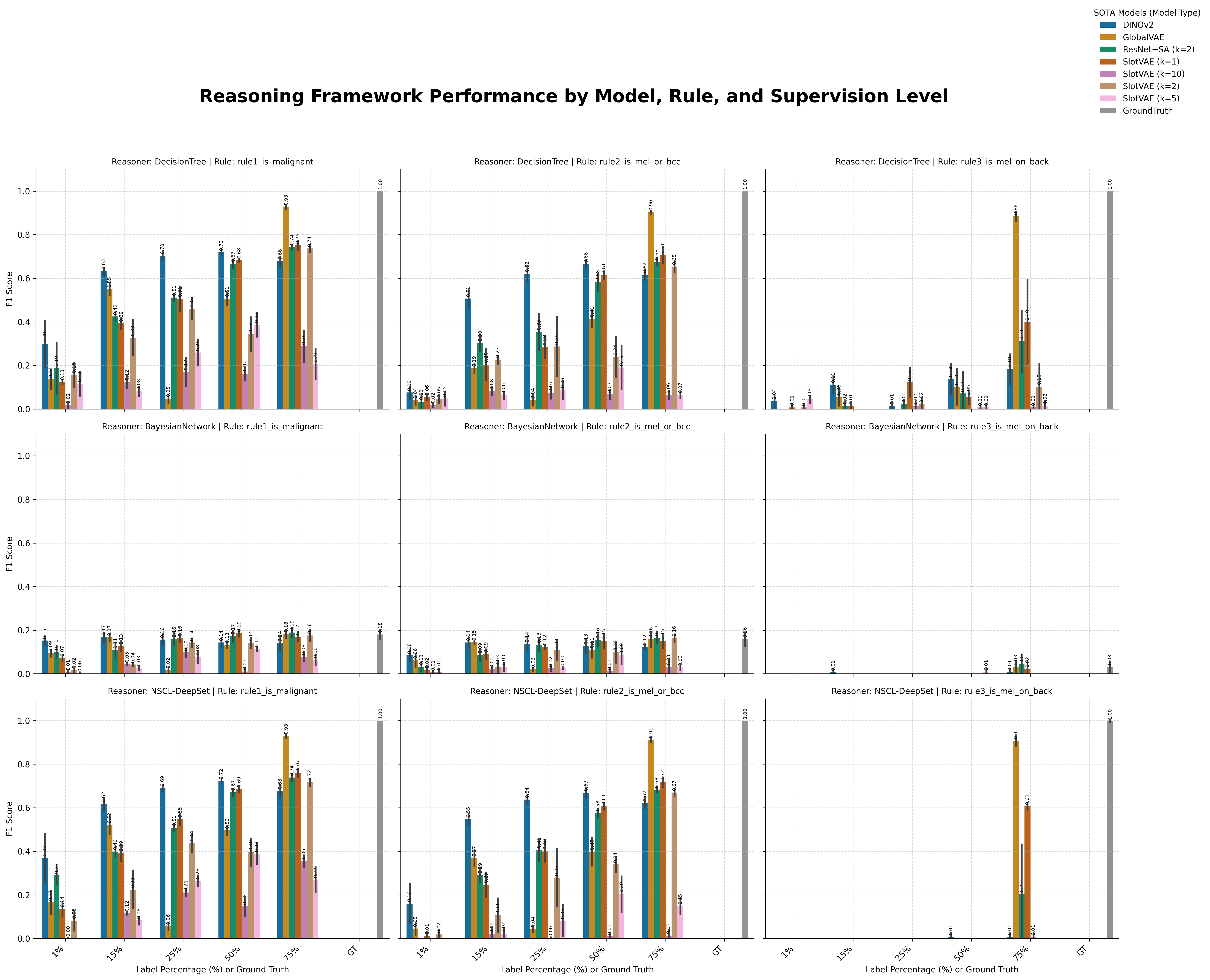}
    \caption{Comparative F1-Scores of downstream reasoning frameworks (Decision Tree/NS-CL) across three rules (Simple, Disjunctive, Conjunctive), supervision levels, and SOTA models on the HAM dataset.}
    \label{fig:reasoning_comparison_rules}
\end{figure}

\textbf{Analysis of \autoref{fig:reasoning_comparison_rules}:}
 provides a comprehensive quantitative summary of the neuro-symbolic pipeline performance on the HAM dataset, showing the F1-scores achieved by the downstream reasoners (NS-CL and  Decision Tree ) when given noisy predicates generated by the perception models (SlotVAE variations, DINOV2, GlobalVAE).

\begin{itemize}
    \item \textbf{Simple and Disjunctive Rules (Rules 1 and 2):} Performance on these rules, which primarily rely on the high-fidelity $\textit{dx}$ predicate (binary malignancy status), scales directly with the predicate quality of the generator model. Models that achieved high in-domain $\textit{dx}$ accuracy (e.g., GlobalVAE and DINOv2 at higher supervision levels) show competitive F1-scores for these simpler rules. This confirms that these rules, requiring only single, readily learned concepts, are solvable once predicate quality surpasses a low threshold.
    \item \textbf{Conjunctive Rule (Rule 3 - $\textit{is\_mel\_on\_back}$):} This rule requires simultaneous high-fidelity prediction of both $\textit{dx}$ (a concept readily learned) and $\textit{localization}$ (a holistic concept that is poorly learned by object-centric models). The plot clearly illustrates the  perceptual bottleneck : almost all models (DINOv2, ResNet+SA\cite{biza2023invariantslotattentionobject}, and SlotVAE variations) fail on this rule, scoring F1-scores close to zero across all supervision levels below 75\%. This universal failure highlights that the low fidelity of the secondary $\textit{localization}$ concept prevents the successful induction of this complex conjunctive rule.
    \item \textbf{GlobalVAE Success (Rule 3):} The exception is the  GlobalVAE at 75\% supervision , which achieves the highest F1-score of approximately $0.90$. This result is the empirical validation of the architectural finding that a global representation is necessary to ground the scene-centric $\textit{localization}$ concept with sufficient fidelity for complex multi-concept reasoning, a finding reinforced by the latent visualization in \autoref{fig:globalvae_75_latent}.
\end{itemize}
This figure confirms that while downstream reasoners are robust to noise for simple rules, the overall success of the pipeline is critically bottlenecked by the perception module's ability to generate accurate predicates for \textit{all} concepts required by a complex rule, especially when the required concepts clash with the model's inherent inductive bias (e.g., object-centric vs. scene-centric).

% \begin{table}[tb]
% \centering
% \caption{Framework Robustness to VAE Noise (F1-Score)}
% \label{tab:robustness}
% \scriptsize
% \begin{tabular}{@{}lcccc@{}}
% \toprule
% & \multicolumn{2}{c}{\textbf{500 Images}} & \multicolumn{2}{c}{\textbf{5000 Images}} \\
% \textbf{Rule} & \textbf{Best} & \textbf{F1-Score} & \textbf{Best} & \textbf{F1-Score} \\
% \midrule
% Cardinality & BN & 0.667 & BN & 0.723 \\
% Conjunctive & ILP & 1.000 & ILP & 1.000 \\
% Disjunctive & ILP & 1.000 & ILP & 1.000 \\
% Simple Exist & ILP & 1.000 & ILP & 1.000\\
% Univ. Quant & BN & 0.762 & NS-CL & 0.897 \\
% \bottomrule
% \end{tabular}
%\end{table}

%%%%%%%%%%%%%%%%%%%%%%%%%%%%%%%%%%%%%%%%%%%%%%%%%%%%%%%%%%%%%%%%%%%%%%
% Bibliography
%%%%%%%%%%%%%%%%%%%%%%%%%%%%%%%%%%%%%%%%%%%%%%%%%%%%%%%%%%%%%%%%%%%%%%
{
    \small
    \bibliographystyle{ieeenat_fullname}
    \bibliography{literature}
}